%% file: S2S_single.tex
\documentclass[runningheads]{llncs}
\usepackage{comment} 
\usepackage{amsmath,amssymb} 
\usepackage{color}
\usepackage{multirow}
\usepackage{graphicx}
\usepackage{times}
\usepackage{epsfig}
\usepackage{url}
\usepackage{mathtools}
\usepackage{subfigure}
\usepackage{hyperref}
\usepackage{blindtext}
\usepackage{caption}
\input{definitions}
\usepackage[width=122mm,left=12mm,paperwidth=146mm,height=193mm,top=12mm,paperheight=217mm]{geometry}
\makeatletter
\newcommand{\printfnsymbol}[1]{%
  \textsuperscript{\@fnsymbol{#1}}%
}
\makeatother

\begin{document}
\pagestyle{headings}
\mainmatter
\def\ECCVSubNumber{5906}  

\title{Sound2Sight: Generating Visual Dynamics \\from Sound and Context} 

\titlerunning{Sound2Sight: Generating Visual Dynamics from Sound and Context}
%
\author{
Anoop Cherian\thanks{Equal contribution. Work done while MC was interning at MERL.}\inst{1}\quad
Moitreya Chatterjee\printfnsymbol{1}\inst{2}\quad
Narendra Ahuja\inst{2} }
\authorrunning{Cherian et \textit{al.}}
%
\institute{Mitsubishi Electric Research Laboratories, Cambridge MA 02139, USA \and
University of Illinois at Urbana-Champaign, Urbana IL 61801, USA \\
\email{cherian@merl.com} \quad \email{metro.smiles@gmail.com} \quad \email{n-ahuja@illinois.edu}}
\maketitle

\begin{abstract}
  Learning associations across modalities is critical for robust multimodal reasoning, especially when a modality may be missing during inference. In this paper, we study this problem in the context of audio-conditioned visual synthesis -- a task that is important, for example, in occlusion reasoning. Specifically, our goal is to generate future video frames and their motion dynamics conditioned on audio and a few past frames. To tackle this problem, we present \emph{Sound2Sight}, a deep variational framework, that is trained to learn a per frame stochastic prior conditioned on a joint embedding of audio and past frames. This embedding is learned via a multi-head attention-based audio-visual transformer encoder. The learned prior is then sampled to further condition a video forecasting module to generate future frames. The stochastic prior allows the model to sample multiple plausible futures that are consistent with the provided audio and the past context. Moreover, to improve the quality and coherence of the generated frames, we propose a multimodal discriminator that differentiates between a synthesized and a real audio-visual clip. We empirically evaluate our approach, vis-\'a-vis closely-related prior methods, on two new datasets  viz. (i) Multimodal Stochastic Moving MNIST with a Surprise Obstacle, (ii) Youtube Paintings; as well as on the existing Audio-Set Drums dataset. Our extensive experiments demonstrate that Sound2Sight significantly outperforms the state of the art in the generated video quality, while also producing diverse video content. 
\end{abstract}

\section{Introduction} 
Evolution has equipped the intelligent species with the ability to create mental representations of sensory inputs and make associations across them to generate world models~\cite{corlett2018conditioned}. Perception is the outcome of an inference process over this world model, when provided with new sensory inputs.  Consider the following situation. You see a kid going into a room which is occluded from your viewpoint, however after sometime you hear the sound of a vessel falling down, and soon enough, a heavy falling sound. In the blink of an eye, your mind simulates a large number of potential possibilities that could have happened in that room; each simulation considered for its coherence with the sound heard, and its urgency or risk. From these simulations, the most likely possibility is selected to be acted upon. Such a framework that can synthesize modalities from other cues is perhaps fundamental to any intelligent system. Efforts to understand such mental associations between modalities dates back to the pioneering work of Pavlov~\cite{pavlov1910work} (on his drooling dogs) who proposed the idea of \emph{conditioning} on sensory inputs. 

In this paper, we explore this multimodal association problem in the context of generating plausible visual imagery given the accompanying sound. Specifically, our goal is to build a world model that learns associations between audio and video dynamics in such a way as to infer visual dynamics when only the audio modality (and the visual context set by a few initial frames) is presented to the system. As alluded to above, such a problem is fundamental to occlusion reasoning. Apart from this, it could help develop assistive technologies for the hearing-impaired, could enable a synergy between video and audio inpainting technologies~\cite{kim2019deep,zhou2019vision}, or could even compliment the current ``seeing through corners'' methods~\cite{lindell2019acoustic,zhao2018through} using the audio modality.

\begin{figure}[t]
    \begin{center}
    \includegraphics[width=12cm,trim={0cm 7.2cm 1.5cm 0cm},clip]{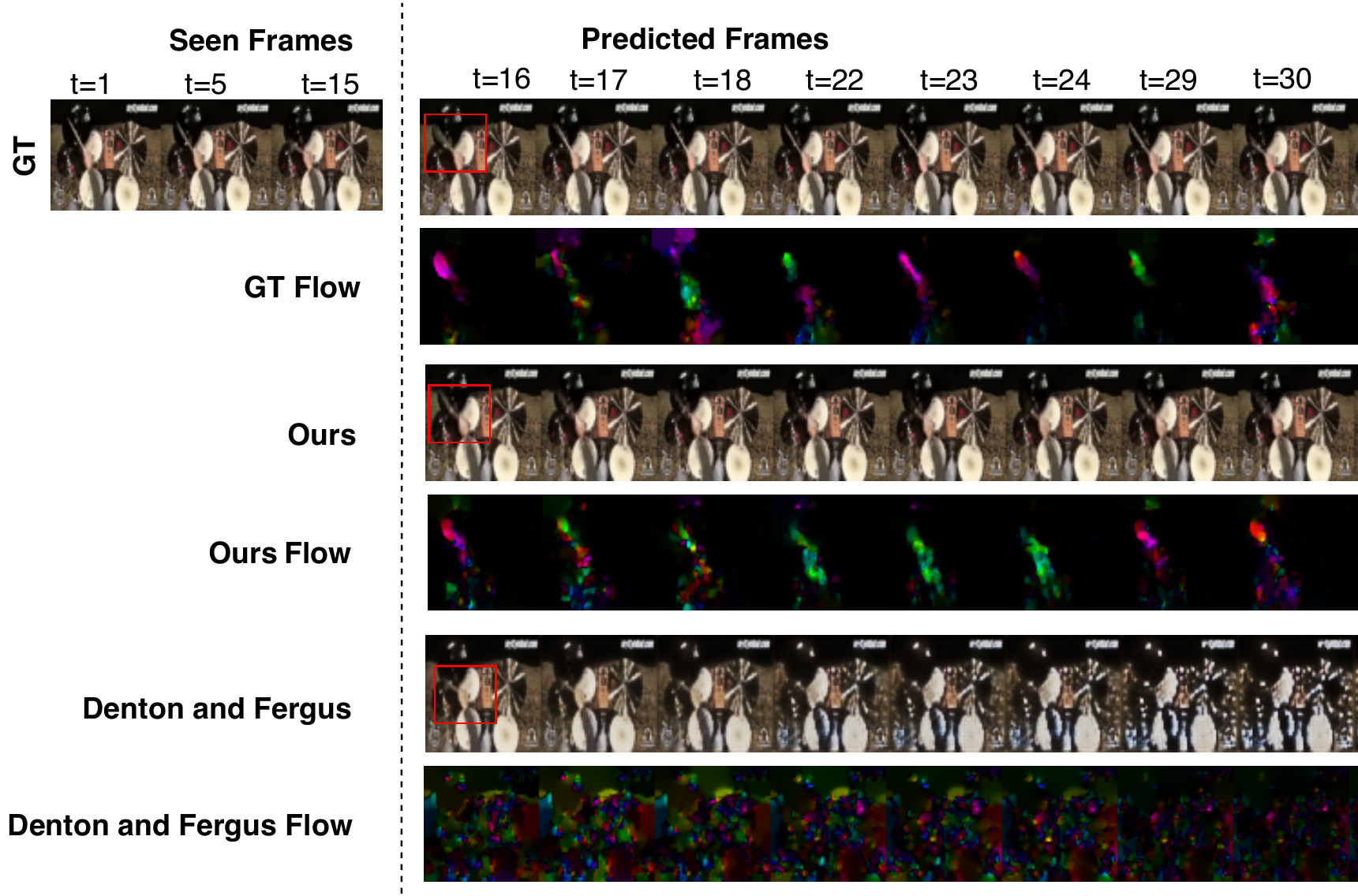}
    \end{center}
   \caption{Video generation using our Sound2Sight against Denton and Fergus~\cite{denton2018stochastic} on AudioSet-Drums~\cite{gemmeke2017audio}. We also show the optical flow between consecutive generated frames. The red square indicates the region of dominant motion.} 
   \vspace{-0.5cm}
    \label{fig:first_samplel}
\end{figure}

From a technical standpoint, the task of generating the pixel-wise video stream from only the audio modality is severely ill-posed. For instance, a drummer playing a drum to a certain beat would sound the same irrespective of the color of his/her attire.  To circumvent this challenge, we condition our video generator using a few initial frames. This workaround not only permits the generation of videos that are pertinent to the situation, but also allows the model to focus on learning the dynamics and interactions of the visual cues assisted by audio. There are several recent works in a similar vein ~\cite{chen2019hierarchical,vougioukas2018end,wav2pix2019icassp} that explore speech-to-video synthesis to generate talking heads, however they do not use the past visual context or assume very restricted motion dynamics and audio priors. On the other hand, methods that seek to predict future video frames~\cite{denton2018stochastic,vondrick2016generating,fragkiadaki2015learning} given only the past frames, assume a continuity of the motion pattern and are unable to adapt to drastic changes in motion that might arise in the future (e.g., the sudden movements of the drummer in Figure~\ref{fig:first_samplel}). We note that there also exist several recent works in the audio-visual synthesis realm, such as generating audio from video~\cite{kidron2005pixels,snd_motions,zhao2018sound} that looks at a complementary problem and multimodal generative adversarial networks (GAN) that generates a single image rather than forecasting the video dynamics ~\cite{chen2017deep,hao2018cmcgan,wan2019towards}.

To tackle this novel task, we present a stochastic deep neural network:~\emph{Sound2Sight}, which is trained end-to-end. Our main backbone is a conditional variational autoencoder (VAE)~\cite{kingma2013auto} that captures the distribution of the \emph{future video frames} in a latent space. This distribution is used as a prior to subsequently condition a video generation framework. A key question that arises then, is how to incorporate the audio stream and its correlations with the video content? We propose to capture this synergy within the prior distribution - through a joint embedding of the audio features and the video frames. The variance of this prior distribution, permits diversity in the video generation model, thereby synthesizing disparate plausible futures. 

An important component in our setup is the audio-visual latent embedding that controls the generation process. Inspired by the recent success of transformer networks~\cite{vaswani2017attention}, we propose an adaptation of multi-head transformers to effectively learn a multimodal latent space through self-attention. As is generally known, pixel generations produced using variational models often lack sharpness, which could be attributed to the Euclidean loss typically used~\cite{lamb2016discriminative}. To this end, in order to improve the generated video quality, we further propose a novel \emph{multimodal discriminator}, that is trained to differentiate between real audio-visual samples and generated video frames coupled with the input audio. This discriminator incorporates explicit sub-modules to verify if the generated frames are realistic, consistent, and synchronized with the audio. 

We conduct experiments on three datasets, two new multimodal datasets: (i) Multimodal Stochastic Moving MNIST with a Surprise Obstacle (M3SO) and (ii) Youtube-Painting, alongside a third dataset -- AudioSet-Drums -- which is an adaptation of the well-known AudioSet datset~\cite{gemmeke2017audio}. The M3SO dataset is an extension of stochastic moving MNIST~\cite{denton2018stochastic}, however incorporates audio based on the location and identity of the digits in the video, while also including a surprise component that requires learning audio-visual synchronization and stochastic reasoning. The Youtube-Painting dataset is created via crawling Youtube for painting videos and provides a challenging setting for Sound2Sight to associate painting motions of an artist and the subtle sounds of brush strokes. Our experiments on these datasets show that Sound2Sight leads to state-of-the-art performances in quality, diversity, and consistency of the generated videos. 

Before moving on, we summarize below the key contributions of this paper.
\begin{itemize}
    \vspace*{-2pt}
    \item We study the novel task of future frame generation consistent with the given audio and a set of initial frames. 
    \item We present \emph{Sound2Sight}, a novel deep variational multimodal encoder-decoder for this task, that combines the power of VAEs, GANs, and multimodal transformers in a coherent learning framework. 
    \item We introduce three datasets for evaluating this task. Extensive experiments are provided, demonstrating state-of-the-art performances, besides portraying diversity in the generation process.
\end{itemize}

\section{Related Works}
In this section, we review prior works that are closely related to our approach.

\noindent\textbf{Audio-Visual Joint Representations:} The natural co-occurrence of audio-and-visual cues is used for better representation learning in several recent works ~\cite{arandjelovic2017look,aytar2016soundnet,harwath2016unsupervised,owens2018audio,owens2016visually,owens2016ambient}. We too draw upon this observation, however, our end-goal of future frame generation from audio is notably different and manifests in our proposed architecture. For example, while both Arandjelovi\'c and Zisserman~\cite{arandjelovic2017look} and Owens and Efros~\cite{owens2018audio} propose a common multimodal embedding layer for video representation, our multimodal embedding module is only used for capturing the prior and posterior distributions of the stochastic components in the generated frame. 

\noindent\textbf{Video Generation:} The success of GANs has resulted in a myriad of image generation algorithms~\cite{deshpande2018generative,goodfellow2014generative,gulrajani2017improved,kingma2013auto,kolouri2018sliced,liu2017unsupervised,wu2019sliced}. Inspired from these techniques, methods for video generation have also been proposed~\cite{saito2017temporal,tulyakov2018mocogan,vondrick2016generating}. These algorithms usually directly map a noise vector sampled from a known or a learned distribution into a realistic-looking video and as such are known as \textit{unconditional video generation} methods. Instead, our proposed generative model uses additional audio inputs, alongside encoding of the past frames. Models like ours are therefore, typically referred to as \textit{conditional video generation} techniques. Prior works~\cite{gupta2018imagine,li2018video,hao2018controllable,pan2019video,wang2018video} have shown the success of conditional generative methods when information, such as the video categories, captions, etc., are available, using which constraints the plausible generations, improving their quality.
Our proposed architecture differs in the modalities we use to constrain the generations and the associated technical innovations required to accommodate them.

\noindent\textbf{Video Prediction/Forecasting:} This is the task of predicting future frames, given a few frames from the past. Prior works in this area typically fall under: (i) \textit{Deterministic}, and (ii) \textit{Diversity-based} methods. Deterministic methods often use an encoder-decoder model to generate video frames autoregressively. The inherent stochasticity within the video data (due to multiple plausible futures or encoding noise) is thus difficult to be incorporated in such models~\cite{ranzato2014video,villegas2017decomposing,finn2016unsupervised,jia2016dynamic,luo2017unsupervised,srivastava2015unsupervised,hsieh2018learning}. Our approach circumvents these issues via a stochastic module. There have been prior efforts to capture this stochasticity from unimodal cues, such as~\cite{walker2017pose,denton2018stochastic,xue2016visual,babaeizadeh2017stochastic}, by learning a parametric prior distribution. Different from these approaches, we model the stochasticity using multimodal inputs. 

We also note that there are several works in the area of generating human face animations conditioned on speech~\cite{jamaludin2019you,karras2017audio,shlizerman2018audio,suwajanakorn2017synthesizing,taylor2017deep}, however these techniques often make use of additional details, such as the identity of the person or leverage strong facial cues such as landmarks, textures, etc. - hindering their applicability to generic videos. There are methods free of such constraints, such as~\cite{oh2019speech2face}, however they synthesize images and not videos. A work similar to ours is Vougioukas \etal~\cite{vougioukas2018end} that synthesizes face motions directly from speech and an initial frame, however it operates in the very restricted domain of generating facial motions only. 

\section{Proposed Method}
\begin{figure}[t]
    \includegraphics[width=12cm, clip]{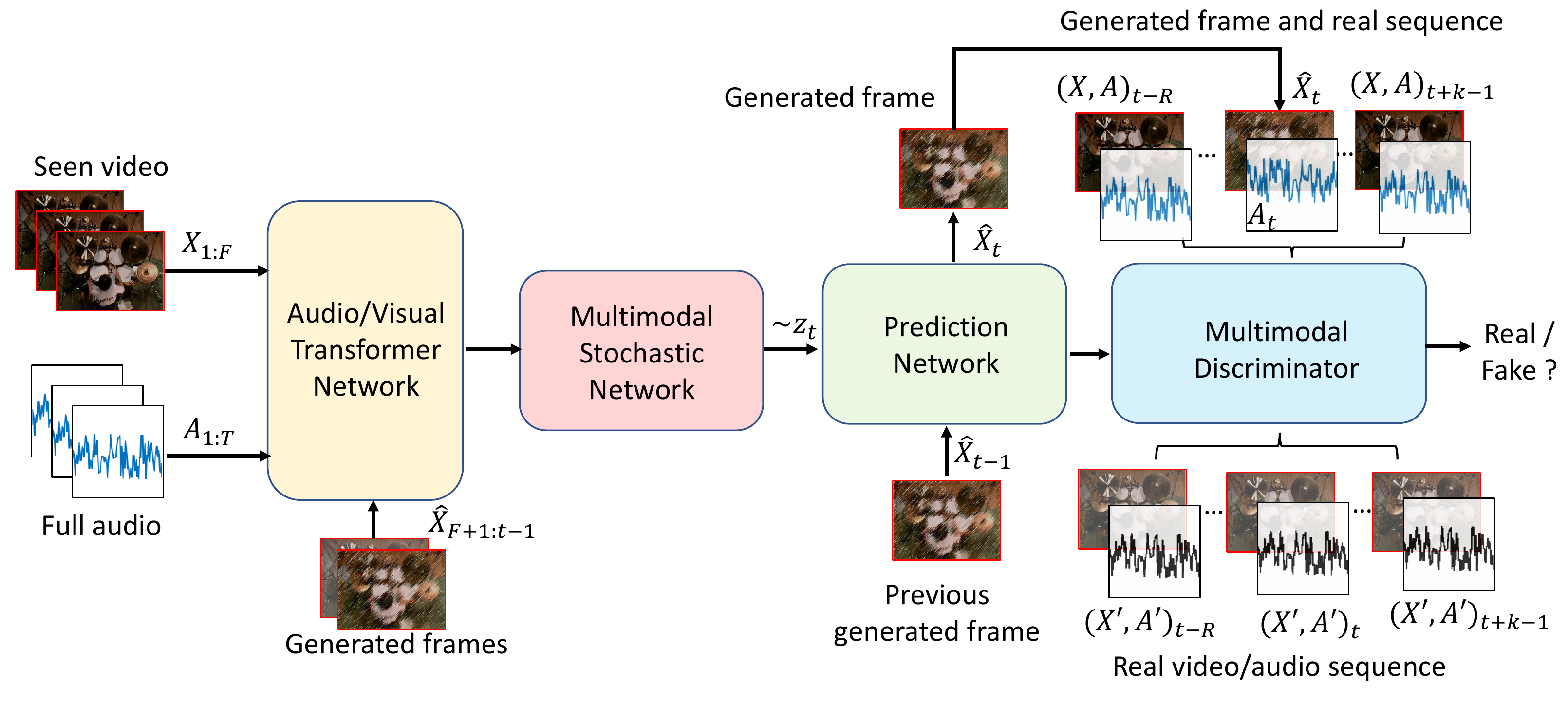} 
   \caption{Overview of the architecture of Sound2Sight. Our model takes $F$ ``seen'' video frames (during inference) and all $T$ audio samples, producing $T-F$ video frames (each denoted by $\hat{X}_t$). During training, the multimodal discriminator predicts if an input video is real or fake. We construct the fake video by replacing the $t$-th frame of the ground truth by $\hat{X}_t$. Note that during training, the generated frames ($\hat{X}_t$) which are input to the audio/visual transformer, are replaced by their real counterparts ($X_t$).} 
    \label{fig:overview_arch}
\end{figure}

Given a dataset $\dataset=\set{\vid_1,\vid_2,\cdots, \vid_N}$ consisting of $N$ video sequences, where each $\vid$ is characterized by a pair $(X_{1:T}, A_{1:T})$ of $T$ video frames and its time-aligned audio samples, i.e., $X_{1:T}=\seq{X_1, X_2, ..., X_F, X_{F+1}, ..., X_{T}}$ and $A_{1:T} = \seq {A_1, A_2, ..., A_{T}}$. We assume that the audio and the video are synchronized in such a way that $A_t$ corresponds to the sound associated with the frame $X_t$ in the duration $(t, t+1)$. Now, given as input a sequence of $F$ frames $X_{1:F}$, ($F<T$) and the audio $A_{1:T}$, our task is to generate frames $\hat{X}_{F+1:T}$ that is as realistic as possible compared to the true frames $X_{F+1:T}$. Given the under-constrained nature of the audio to video generation problem, we empirically show that it is essential to provide the past frames $X_{1:F}$ to set the visual context besides providing the audio input.

\noindent\textbf{Sound2Sight Architecture:}
In this section, we first present an overview of the proposed model, before discussing the details. Figure~\ref{fig:overview_arch} illustrates the key components in our model and the input-output data flow. In broad strokes, our model follows an encoder-decoder auto-regressive generator architecture, generating the video sequentially one frame at a time. This generator module has two components, viz. the \textit{Prediction Network} and the \textit{Multimodal Stochastic Network}. The former module takes the previous frame $X_{t-1}$ as input,\footnote{$X_{t-1}$ is the \emph{real} frame during training, however during inference, it is the generated frame $\hat{X}_{t-1}$ if $t-1>F$.} encodes it into a latent space, concatenates it with a prior latent sample $z_t$ obtained from the stochastic network, and decodes it to generate a frame $\hat{X}_{t}$, which approximates the target frame $X_t$.  Sans the sample $z_t$, the prediction network is purely deterministic and unimodal, and hence can fail to capture the stochasticity in the motion dynamics. This challenge is mitigated by the multimodal stochastic network, which uses transformer encoders~\cite{vaswani2017attention} on the audio and visual input streams to produce (the parameters of) a prior distribution from which $z_t$ is sampled. The generator can thus be thought of as a non-linear heteroskedastic dynamical system (whose variance is decided by an underlying neural network), which generates $\hat{X}_{t}$ from the pair $(\hat{X}_{t-1}, z_t)$, and implicitly conditioned on the (latent) history of previous samples and the given audio.

During training, two additional data flows happen. (i) The transformer and the stochastic network take as input the true video sequence $X_{1:t}$ as well. This is used to estimate a posterior distribution which is in turn used to train the stochastic prior so that it effectively captures the distribution of real video samples. (ii) Further, the generated frames are evaluated for their realism and audio-visual alignment using a multimodal adversarial discriminator~\cite{goodfellow2014generative} (shown in Figure~\ref{fig:overview_arch}). This discriminator uses $\hat{X}_{t}$ -- the synthetic frame, inserted at the $t$-th index of the original sequence, and $X_{t-R:t+(k-1)}$ the set of $R$ past, and $(k-1)$ future frames, along with the corresponding audio, and compares it with real (arbitrary) audio-visual clips of length $R+k$ from the dataset. Since discriminators match distributions, rather than matching individual samples, this ensures that incorporating the generated frame $\hat{X}_{t}$ results in a coherent video that is consistent with the input audio, while permitting diversity. In the following, we revisit each of the above modules in greater detail and layout our training strategy. 

\begin{figure}[t]
    \centering
    \includegraphics[width=12cm,trim={0cm 1.0cm 3.5cm 6cm},clip]{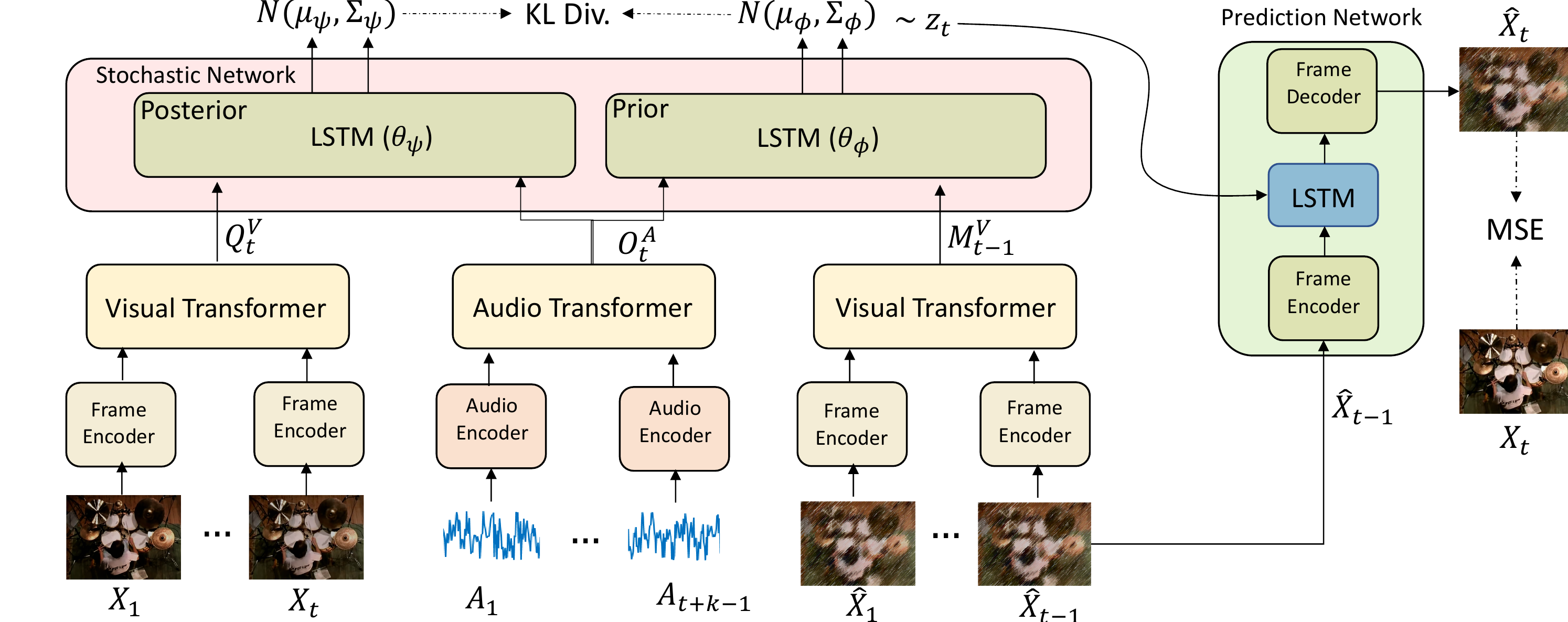} 
   \caption{Details of our Multimodal Stochastic Network and our Prediction Network.} 
    \label{fig:msn_pn_modules}
\end{figure}

\noindent\textbf{Prediction Network:}
Broadly speaking, the prediction network (PN) is a standard sequence-to-sequence encoder-decoder network. It starts off by embedding the previous frame $X_{t-1}$ into a latent space. We denote this embedding by $f(X_{t-1})$, where $f(\cdot)$ abstracts a convolutional neural network (CNN)~\cite{lecun1998gradient}. Each layer of this CNN consists of a set of convolution kernels, followed by 2D-Batch Normalization~\cite{ioffe2015batch} layers, Leaky ReLU activations, and has skip-connections to the decoder part of the network. These skip connections facilitate reconstruction of static parts of the video~\cite{ronneberger2015u}. The embedding of the frame $f(X_{t-1})$ is then concatenated with a sample $z_t\sim\normal(\mu_{\phi},\Sigma_{\phi})$, a Gaussian prior provided by the stochastic module (described next) where $\mu_{\phi}$ and $\Sigma_{\phi}$ denote the mean and a diagonal covariance matrix of this Gaussian prior. Our key idea is to have $z_t$ capture the cues about the future as provided by the available audio, as well as the randomness in producing the next frame. We then feed the pair $\tuple{f(X_{t-1}), z_t}$ to a Long-Short Term Memory (LSTM)~\cite{hochreiter1997long}, parametrized by $\theta_L$ within the PN; this LSTM keeps track of the previously generated frames via its internal states. Specifically, if $h_{t-1}$ denotes the hidden state of this LSTM, then we define its output $\eta_t$ as:

\begin{equation}
 \eta_t = \lstm_{\theta_L}\!\!\left(\tuple{f(X_{t-1}), z_t},  h_{t-1}\right).
\end{equation}

\noindent The LSTM output $\eta_t$ is then passed to the decoder network $g$, to generate the next frame, i.e., $\hat{X}_{t} = g(\eta_t)$. The decoder consists of a set of deconvolution layers with Leaky ReLU activations, coupled with 2D-Batch Normalization layers.

\noindent \textbf{Multimodal Stochastic Network:} 
Several prior works have underscored the importance of modeling the stochasticity in video generation~\cite{babaeizadeh2017stochastic,denton2018stochastic,walker2017pose,xue2016visual}, albeit using a single modality. Inspired by these works, we introduce the multimodal stochastic network (MSN) that takes both the audio and video streams as inputs to model the stochastic elements in producing the target frame $X_t$. As alluded to earlier, such a stochastic element allows for capturing the randomness in the generated frame, while also permitting the sampling of multiple plausible futures conditioned on the available inputs.
As shown in Figure~\ref{fig:msn_pn_modules}, the stochastic network is effectuated by computing a prior and a posterior distribution in the embedding space (from which $z_t$ is sampled) and training the model to minimize their mutual discrepancy. The prior distribution is jointly conditioned on an embedding of the audio sub-clip $A_{1:t+(k-1)}$ and an embedding of the video frames $X_{1:t-1}$, both obtained via transformer encoders. We denote the $t$-th audio encoding by $O^A_{t}$, while the $(t-1)$-th video encoding is denoted by $M^V_{t-1}$. Let the prior distribution be $p_{\phi} (z_t | O^A_{t}, M^V_{t-1})$, parametrized as a Gaussian, with mean $\mu_\phi$ and diagonal covariance $\Sigma_\phi$. Likewise, the posterior distribution $p_{\psi} (z_t | O^A_{t}, Q^V_{t})$, which is also assumed to be a Gaussian $\normal(\mu_{\psi}, \Sigma_{\psi})$, is jointly conditioned on audio clips $A_{1:t+(k-1)}$ and visual frames $X_{1:t}$. Its audio embedding is shared with the prior distribution and its visual input is obtained from the $t$-th transformer encoding is denoted  $Q^V_{t}$. Here, it is worth noting that the visual conditioning of the prior distribution, unlike the posterior, is \emph{only upto frame $t-1$}, i.e. the past visual frames. Since the posterior network has access to the $t$-th frame in its input, it may attempt to directly encode this frame to be decoded by the prediction network decoder to produce the next frame. However, due to the KL-divergence loss between the prior and the posterior distributions, such a direct decoding cannot happen; unless the prior is trained well such that the KL-loss is minimized; which essentially implies the prior $p_{\phi} (z_t | O^A_{t}, M^V_{t-1})$ will be able to predict the latent distribution of the future samples (as if from the posterior $p_{\psi} (z_t | O^A_{t}, Q^V_{t})$), which is essentially what we require during inference.

To generate the prior distribution, we concatenate the embedded features $M^V_{t-1}$ and $O^A_{t}$ as input to an $\lstm_\phi$. Different from standard LSTMs, this LSTM predicts the parameters of the prior distribution directly, i.e.,

\begin{equation}
\mu_\phi, \log \Sigma_\phi = \lstm_\phi(O^A_{t}, M^V_{t-1}).
\end{equation}
\noindent The posterior distribution parameters are estimated similarly, using a second LSTM, denoted $\lstm_\psi$ that takes as input the embedded and concatenated audio-video features $O^A_{t}$ and $Q^V_{t}$ to produce:

 \begin{equation}
 \mu_\psi, \log \Sigma_\psi = \lstm_{\psi}( O^A_{t}, Q^V_{t}).
 \end{equation}

\noindent\textbf{Audio-Visual Transformer Encoder:}
Next, we describe the process of producing the prior and posterior distributions from audio-visual joint embeddings. As we want these embeddings to be ``temporally-conscious'' while computable efficiently, we bank on the very successful \textit{Transformer Encoder Networks}~\cite{vaswani2017attention}, which are armed with self-attention modules that are well-known to produce powerful multimodal representations. Re-using the encoder CNN $f$ from the prediction network, our visual transformer encoder takes as input the matrix $\mathcal{F}=\seq{f'(X_1), f'(X_2), \cdots, f'(X_{t-1})}$ with $f'(X_i)$ in its $i$-th column, where $f'(X_i)$ denotes the feature encoding $f(X_i)$ augmented with the respective temporal position encoding of the frame in the sequence, as suggested in~\cite{vaswani2017attention}.
We then apply $\ell$-head self-attention to $\mathcal{F}$ by designing \textit{Query} ($\mathcal{Q}$), \textit{Key} ($\mathcal{K}$), and \textit{Value} ($\mathcal{V}$) triplets via linear projections of our frame embeddings $\mathcal{F}$; i.e., $\mathcal{Q}_j=W_q^j\mathcal{F}$, $\mathcal{K}_j=W_k^j\mathcal{F}$, and $\mathcal{V}_j=W_v^j\mathcal{F}$, where $W^j_q,W_k^j,W_v^j$ are matrices of sizes $d_k\times d$, $d$ is the size of the feature $f'$, and $j=1,2,\cdots, \ell$. Using an $\ell d_k\times d$ weight matrix $W_h$, our self-attended feature $\hat{M}_{t-1}^V$ from this transformer layer is thus:

\begin{equation}
    \hat{M}_{t-1}^V = \concat_{j=1}^{\ell}\left(\softmax\left(\frac{\mathcal{Q}_j\mathcal{K}_j^{\top}}{\sqrt{d_k}}\right)\mathcal{V}_j\right)\!W_h,
\end{equation}
\noindent where $\concat$ denotes the concatenation operator. We use four consecutive self-attention layers within every transformer encoder, which are then combined via feed-forward layers to obtain the final encoding~\cite{vaswani2017attention} $M^V_{t-1}$, which is subsequently used in the MSN module. Likewise, the re-purposed visual features for the posterior distribution, $Q^V_{t}$, can also be computed by employing a separate transformer encoder module, which ensures a separation of the visual components of the prior and the posterior networks. To produce the audio embeddings $O^A_{t}$, we first compute STFT (Short-Time Fourier Transform) features $(S_1, S_2, ..., S_{t+(k-1)})$ from the raw audio by choosing appropriate STFT filter sizes and strides, where each $S_i \in \mathbb{R}^{d_{H_A} \times d_{W_A}}$ and encode them using an audio transformer. 

\noindent \textbf{Generator Loss:} To train our generator model, we directly maximize the variational \textit{Empirical Lower BOund} (ELBO)~\cite{kingma2013auto} by optimizing the following objective: 

\begin{equation*} 
\mathcal{L}_\textit{V} = \sum_{t=F+1}^T \expect_{z_t \sim p_{\phi}} \log p_{\phi}(\hat{X}_t | M^V_{t-1}, z_{t})\; - \beta \kl\left(p_{\psi}\!\left(z_t | O^A_{t}, Q^V_{t}\right) \dd p_{\phi}\!\left(z_t | O^A_{t}, M^V_{t-1}\right)\right),
\end{equation*}

\noindent where the KL-divergence matches the closeness of the posterior distribution and the prior, while $\beta$ is a regularization constant. Casting the above objective as a minimization and approximating the first term by the pixel-wise $\ell_2$ error, reduces the objective to: 
\begin{equation}
  \begin{aligned}
\mathcal{L}_\textit{V} \approx \sum_{t=F+1}^T \lVert X_t - \hat{X}_t \rVert^2 _2 \; + \; \beta KL( p_{\psi} || p_{\phi} ).
  \end{aligned}
  \label{eq:l2_vae}
\end{equation}

\noindent\textbf{Multimodal Discriminator Network:}
Matching the synthesized frames against the ground-truth using only pixel-wise loss, as in~\eqref{eq:l2_vae}, implies biasing one possible outcome over other plausible ones; which can discourage generative diversity. We seek to rectify this shortcoming by further scrutinizing the output of our PN module using a multimodal discriminator (see Fig.~\ref{fig:overview_arch}). This discriminator is designed to match the distribution of synthesized frames against the ground truth distribution of audio-video pairs. In contrast to conventional image-based GAN discriminators~\cite{brock2018large,goodfellow2014generative}, our variant couples a classifier, denoted $D_{std}$, and an LSTM $D$, to produce binary labels indicating if the $t$-th frame is from the real data (drawn from the distribution $p_{\dataset}$) or a synthetic distribution produced by the generator $p_G$. This is done via using a set of ground-truth audio-visual frames from the neighborhood of the generated frame, where this neighborhood spans the previous $R$ and future $(k-1)$ frames. When judging its inputs, the discriminator, besides looking into whether the $t$-th frame appears real or fake, also looks at how well the regularities of object motions are preserved with respect to the neighborhood via a motion dynamics (MD) loss, and if the frames are synchronized with the audio via an audio alignment (AA) loss. With these additional terms, our discriminator loss is: 

\begin{equation}\label{eq:gan}
\begin{aligned}
& \mathcal{L}_{\textit{adv}} = -\!\!\sum_{t=F+1}^T \expect_{X'_t \sim p_\dataset}\!\!\log D_{std}(X'_t)\!+\!\expect_{\hat{X}_t \sim p_{G}} \log (1-D_{std}(\hat{X}_t))\; 
\\ & + \;\!\!\!\expect_{X'_t \sim p_{\dataset}}\!\!\!\!\log \underbrace{D(X'_t| A'_t, B'_{t+(k-1)}, \cdots, B'_{t+1}, B'_{t-1}, \cdots, B'_{t-R})}_\textrm{Real Data - Motion Dynamics (MD)}\; \\
&+ \!\!\!\expect_{X'_t\sim p_\dataset}\!\!\!\! \log\underbrace{(1- D(X'_t|A'_{t'}, C'_{t+(k-1), t'+(k-1)}, \cdots, C'_{t+1, t'+1}, C'_{t-1, t'-1},\!\cdots,\!C'_{t-R, t'-R}))}_\textrm{Real Data - Audio Alignment (AA)}\; \\
& + \; \!\!\!\expect_{\hat{X}_t\sim p_G}\!\!\!\!\log\underbrace{(1 - D(\hat{X}_t|A_t, B_{t+(k-1)}, \cdots, B_{t+1}, B_{t-1}, ...,B_{t-R}))}_\textrm{Synthetic Frame - Motion Dynamics (MD)}\; \\
&+ \!\!\!\expect_{\hat{X}\sim p_G}\!\!\!\! \log\underbrace{(1- D(\hat{X}_t|A_{t'}, C_{t+(k-1), t'+(k-1)}, \cdots, C_{t+1, t'+1}, C_{t-1, t'-1},\!\cdots,\!C_{t-R, t'-R}))}_\textrm{Synthetic Frame - Audio Alignment (AA)}\; \\
\end{aligned}
\end{equation}

\noindent where $(X'_t, A'_t)$ denotes a visual frame $X'_t$ and its associated audio $A'_t$ from a clip $B'=(X'_{1:T}, A'_{1:T})$ arbitrarily sampled from the training set. Similarly, we define $C_{t, t'} = (X_t, A_{t'})$, $t' \neq t$, $B_{t} = (X_t, A_{t})$, $C'_{t, t'} = (X'_t, A'_{t'})$, $B'_{t} = (X'_t, A'_{t})$, $X_t \neq X'_t$, $A_t \neq A'_t$. The first term in~\eqref{eq:gan} defines a standard image-based GAN loss, while $D$ in the other terms denotes a convolutional LSTM. The motion dynamics loss captures the consistency of the generated frame against other frames in the sequence (i.e., $X_t'$ against $B'$ on the real, and $\hat{X}_t$ against $B$ on the generated), while the audio alignment of the generated frame $\hat{X}_t$ against arbitrary audio samples $A'$ is captured in the last term. We optimize for the discriminator parameters via minimizing this loss above. 

Combining the adversarial losses above with~\eqref{eq:l2_vae}, our final objective for optimizing the generator is:

\begin{equation}
    \mathcal{L} = \mathcal{L}_\textit{V} \; - \; \gamma \mathcal{L}_{\textit{adv}},
    \label{eq:loss}
\end{equation}

\noindent where $\gamma$ is a constant. We minimize~\eqref{eq:loss} using ADAM~\cite{kingma2014adam}, while employing the reparameterization trick~\cite{kingma2013auto} to ensure differentiability of the stochastic sampler.

\section{Experiments}
\label{sec:expts-MS}
To benchmark the performance of our model, we present empirical experiments on a synthetic and two real world datasets, which will be made publicly available.

\noindent \textbf{Multimodal MovingMNIST with a Surprise Obstacle (M3SO):} is a novel extension of the stochastic MovingMNIST dataset~\cite{denton2018stochastic} adapted to our multimodal setting. Stochastic MovingMNIST consists of digits from the MNIST dataset~\cite{lecun1998gradient} moving along rectilinear paths in a fixed size box ($48 \times 48$) which bounce in random directions upon colliding with the box boundaries. In M3SO, we make the following changes: (i) we equip each digit with a unique tone, (ii) the amplitude of this tone is inversely proportional to the digit's distance from the origin, and (iii) the tone changes momentarily when the digit bounces off the box edge. We make this task even more challenging by introducing an obstacle (square block of fixed size) at a random location within the unseen part of the video. When the digit bounces against the block, a unique audio frequency is emitted. The task on this dataset is not only to generate the frames, but also to predict the location of the block by listening to the tone changes. See supplementary materials for details. We also construct a version of the dataset, where no block is introduced, called M3SO-NB. We produced 8,000 training, 1,000 validation, and 1,000 test samples for both M3SO and M3SO-NB. 

\noindent\textbf{AudioSet-Drums:} includes videos from the \emph{Drums} class of the AudioSet dataset~\cite{gemmeke2017audio}. We clipped and retained only those video segments from this dataset for which the drum player is visible when the drum beat is heard. This yielded a dataset consisting of 8K clips which we split as  6K for training, 1K for validation, and 1K for test. Each video is of $64 \times 64$ resolution, 30fps, and is 3 seconds long. 

\noindent\textbf{YouTube Painting}: To analyze Sound2Sight in a subtle, yet real world setting, we introduce the \textit{Youtube Painting} dataset. The videos in this dataset are manually collected via crawling painting videos on Youtube~\cite{Youtube:Taylor}. We selected only those videos that contain a painter painting on a canvas in an indoor environment, and which have a clear audio of the brush strokes. These videos provide a wide assortment of brush strokes and painting colors. The painter's motions and the camera viewpoints are often arbitrary which adds to the complexity and diversity, making it a very challenging dataset. Here the task is to generate video frames showing the dynamics of the painter's arms, while preserving the static components in the scene. We collected 4.8K videos for training, 500 for validation and 500 for test.  Each video is of $64\times 64$ resolution, 30fps, and 3s long.

\noindent\textbf{Evaluation Setup:} On the M3SO dataset, we conduct experiments in two settings: (i) in M3SO-NB, all methods are shown 5 frames and the full audio, with the task to predict the next 15 frames at training and 20 frames at test time, and (ii) using M3SO in which blocks are presented, we show 30 frames at training and 30 frames are predicted, however the block appears at the 42-nd frame. We predict 40 frames at test time. For both the real-world datasets, we train all algorithms on 15 seen frames and task is to predict the next 15 during training and 30 at test.

\noindent \textbf{Baselines:} To the best of our knowledge, this is the first attempt at future frame generation from audio applied to generic videos. To this end, we compare our algorithm against the following closely related baselines which we adapt to make them applicable to our setting: (i) \textit{Audio-Only}: we train a  sequence-to-sequence model~\cite{sutskever2014sequence} taking only the audio as input and generate the frames using an LSTM (thus, the past context is missing), (ii) \textit{Video-Only}, where we design three baselines: (ii-a) the recent work of Denton and Fergus~\cite{denton2018stochastic}, (ii-b) the approach of Hsieh et al.~\cite{hsieh2018learning}, and (iii-c) an ablated variant of our model with the audio modality turned off (Ours - No audio), and (iii) \textit{Multimodal}: in this scenario too, we develop multiple baselines: (iii-a) we compare against the approach of Vougioukas et al.~\cite{vougioukas2018end}, that predicts the video from audio and the first frame, albeit in a very restricted setting of generating faces from speech, (iii-b) the approach of Vougioukas et al.~\cite{vougioukas2018end}, modified to use a set of seen frames rather than one (Multiframe~\cite{vougioukas2018end}), (iii-c) an ablated variant of our model without the AA loss term in the discriminator (Ours - No AA), and (iii-d) an ablated variant of our model without the AA and MD loss terms in the discriminator (Ours - No AA, MD). 

\noindent \textbf{Evaluation Metrics}: We use the standard strutural similarity (SSIM)~\cite{wang2004image} and the Peak Signal to Noise Ratio (PSNR) scores for quantitative evaluation of the quality of the generated frames against the ground-truth.  Additionally, for the real-world datasets, we also report their realism and audio consistency as measured via the discriminator fooling rate, and through human preference evaluation on a subset of our test sets.

\noindent \textbf{Implementation Details}: The PN module uses an LSTM with two layers and produces 128-D frame embeddings. We use 10-D stochastic samples ($z_t$). The prior and posterior LSTMs are both single-layered, each with 256-D inputs from audio-frame embeddings (which are each 128-D). All LSTMs have 256-D hidden states. Each transformer module has one layer and four heads with 128-D feedforward layer. The discriminator uses an LSTM with a hidden layer of 256-D, a frame-history $R=2$, and look-head $k=1$. We train the generator and discriminator jointly with a learning rate of 2e-3 using ADAM~\cite{kingma2014adam}. We set both $\beta$ and $\gamma$ as 0.0001, and increased $\gamma$ by a factor of 10 every 300 epochs. All hyper-parameters are chosen using the validation set. During inference, we sample 100 futures per time step, and use sequences that best matches the ground-truth, for our method and the baselines. 

\subsection{Experimental Results}

\begin{table}[t]
\footnotesize
\centering
\caption{SSIM, PSNR for M3SO-NB and M3SO. \textbf{Highest}, \textcolor{blue}{Second} highest scores.} 
\label{tab:mnist}
\begin{tabular}{l|c|c|c|c||c|c|c}
\hline \hline
\multicolumn{8}{c}{\textit{Experiments with M3SO-NB with 5 seen frames}} \\ \hline
\multirow{1}{*}{\textbf{Method}} & \multirow{1}{*}{\textbf{Type}} & \multicolumn{3}{c}{\textbf{SSIM}} & \multicolumn{3}{c}{\textbf{PSNR}} \\ \cline{3-8}
& & Frame 6 & Frame 15 &	Frame 25 & Frame 6 &	Frame 15 &	Frame 25 \\ \hline
Our Method & Multimodal & \textbf{0.9575}	& \textbf{0.8943} &	\textbf{0.8697} &	21.69 &	\textbf{17.62} &	\textcolor{blue}{16.84} \\ 
Ours - No AA  & Multimodal  & 0.9547 &	0.8584 &	0.8296 &	\textcolor{blue}{21.80} &	\textcolor{blue}{17.36} &	\textbf{16.97} \\
Ours - No AA, MD  & Multimodal  & 0.9477 &	0.8546 &	0.8251 &	21.16 &	16.16 &	15.49 \\ 
Ours - No audio  & Unimodal - V  & \textcolor{blue}{0.9556} &	0.8351 &	0.6920 &	\textbf{22.66} &	15.59 &	12.40 \\ \hline
Multiple Frames - ~\cite{vougioukas2018end} & Multimodal  & 0.9012 &	\textcolor{blue}{0.8690} &	0.8693 &	18.09 &	15.23 &	15.33 \\
Vougioukas \etal ~\cite{vougioukas2018end} & Multimodal  & 0.8600 &	0.8571 &	0.8573 &	15.17 &	14.99 &	15.01 \\ 
 \hline
Denton and Fergus~\cite{denton2018stochastic} & Unimodal - V  & 0.9265 &	0.8300 &	0.7999 &	18.59 &	14.65 &	13.98 \\
Audio Only  & Unimodal - A  & 0.8499 &	0.8659 &	0.8662 &	13.71 &	13.16 &	12.94 \\ \hline
\multicolumn{8}{c}{\textit{Experiments on M3SO with 30 seen frames (Block is introduced in the $42^{nd}$ frame)}} \\ \hline
& & Frame 31 & Frame 42 &	Frame 70 & Frame 31 &	Frame 42 &	Frame 70 \\ \hline
Our Method & Multimodal & \textbf{0.8780}	& \textbf{0.6256} &	\textbf{0.6170} &	\textbf{19.50} &	\textbf{9.39} &	\textbf{9.41} \\ \hline
Multiple Frames - ~\cite{vougioukas2018end} & Multimodal  & \textcolor{blue}{0.8701} &	\textcolor{blue}{0.6073} &	\textcolor{blue}{0.6050} &	\textcolor{blue}{15.41} &	8.53 &	8.53 \\
Vougioukas \etal ~\cite{vougioukas2018end} & Multimodal  & 0.8681 &	0.6009 &	0.6007 &	15.17 &	8.48 &	8.48 \\ \hline
Denton and Fergus~\cite{denton2018stochastic} & Unimodal - V  & 0.7353 &	0.5115 &	0.4991 &	12.25 &	7.13 &	7.00 \\
Audio Only  & Unimodal - A  & 0.6474 &	0.5397 &	0.5315 &	12.39 &	\textcolor{blue}{9.25} &	\textcolor{blue}{8.84} \\ \hline
\end{tabular}
\end{table}

\begin{table*}[t]
\footnotesize 
\centering
\caption{SSIM, PSNR for AudioSet, YouTube Painting.  \textbf{Highest}, \textcolor{blue}{Second} highest scores. }\label{tab:drums_paint} 
\begin{tabular}{l|c|c|c|c||c|c|c}
\hline \hline
\multicolumn{8}{c}{\textit{Experiments on the AudioSet Dataset~\cite{gemmeke2017audio}, with 15 seen frames}} \\ \hline
\multirow{1}{*}{\textbf{Method}} & \multirow{1}{*}{\textbf{Type}} & \multicolumn{3}{c}{\textbf{SSIM}}  & \multicolumn{3}{c}{\textbf{PSNR}} \\ \cline{3-8} 
& & Frame 16 & Frame 30 &	Frame 45 & Frame 16 &	Frame 30 &	Frame 45 \\ \hline
Our Method & Multimodal & \textbf{0.9843}	& \textbf{0.9544} &	\textbf{0.9466} &	\textbf{33.24} &	\textbf{27.94} &	\textbf{26.99} \\ \hline
Multiple Frames - ~\cite{vougioukas2018end} & Multimodal  & 0.9398 &	\textcolor{blue}{0.9037} &	\textcolor{blue}{0.8959} &	26.21 &	\textcolor{blue}{23.78} &	\textcolor{blue}{23.29} \\
Vougioukas \etal ~\cite{vougioukas2018end} & Multimodal  & 0.8986 &	0.8905 &	0.8866 &	23.62 &	23.14 &	22.91 \\ \hline
Denton and Fergus~\cite{denton2018stochastic} & Unimodal - V  & \textcolor{blue}{0.9706} &	0.6606 &	0.5097 &	\textcolor{blue}{30.01} &	16.57 &	13.49 \\
Hsieh \etal~\cite{hsieh2018learning}  & Unimodal - V  & 0.1547 &	0.1476 &	0.1475 &	9.42 &	9.54 &	9.53 \\
Audio Only  & Unimodal - A  & 0.6485 &	0.6954 &	0.7277 &	18.81 &	19.79 &	20.50 \\ \hline
\multicolumn{8}{c}{\textit{Experiments on the novel YouTube Painting Dataset, with 15 seen frames}} \\ \hline
& & Frame 16 & Frame 30 &	Frame 45 & Frame 16 &	Frame 30 &	Frame 45 \\ \hline
Our Method & Multimodal & \textcolor{blue}{0.9716}	& \textbf{0.9291} &	\textbf{0.9110} &	\textbf{32.73} &	\textbf{27.27} &	\textbf{25.57} \\ 
Ours - No AA  & Multimodal  & 0.9588 &	\textcolor{blue}{0.9244} &	\textcolor{blue}{0.9076} &	30.32 &	\textcolor{blue}{26.58} &	24.75 \\
Ours - No AA, MD  & Multimodal  & 0.9552 &	0.9196 &	0.9022 &	29.45 &	26.04 &	24.75 \\
Ours - No audio  & Unimodal - V  & 0.9524 &	0.9166 &	0.8986 &	29.24 &	25.90 &	24.59 \\ \hline
Multiple Frames - ~\cite{vougioukas2018end} & Multimodal  & 0.9657 &	0.9147 &	0.8954 &	30.09 &	25.40 &	24.08 \\
Vougioukas \etal ~\cite{vougioukas2018end} & Multimodal  & 0.9281 &	0.9126 &	0.9027 &	26.97 &	25.58 &	\textcolor{blue}{24.78} \\ \hline
Denton and Fergus~\cite{denton2018stochastic} & Unimodal - V  & \textbf{0.9779} &	0.6654 &	0.4193 &	\textcolor{blue}{32.52} &	16.05 &	11.84 \\
Hsieh \etal~\cite{hsieh2018learning}  & Unimodal - V  & 0.1670 &	0.1613 &	0.1618 &	9.11 &	9.57 &	9.72 \\
Audio Only  & Unimodal - A  & 0.5997 &	0.6462 &	0.6743 &	16.75 &	17.53 &	18.04 \\ 
\hline \hline
\end{tabular}
\end{table*}
 
\begin{figure}[t]
    \centering
    \subfigure[]{\includegraphics[scale=0.28]{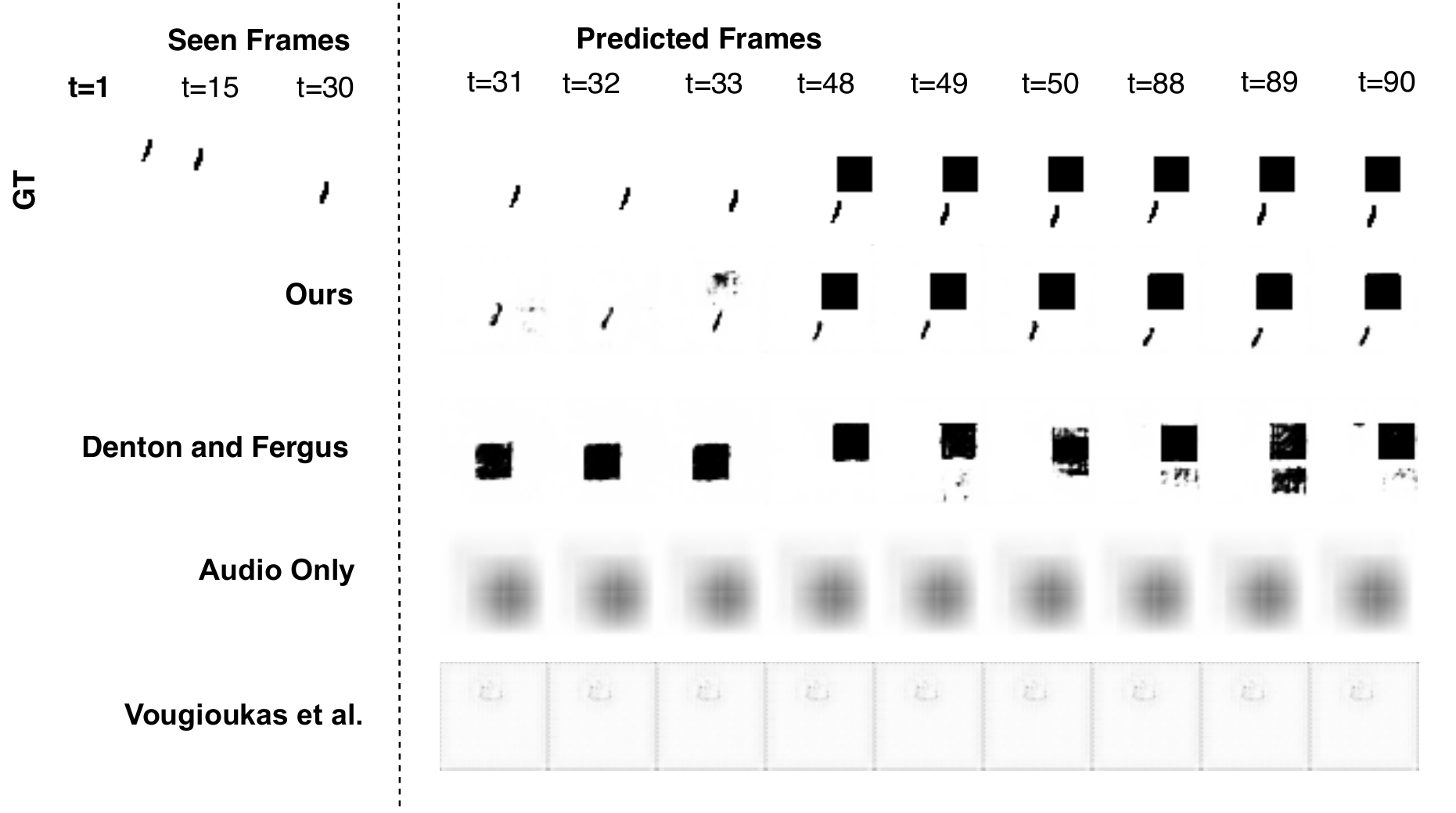}}
    \subfigure[]{\label{fig:samp_div}\includegraphics[scale=0.28]{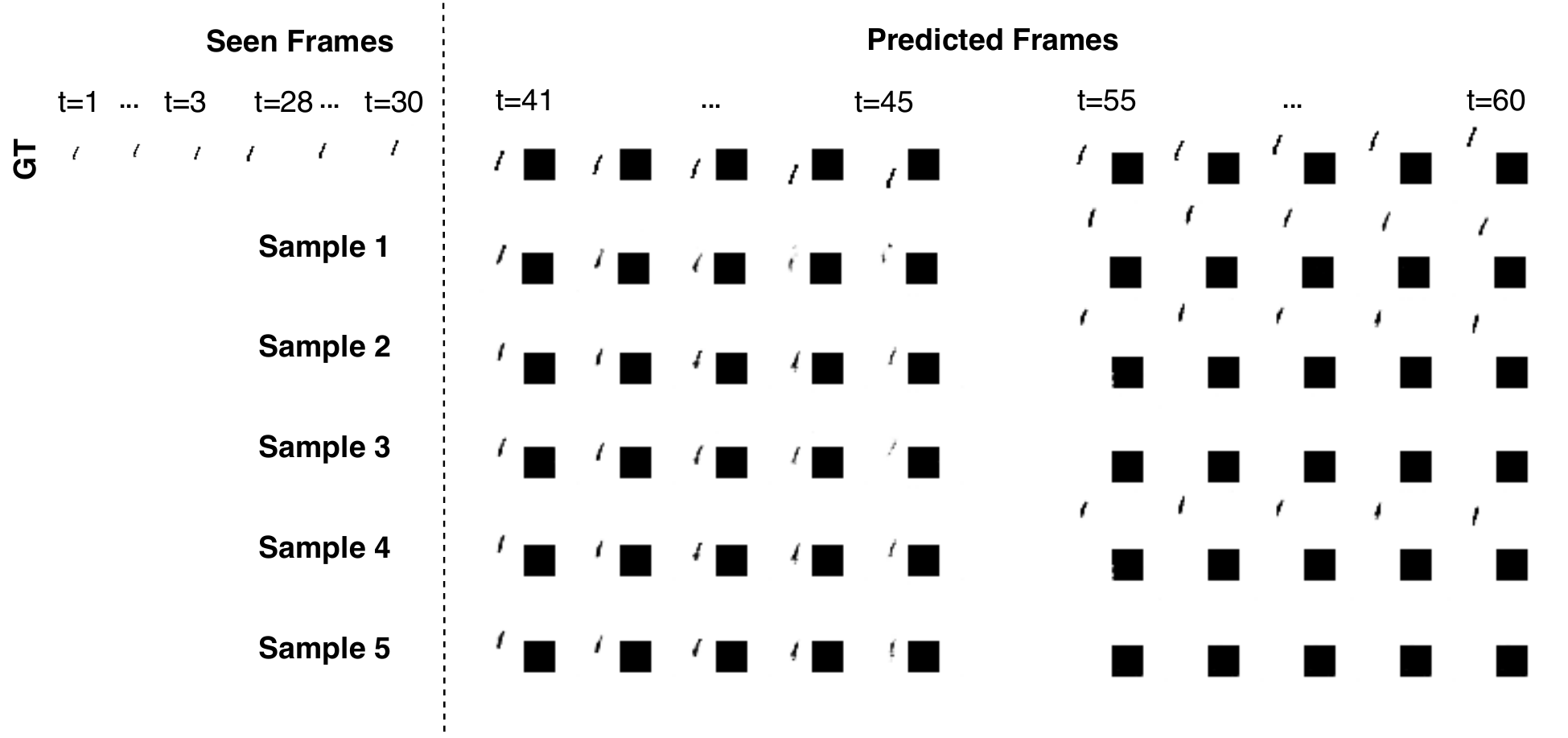}}
    \caption{Generated M3SO digits against baselines (left) and diversity (right).}
    \label{fig:m3so-quals}
\end{figure}

\noindent \textbf{M3SO Results:} Table~\ref{tab:mnist} shows the performance of our model versus competing baselines on the M3SO dataset in two settings: (i) without block (M3SO-NB) and (ii) with block (M3SO). For M3SO-NB, we observe that our method attains significant improvements over prior works, even on long-range generation. In M3SO, when the block is introduced at the $42$-nd frame, the generated frame quality drops across all methods. Nevertheless, our method continues to demonstrate better performance. Figure~\ref{fig:m3so-quals} (left)  presents a visualization of the generated frames by our method vis-\'a-vis prior works on the M3SO dataset. Contrasting the output by our method against prior works clearly reveals the superior generation quality of our method, which closely resembles the ground truth. The figure reveals that among existing approaches, Denton and Fergus~\cite{denton2018stochastic} comes closest to generating frames that look realistic, however they fail to capture the appearance of the digit. We find that the method of~\cite{denton2018stochastic} fares well under uncertainty, however our task demands reasoning over audio - an element missing in their setup. Further note that our model localizes the block in time (i.e. after the $42^{nd}$ frame) better than other methods. This is quantitatively analyzed in Table~\ref{tab:mnist_30_30_block} by comparing the mean IoU of the predicted block location in the final generated frame against the ground truth;  our scheme outperforms the closest baseline~\cite{denton2018stochastic} by $\sim$30\%.

\begin{table}[t]
\footnotesize
\parbox{.45\linewidth}{
\centering
\caption{Block localization on M3SO.}\label{tab:mnist_30_30_block}
\begin{tabular}{l|c}
\hline \hline
\textbf{Method} & \textbf{Localization IoU} \\ 
\hline \hline
Our Method & \textbf{0.5801} \\ \hline
Denton and Fergus~\cite{denton2018stochastic} & 0.2577 \\
 \hline
Vougioukas \etal ~\cite{vougioukas2018end} & 0.1289 \\ \hline 
\end{tabular}
}
\quad
\parbox{0.5\linewidth}{
\centering
\caption{Human preference score on samples from our method vs.~\cite{vougioukas2018end}}\label{tab:human_eval}
\begin{tabular}{l|c}
\hline \hline
\textbf{Datasets} & \textbf{ Prefer ours} \\ 
\hline \hline
AudioSet- Ours vs.~\cite{vougioukas2018end}  & \textbf{83\%} \\ \hline
YouTube Painting-Ours vs.~\cite{vougioukas2018end} &\textbf{92\%} \\ \hline
\end{tabular}
}
\end{table}

\begin{figure}[t]
    \begin{minipage}[c]{0.5\linewidth}
    \hspace{-0.25cm}
    \includegraphics[scale=0.33]{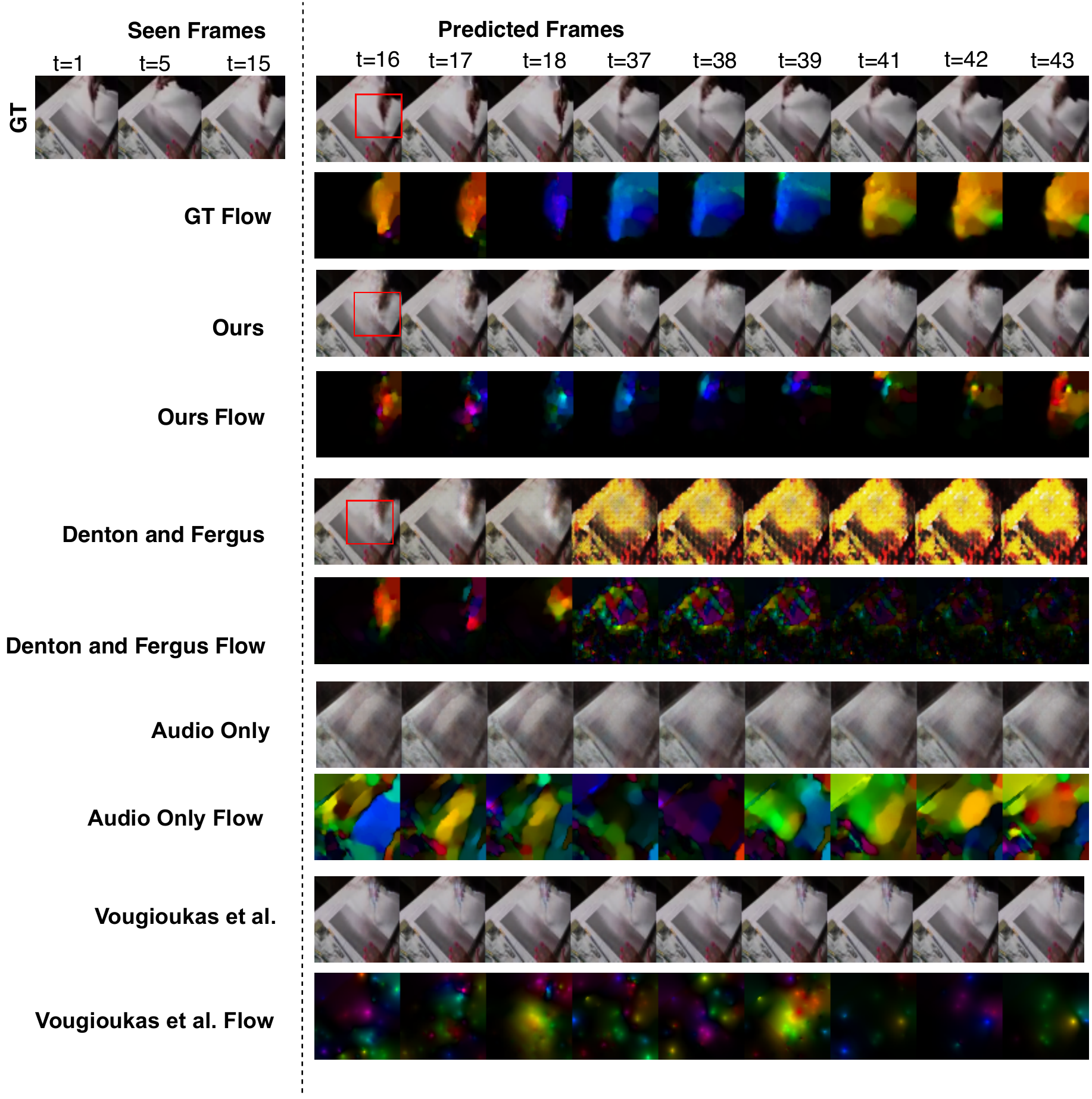}
    \end{minipage}
    \begin{minipage}[c]{0.45\linewidth}
    \includegraphics[scale=0.33]{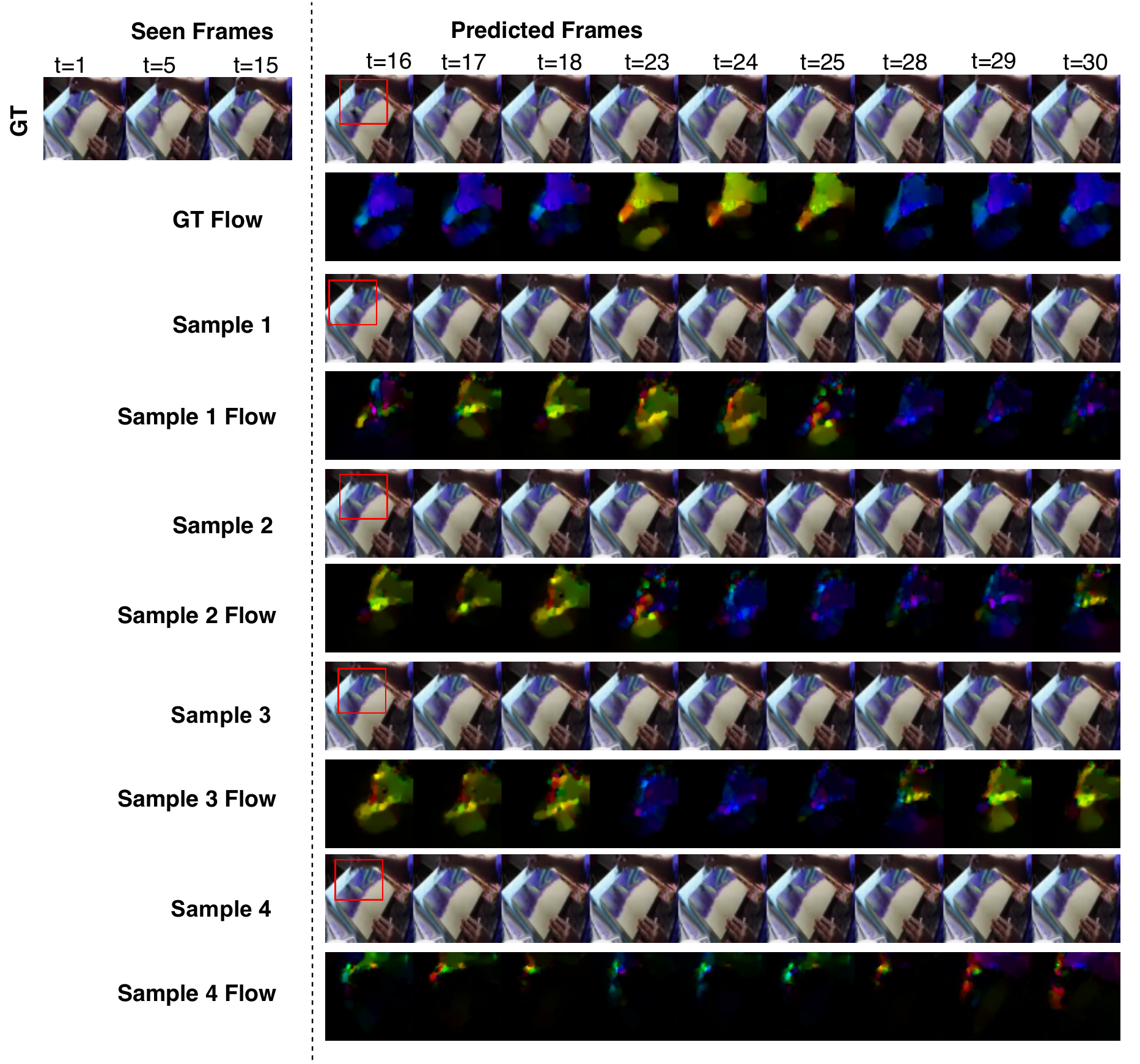}
    \end{minipage}
   \caption{Qualitative comparisons of generated frames and optical flow images against baselines for YouTube Painting (left). We show diverse generations on the right.}
    \label{fig:smmnist_audioset_qual}
\end{figure}

\noindent\textbf{Comparisons on Real-world Datasets:} As with M3SO, we see from Table~\ref{tab:drums_paint} that our approach outperforms the baselines, even at long-range generation. Due to the similarity in visual content (e.g., background) of the unseen frames to the seen frames, prior methods (e.g., ~\cite{vougioukas2018end} and~\cite{denton2018stochastic}) are seen to copy the seen frames as \emph{predicted} ones, yielding relatively high SSIM/PSNR early on (Figures~\ref{fig:first_samplel} and~\ref{fig:smmnist_audioset_qual} that show that drummer's and painter's arms remain fixed); however their performances drop in the long-range. Instead, our method captures the hand motions. Further, our generations are free from artifacts, as corroborated by the fooling rate on the fully-trained discriminator, that achieves $79.26\%$ for AudioSet Drums and $65.99\%$ for YouTube Painting. 

\noindent\textbf{Human Preference Scores:} To subjectively assess the video generation quality, we conducted a human preference evaluation 
between a randomly selected subset  of our generated videos and those produced by the closest competitor- Vougioukas \etal ~\cite{vougioukas2018end} on both the real-world datasets. The results in Table~\ref{tab:human_eval} evince that humans preferred our method for more than 80-90\% of the videos against those from~\cite{vougioukas2018end}.
\begin{figure}[!h]
    \centering
    \subfigure[Ablation Study]{\label{fig:smmnist_ssim_ablation}\includegraphics[width=4.0cm,trim={0cm 6.3cm 1.7cm 2cm},clip]{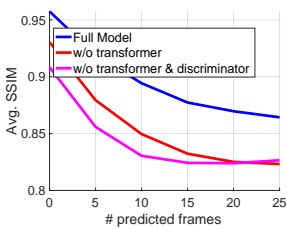}} 
    \subfigure[Inter-SSIM]{\label{fig:mnist_div_ssim}\includegraphics[width=4.0cm,trim={0cm 6.3cm 1.7cm 2cm},clip]{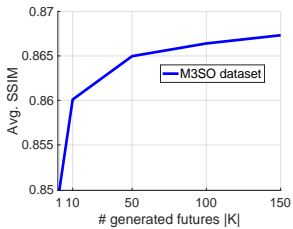}}
    \subfigure[Intra-SSIM]{\label{fig:realwrld_div_ssim}\includegraphics[width=4.0cm,trim={0cm 6.3cm 1.7cm 2cm},clip]{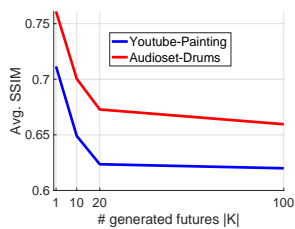}}
   \caption{Ablation and diversity studies (see text for details).} 
\end{figure}

\noindent\textbf{Sample Diversity:} In Figure~\ref{fig:samp_div}, we show the diversity in the samples generated on the M3SO dataset. We see from this figure that the trajectory of the digit (`1') changes across generated sequences, each plausible. Figure~\ref{fig:mnist_div_ssim} shows quantitative evaluation of diversity. Specifically, we generated a set of $\mathcal{K}$ futures at every time step (for $|\mathcal{K}|$ ranging from $1-100$), and plotted the SSIM of the samples which matched maximally with the ground truth. As is clear, this plot shows an increasing trend suggesting that samples closer to the ground-truth are obtainable by increasing $\mathcal{K}$; suggesting generative diversity. We further analyze this over SSIMs on optical flows computed from the Youtube Painting and Drums datasets. In Figure~\ref{fig:realwrld_div_ssim}, we plot the intra-sample diversity, i.e., the average pairwise SSIMs for sequences in $\mathcal{K}$; showing a downward trend, suggesting these sequences are self-dissimilar within $\mathcal{K}$. 

\noindent\textbf{Ablation Results:} To study the influence of the transformer network, we contrast our model by substituting the transformer by an LSTM with 128-D hidden states. Figure~\ref{fig:smmnist_ssim_ablation} shows the result, clearly suggesting the benefits of using transformers. From this plot, we also find that having our discriminator is important. Tables~\ref{tab:mnist} and~\ref{tab:drums_paint} show that removing the AA and MD loss terms from the discriminator  hurts performance.

\section{Conclusions}

In this work, we explored the novel task of video generation from audio and the visual context for generic videos. We proposed a novel deep variational encoder-decoder model for this task, that also characterizes the underlying stochasticity in real-world videos. We combined our video generator with a multimodal discriminator to improve its quality and diversity. Empirical evaluations on three datasets demonstrated the superiority of our method over closely-related baselines.

\vspace{4pt}
\noindent\textbf{Acknowledgements.} MC thanks the support from the Joan and Lalit Bahl Fellowship. 
%
%
\bibliographystyle{splncs04}
\bibliography{sound2sight_bib}


\appendix
\section*{Appendix}
In this supplementary materials, we include the following:
\begin{enumerate}
    \item Additional details about the three datasets and the associated tasks in Sections~\ref{sec:m3so}, ~\ref{sec:movingmnist}, ~\ref{sec:drums}, and ~\ref{sec:painting} respectively. 
    \item The architecture of \emph{Sound2Sight} and details of the training procedure.
    \item Standard Deviation measures of the model performance.
    \item Auxiliary evaluation of the quality of the generated videos.
    \item Performances from ablative studies of our model, the effect of the choice of hyper-parameters, per-sample comparisons with competing methods, and a study of the effectiveness of teacher forcing training strategy. Also included are plots showcasing the diversity of our model. Further, the alignment of the generated frames against the input audio is quantitatively evaluated.
    \item Qualitative experimental results vis-\'a-vis competitive baselines and figures illustrating the diversity of the samples generated by our model.
    \item Failure cases of our model.
\end{enumerate}


\section{Datasets and Tasks}
As described in the main paper, we present results on three Audio-Visual datasets for our video generation task, namely (i) the Multimodal Moving MNIST with (and without) a Surprise Obstacle (M3SO), and its variant without the obstacle (M3SO-NB), (ii) the Audioset-Drums~\cite{gemmeke2017audio}, and (iii) YouTube-Painting. Here, we provide more details of these tasks. \textbf{Please see the supplementary video samples for a better understanding of the task}. Figure~\ref{fig:dataset_samples} shows samples from all three datasets and Table~\ref{tab:dataset_stat} presents the statistics of these datasets and the training/val/test splits.

\subsection{Multimodal Moving MNIST with a Surprise Obstacle (M3SO)}
\label{sec:m3so}
 This is a novel synthetic dataset, where a randomly chosen MNIST digit~\cite{lecun1998gradient} moves in a box (of size $48 \times 48$), along rectilinear trajectories, in the seen part of the sequence (30 frames for example). While doing so, the digit occasionally bounces against the walls of the box which alters its trajectory of motion randomly. The digit then traverses this path of motion linearly till the next collision happens. Additionally, we associate audio with each digit (a unique tone for the digit), and the amplitude of this tone is inversely proportional to the distance of the moving digit from the lower-left corner of the box. When the digit bounces off the box edges, a different tone is emanated (i.e. the frequency of the audio changes). The audio then reverts to the original tone of the digit as it continues its motion. This process sustains as we transition from the ``seen'' frames to the ensuing ``unseen'' part. 
 
 At a pre-selected frame in the ``unseen'' part (the $42^{nd}$ frame, i.e. the $12^{th}$ unseen frame) , a square obstacle is introduced in the box at a randomly chosen spatial location. When the digit bounces off this obstacle yet another different but unique tone is played (i.e. the frequency changes again). Post the collision, the digit continues its linear motion in a random direction, with its accompanying tone switching back to the original tone of the digit. 
 
\begin{figure*}[t]
    \centering
    \includegraphics[width=12cm]{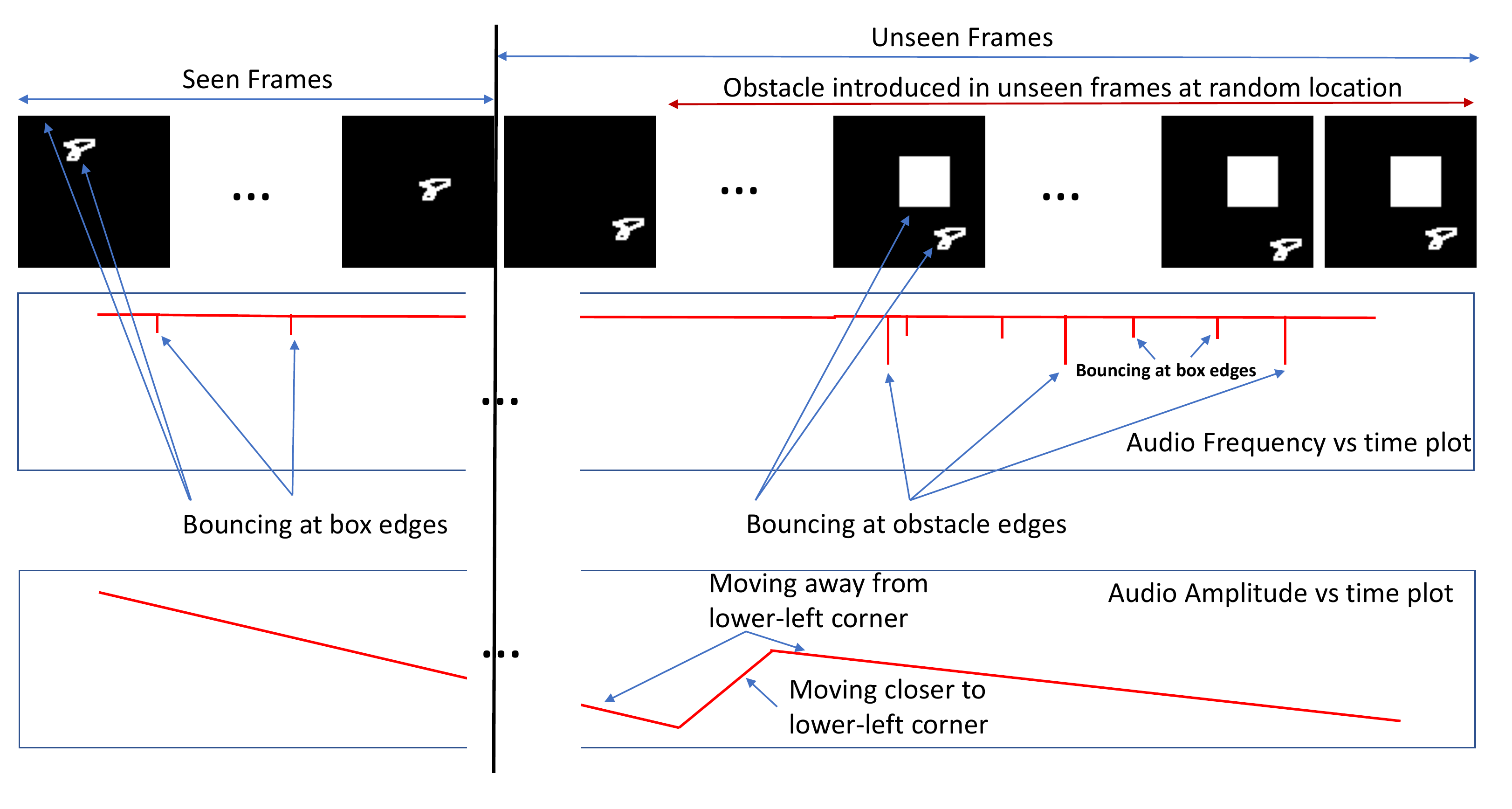}
    \caption{An illustration of our proposed ``Multimodal Moving MNIST with a Surprise Obstacle'' (M3SO) dataset. The top row shows a few salient frames in a sample video. The two rows below it show the frequency (middle row) and amplitude (bottom row) of the audio signal plotted against time. When the digit bounces on the edges, it has a momemtary tone change, seen on the left part of the middle row, and when the digit bounces against the block, there is a different tone played mometarily, as is shown by the first spike on the right side of the middle row. The last row shows the change in amplitude of the tone as the digit moves closer/farther away from the bottom left corner of the canvas.} 
    \label{fig:mnist}
\end{figure*}

 Given this setting, our task is to use sound to generate the unseen frames which involves capturing the dynamics of the digit, and also placing the obstacle accurately both in time and space. For our experiments, we train all algorithms with 30 seen frames and the task is to predict the next 30 for training. Additionally, in order to evaluate the generalizability of the model into the distant future, we ask all methods to predict 60 unseen frames at test time. The obstacle is introduced in the $42^{nd}$ frame, i.e. the $12^{th}$ unseen frame.  Figure~\ref{fig:mnist} represents this setup visually. Figure~\ref{fig:dataset_samples} shows some sample frames from a clip in this dataset.

\subsection{M3SO-NB}
\label{sec:movingmnist}
We also conduct experiments on the proposed Multimodal Moving MNIST dataset without the surprise obstacle component (M3SO-NB). In this setting, we train with 5 seen frames and predict the next 15 frames. However at test time all competing models predict 25 frames. 

\subsection{Audioset-Drums}
\label{sec:drums}
This dataset was constructed by collecting videos from the \textit{Drums} class of the AudioSet dataset~\cite{gemmeke2017audio}. This particular class of videos is unique in the sense that most of the videos in this class have correlated visual and auditory information. We selected those videos from this class which clearly had some body-parts (mostly hands and head) of the drummer visible - playing his (or her) drum kit in an indoor environment. We removed videos that had animations, and those clips in which the sound source (i.e., the drum kit) was not clearly visible.  All video clips were resized to a frame resolution of $64 \times 64$ at 30fps and the audio was sampled at 44kHz. Figure~\ref{fig:dataset_samples} shows some sample frames from this dataset. For this dataset, we train all competing techniques with 15 seen frames and predict the next 15. At test time however, we predict further into the future by predicting the next 30 frames after the 15 seen frames.

\subsection{YouTube-Painting}
\label{sec:painting}
Apart from the constrained or simplified audio-visual prediction context, as captured in the previous two datasets, we decided to introduce a more challenging dataset, that is still constrained in its context, however is diverse and loose in its spatio-temporal dynamics. After looking at a tradeoff between various possibilities on a large collection of Youtube videos, including diversity in camera view angles, types of actor motions, kinds of visual context, and a clear and distinctive audio cue, we decided to use videos containing an actor painting some art. We realized that there exists a good collection of such videos on Youtube, which can provide a good training set for our scheme. While, this dataset may not be characteristic of large motions, it contains videos taken from multiple viewing angles, periodic motions (while painting), diversity in the visual context (such as different mixes of colors, paint brushes, etc.), different painters, and different orientations and types of painting canvas, brush, etc at the same time containing clear sound of the painter's brush touching the canvas. With this idea, we present our Youtube-Paining dataset.

\begin{figure*}[]
    \begin{center}
    \includegraphics[width=1.0\linewidth]{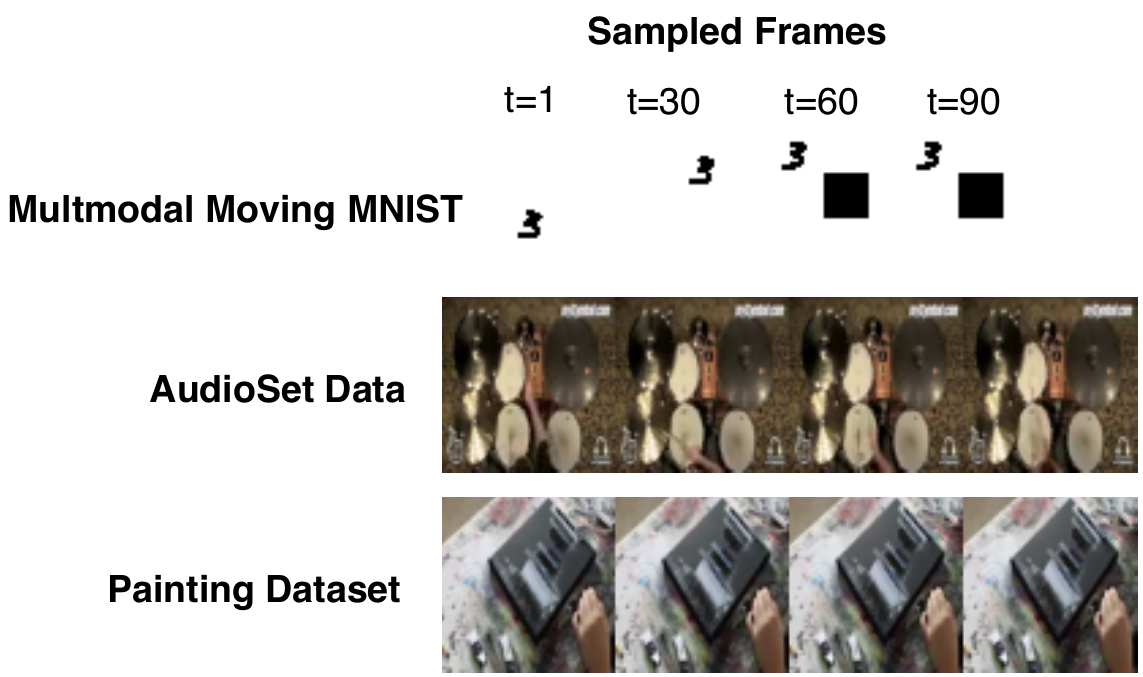}
    \end{center}
   \caption{Sample frames from the three datasets that we used in our experiments.}
    \label{fig:dataset_samples}
\end{figure*}

The YouTube-Painting dataset was constructed by collecting videos from YouTube where a painter (or a part of his/her body) is seen painting an acrylic painting, inside a room~\cite{Youtube:Taylor}. We used ``Taylor ASMR'' as the search query to crawl these videos. Besides the visuals of the painter painting, there is accompanying audio emanating as a result of the painter's brush movements upon the canvas. The videos do not feature too much camera motion and thus the change in audio frequency is a cue for the distance of the brush from the camera. Akin to the AudioSet-Drums dataset, the chosen videos in this dataset also do not contain animations, shot changes, etc. Further, videos for which the sound source, i.e., the brush, was not clearly seen, were dropped from the dataset. Video clips in this dataset have frames of size $64 \times 64$ at 30fps, while the audio is sampled at 44kHz. Figure~\ref{fig:dataset_samples} shows sample frames from this dataset. Here too, we train our models with 15 seen frames and predict the next 15. At test time however, we predict further into the future by predicting the next 30 frames after having seen the first 15 frames.

\begin{table}[t]
\centering
\caption{A summary of the statistics of the different datasets.}\label{tab:dataset_stat}
\begin{tabular}{l|c|c|c|c}
\hline \hline
\textbf{Datasets} & \textbf{\# Train} & \textbf{\# Test} & \textbf{\# Val} & \textbf{\# Frames/video} \\ 
\hline \hline
M3SO  &  8000 & 1000 & 1000 & 100 \\ \hline
AudioSet-Data &  6000 & 1000 & 1000 & 90 \\ \hline
YouTube-Painting &  4800 & 500 & 500 & 90  \\ \hline
\end{tabular}
\end{table}

\section{Network Architecture and Training Details}
We use an LSTM with 2 layers in the prediction module, the input to which is of 138 dimensions (128 dimensions of features and 10 dimensional $z_t$). The prior and posterior LSTMs are both single-layered. All LSTMs have a hidden state size of 256 dimensions. Each transformer module has one layer and four heads for capturing multi-head self-attention. The discriminator uses an LSTM with a single hidden layer of 256 dimensions, and a frame-history $R=2$ and a look-ahead window of size $k=1$. We train the generator and discriminator jointly with an initial learning rate of 0.002 for both, using the ADAM~\cite{kingma2014adam} optimizer. During inference, we sample 100 futures per time step, and use the one that maximally matches the ground-truth for evaluating our method. We use the same evaluation for all baseline methods that can generate multiple plausible futures. The weighting term on the KL-loss, $\beta$, and the weight on the discriminator loss, $\gamma$, were both set to $0.0001$ for all datasets. However, $\gamma$ was increased by a factor of 10 every 300 training epochs. All hyper-parameters were chosen using cross-validation on the validation category of every dataset.

\section{Standard Deviation Measures of Model Performance}

\begin{table}[t]
\footnotesize
\centering
\caption{Standard Deviation Scores of SSIM, PSNR  on the test set of M3SO-NB and M3SO Datasets.} 
\label{tab:std_mnist}
\begin{tabular}{l|c|c|c|c||c|c|c}
\hline \hline
\multicolumn{8}{c}{\textit{Experiments with M3SO-NB with 5 seen frames}} \\ \hline
\multirow{1}{*}{\textbf{Method}} & \multirow{1}{*}{\textbf{Type}} & \multicolumn{3}{c}{\textbf{SSIM}} & \multicolumn{3}{c}{\textbf{PSNR}} \\ \cline{3-8}
& & Frame 6 & Frame 15 &	Frame 25 & Frame 6 &	Frame 15 &	Frame 25 \\ \hline
Our Method & Multimodal & 0.0011	& 0.0051 &	0.0037 &	0.157 &	0.197 &	0.142 \\ \hline
Multiple Frames - ~\cite{vougioukas2018end} & Multimodal  & 0.0008 &	0.0016 &	0.0006 &	0.032 &	0.100 &	0.112 \\
Vougioukas \etal ~\cite{vougioukas2018end} & Multimodal  & 0.0169 &	0.0175 &	0.0169 &	0.069 &	0.120 &	0.121 \\ 
 \hline
Denton and Fergus~\cite{denton2018stochastic} & Unimodal - V  & 0.0009 &	0.0021 &	0.0014 &	0.090 &	0.064 &	0.057 \\
Audio Only  & Unimodal - A  & 0.0045 &	0.0036 &	0.0034 &	0.120 &	0.131 &	0.102 \\ \hline
\multicolumn{8}{c}{\textit{Experiments on M3SO with 30 seen frames (Block is introduced in the $42^{nd}$ frame)}} \\ \hline
& & Frame 31 & Frame 42 &	Frame 70 & Frame 31 &	Frame 42 &	Frame 70 \\ \hline
Our Method & Multimodal & 0.0050	& 0.0064 &	0.0064 &	0.178 &	0.127 &	0.144 \\ \hline
Multiple Frames - ~\cite{vougioukas2018end} & Multimodal  & 0.0011 &	0.0029 &	0.0032 &	0.142 &	0.038 &	0.027 \\
Vougioukas \etal ~\cite{vougioukas2018end} & Multimodal  & 0.0008 &	0.0024 &	0.0014 &	0.076 &	0.002 &	0.018 \\ \hline
Denton and Fergus~\cite{denton2018stochastic} & Unimodal - V  & 0.0106 &	0.0226 &	0.0193 &	0.417 &	0.182 &	0.147 \\
Audio Only  & Unimodal - A  & 0.0123 &	0.0178 &	0.0179 &	0.307 &	0.306 &	0.307 \\ \hline
\end{tabular}
\end{table}


\begin{table*}[t]
\footnotesize 
\centering
\caption{Standard Deviation Scores of SSIM, PSNR  on the test set of AudioSet, YouTube Painting Datasets.}\label{tab:std_drums_paint} 
\begin{tabular}{l|c|c|c|c||c|c|c}
\hline \hline
\multicolumn{8}{c}{\textit{Experiments on the AudioSet Dataset~\cite{gemmeke2017audio}, with 15 seen frames}} \\ \hline
\multirow{1}{*}{\textbf{Method}} & \multirow{1}{*}{\textbf{Type}} & \multicolumn{3}{c}{\textbf{SSIM}}  & \multicolumn{3}{c}{\textbf{PSNR}} \\ \cline{3-8} 
& & Frame 16 & Frame 30 &	Frame 45 & Frame 16 &	Frame 30 &	Frame 45 \\ \hline
Our Method & Multimodal & 0.0092	& 0.0065 &	0.0065 &	0.123 &	0.519 &	0.266 \\ \hline
Multiple Frames - ~\cite{vougioukas2018end} & Multimodal  & 0.0168 &	0.0073 &	0.0148 &	0.964 &	0.231 &	0.320 \\
Vougioukas \etal ~\cite{vougioukas2018end} & Multimodal  & 0.0162 &	0.0205 &	0.2650 &	0.497 &	0.402 &	0.182 \\ \hline
Denton and Fergus~\cite{denton2018stochastic} & Unimodal - V  & 0.0168 &	0.0102 &	0.0084 &	1.098 &	0.319 &	0.054 \\
Hsieh \etal~\cite{hsieh2018learning}  & Unimodal - V  & 0.0050 &	0.0016 &	0.0082 &	0.042 &	0.051 &	0.006 \\
Audio Only  & Unimodal - A  & 0.0197 &	0.0068 &	0.0072 &	0.230 &	0.388 &	0.177 \\
 \hline
\multicolumn{8}{c}{\textit{Experiments on the novel YouTube Painting Dataset, with 15 seen frames}} \\ \hline
& & Frame 16 & Frame 30 &	Frame 45 & Frame 16 &	Frame 30 &	Frame 45 \\ \hline
Our Method & Multimodal & 0.0025	& 0.0093 &	0.0146 &	1.369 &	0.718 &	0.250 \\ \hline
Multiple Frames - ~\cite{vougioukas2018end} & Multimodal  & 0.0020 &	0.0040 &	0.0050 &	0.181 &	0.536 &	0.879 \\
Vougioukas \etal ~\cite{vougioukas2018end} & Multimodal  & 0.0143 &	0.0028 &	0.0150 &	0.212 &	0.216 &	0.449 \\ \hline
Denton and Fergus~\cite{denton2018stochastic} & Unimodal - V  & 0.0008 &	0.0193 &	0.0390 &	0.431 &	0.338 &	0.459 \\
Hsieh \etal~\cite{hsieh2018learning}  & Unimodal - V  & 0.0028 &	0.0019 &	0.0030 &	0.033 &	0.133 &	0.150 \\
Audio Only  & Unimodal - A  & 0.0115 &	0.0069 &	0.0226 &	0.353 &	0.294 &	0.426 \\ 
\hline \hline
\end{tabular}
\end{table*}

Tables~\ref{tab:std_mnist},~\ref{tab:std_drums_paint} present the standard deviation of SSIM and PSNR scores on the test set of M3SO-NB, M3SO, AudioSet-Drums, and YouTube Painting datasets. The low standard deviation scores on both SSIM and PSNR, across all datasets, underscore the gains of our method over competing methods. For the mean test set SSIM and PSNR scores on these datasets, we refer the interested reader to Tables 1 and 2 of the main paper. 

\section{Auxiliary Evaluation of Generated Video Quality}

\begin{table*}[t]
\centering
\caption{Average discriminator fooling rates for different methods on real-world data.}\label{tab:disc_fool}
\begin{tabular}{l|c|c|c|c}
\hline \hline
\textbf{Datasets} & \textbf{Our Method} & \textbf{Denton and Fergus}~\cite{denton2018stochastic} & \textbf{Audio-Only}  & \textbf{ Multiframe }~\cite{vougioukas2018end} \\ 
\hline \hline
AudioSet Data     & \textbf{0.7926} & 0.3372 & 0.5622 & 0.6095 \\ \hline
YouTube Painting Data   & \textbf{0.6599} & 0.4282 & 0.4687 & 0.6507 \\ \hline
\end{tabular}
\end{table*}

\begin{table*}[t]
\centering
\caption{Human preference score on samples generated by our method vs.~\cite{vougioukas2018end}}\label{tab:human_eval_supp}
\begin{tabular}{l|c}
\hline \hline
\textbf{Datasets} & \textbf{ Prefer ours} \\ 
\hline \hline
M3SO- Ours vs. Multiframe~\cite{vougioukas2018end} &  \textbf{88\%} \\ \hline
AudioSet- Ours vs. Multiframe~\cite{vougioukas2018end}  & \textbf{83\%} \\ \hline
YouTube Painting-Ours vs. Multiframe~\cite{vougioukas2018end} &\textbf{92\%} \\ \hline
\end{tabular}
\end{table*}

\begin{table*}[t]
\centering
\caption{mIoU on block localization, evaluated for the final frame of the generated sequences on M3SO with block.}\label{tab:mnist_30_30_block_supp}
\begin{tabular}{l|c}
\hline \hline
\textbf{Method} & \textbf{Localization IoU} \\ 
\hline \hline
Our Method & \textbf{0.5801} \\ \hline
Denton and Fergus~\cite{denton2018stochastic} & 0.2577 \\
 \hline
 Vougioukas \etal ~\cite{vougioukas2018end} & 0.1289 \\ \hline
 Audio-Only & 0.1030 \\  \hline
 
\end{tabular}
\end{table*}
Besides evaluating our method against the baselines using the SSIM and PSNR (see Table 1 and 2 in the main paper), we use some auxiliary measures to further evaluate the quality of synthesis. Table~\ref{tab:disc_fool} presents the fooling rate of a discriminator trained to distinguish real video clips of length $R + (k-1)$ (which equals $2$ in our case) frames (and their audio) from synthetic ones, on both real world-datasets . We see that our approach outperforms all baselines, attesting to the quality of the frames that are generated by our method. Note that, the discriminator is trained to judge the audio-visual alignment as well as the quality of frames. Thus, a higher discriminator fooling rate implies that the respective model generates more real looking frames, realistic dynamics, and better alignment with audio. Additionally in Table~\ref{tab:human_eval_supp}, we present the human preference score for samples of our method versus the competitive method of Multiframe~\cite{vougioukas2018end}. \textbf{The results evince that human annotators prefer samples generated by our method overwhelmingly}. For the M3SO dataset, we also measure the intersection over union of the predicted box location against the ground truth location. Table~\ref{tab:mnist_30_30_block_supp} shows that our method outperforms competing methods by a very significant margin (nearly 30\%).  A high accuracy for this localization task, such as that of our method, demands good capture of the visual dynamics of the digits, along with synchronization against the tones corresponding to the bouncing of the digits with the obstacle. 

\section{Additional Experimental Details}
\subsection{Ablative Analysis}

\begin{table}[t]
\footnotesize
\centering
\caption{SSIM scores for our full model vis-\'a-vis different variants of our model on M3SO-NB and YouTube Painting Datasets. Highest scores are in \textbf{bold}.} 
\label{tab:ablation_mnist_painiting}
\begin{tabular}{l|c|c|c|c||c|c|c}
\hline \hline
\multirow{1}{*}{\textbf{Method}} & \multirow{1}{*}{\textbf{Type}} & \multicolumn{3}{c}{\textbf{M3SO-NB}} & \multicolumn{3}{c}{\textbf{YouTube Painting}} \\ \cline{3-8}
& & Frame 6 & Frame 15 &	Frame 25 & Frame 16 &	Frame 30 &	Frame 45 \\ \hline
Our Method & Multimodal & \textbf{0.9575}	& \textbf{0.8943} &	\textbf{0.8697} &	\textbf{0.9716} &	0.9291 &	\textbf{0.9110} \\ 
Ours - Only $L_2$  & Multimodal  & 0.9543 &	0.7624 &	0.5610 &	0.9524 &	0.9166 &	0.8986 \\ 
Teacher - Forcing  & Multimodal  & 0.9412 &	0.8819 &	0.8519 &	0.9695 &	\textbf{0.9293} &	0.9109 \\ 
AV Mismatch  & Multimodal  & 0.9428 &	0.8569 &	0.8234 &	0.9496 &	0.8784 &	0.8520 \\ \hline \hline
\end{tabular}
\end{table}

\begin{figure*}[t]
    \begin{center}
    \subfigure[SSIM]{\label{fig:mnist_ablation_ssim}\includegraphics[width=0.49\linewidth]{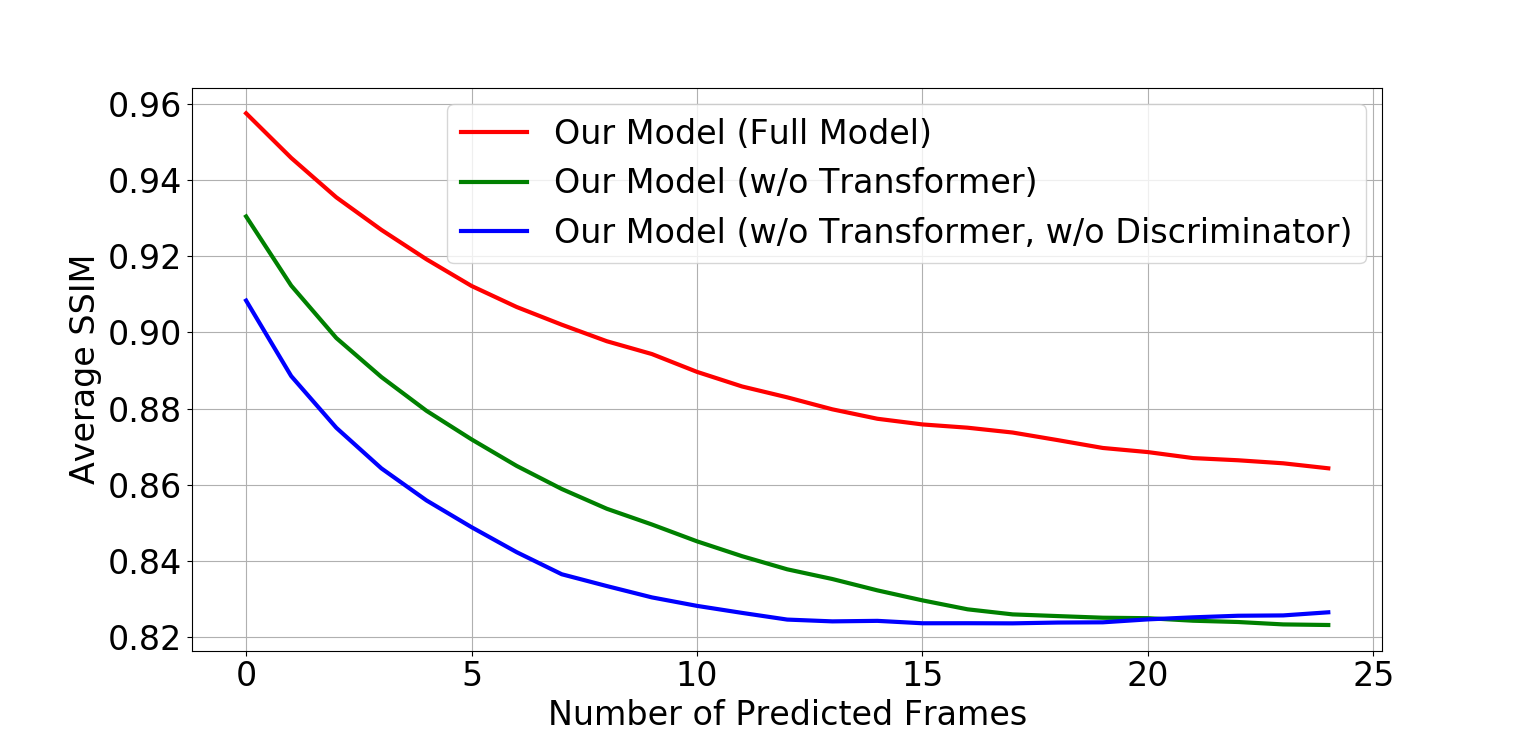}}
    \subfigure[PSNR]{\label{fig:mnist_ablation_psnr}\includegraphics[width=0.49\linewidth]{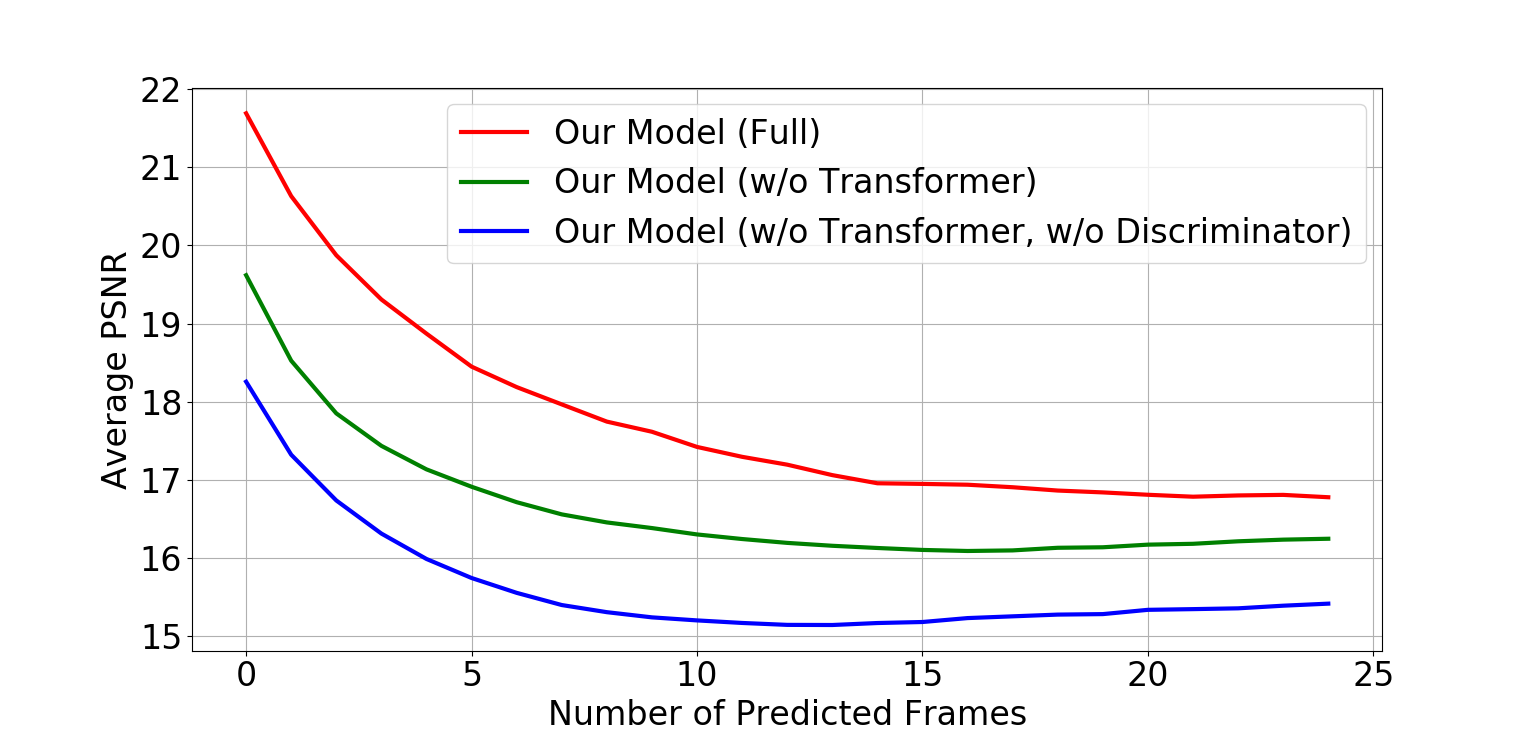}}
    \end{center}
   \caption{SSIM (left) and PSNR (right) on the M3SO-NB dataset with ablated variants of our Sound2Sight model.}
    
\end{figure*}

Figures~\ref{fig:mnist_ablation_ssim} and ~\ref{fig:mnist_ablation_psnr} show the performance variations of different ablated variants of our model. Both plots highlight the gains obtained by using: (i) transformer encoder networks~\cite{vaswani2017attention} to encode the input audio and visual modalities in the stochasticity module; and (ii) the multimodal discrimator network. The transformer encoders help to emphasize the salient components of the features, while attenuating the others by leveraging self-attention. Using the multimodal discriminator discriminator, on the other hand, ensures that the synthesized video clips are realistic and well-aligned with the audio.  

Table~\ref{tab:ablation_mnist_painiting} contrasts our full model against an ablated variant of our model, which does not have the posterior network or the multimodal discriminator (Ours - Only $L_2$). This variant is trained with only the reconstruction loss ($L_2$ loss of Equation 5 of the main paper) and serves to disentangle its effect from the other ones in the final objective (Equation 7 in the paper). As is evident from the results, merely training with the reconstruction loss leads to sub-optimal performance.

\subsection{Sensitivity of Hyperparameters}
\begin{figure*}[htbp]
    \begin{center}
    \subfigure[$\gamma$ plot]{\label{fig:mnist_ablation_gamma_ssim}\includegraphics[width=.49\linewidth]{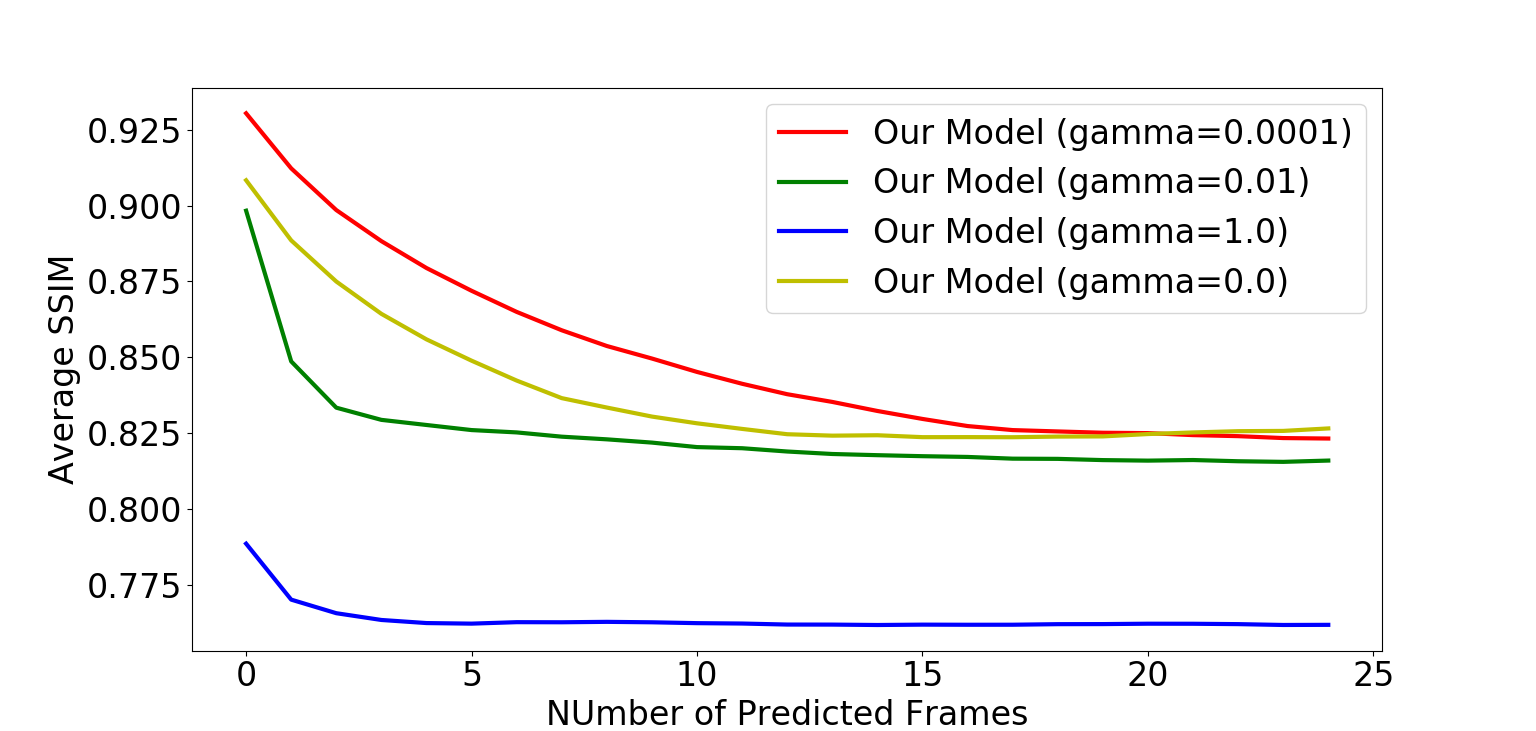}}
    \subfigure[$\beta$ plot] {\label{fig:mnist_ablation_beta_ssim}\includegraphics[width=0.49\linewidth]{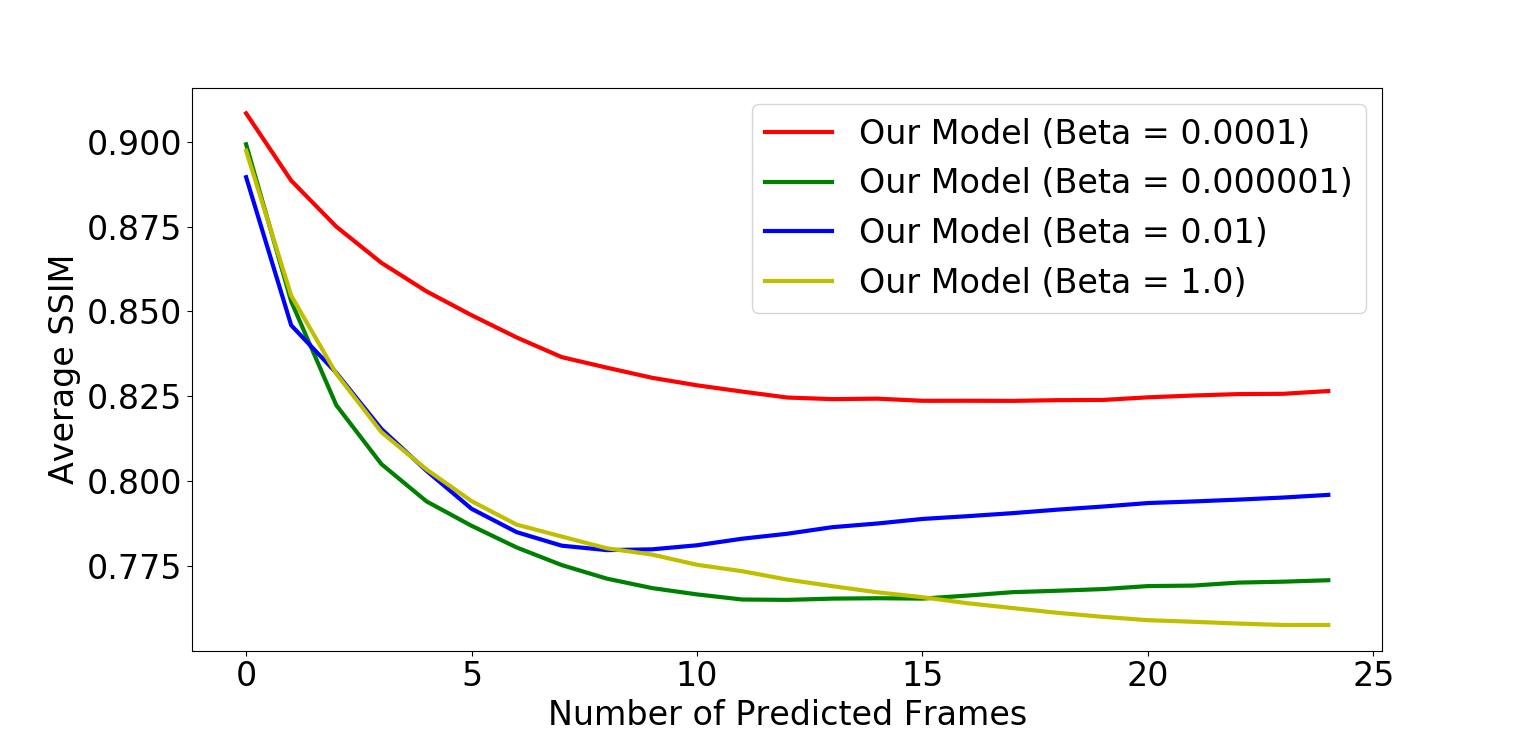}} 
    \end{center}
   \caption{Hyperparameter Study: Variations in performance on the M3SO-NB dataset against different choices of weighting on the discriminator loss, $\gamma$, as measured by SSIM (left) and different choices of weighting on the KL-Divergence loss, $\beta$, as measured by SSIM (right)}
\end{figure*}

Figures~\ref{fig:mnist_ablation_gamma_ssim} and~\ref{fig:mnist_ablation_beta_ssim}  show the empirical sensitivity analysis of how performance varies with various choices of hyper-parameters $\beta$ and $\gamma$, respectively on the M3SO-NB dataset. As can be seen from these plots, our model attains its peak performance when both these parameters are set to $0.0001$.

\subsection{Evaluating Diversity of the Prediction Network}
\begin{figure*}[t]
    \begin{center}
    \subfigure[SSIM]{\label{fig:mnist_ablation_futcand_ssim}\includegraphics[width=0.49\linewidth]{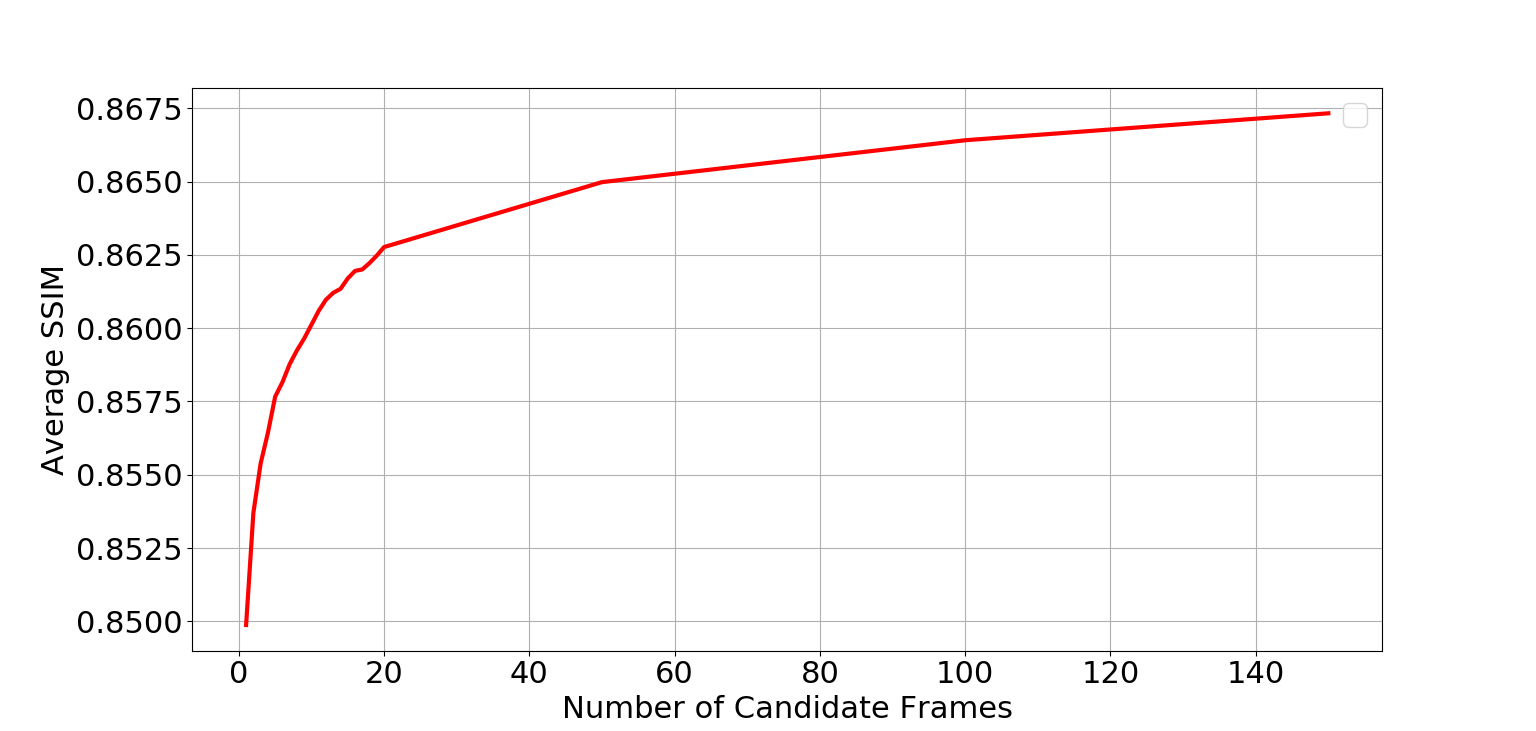}}
    \subfigure[PSNR]{\label{fig:mnist_ablation_futcand_psnr}\includegraphics[width=.49\linewidth]{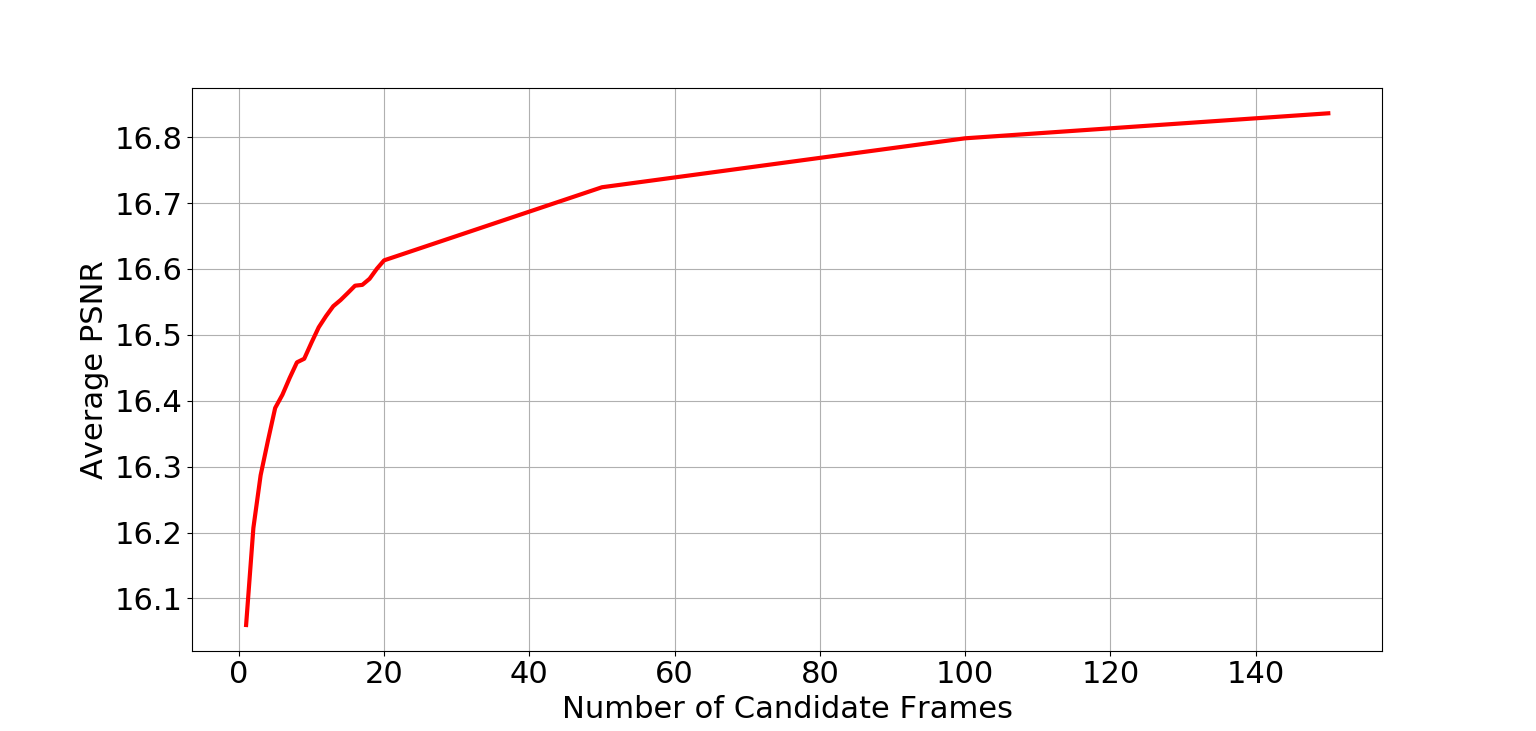}}
    \end{center}
   \caption{Variation in performance against ground truth video, on the Vanilla MovingMNIST without obstacle with increasing number of candidates (number of futures generated) at every time step, measured by SSIM (left) and PSNR (right).}
\end{figure*}

\begin{figure}
    \begin{center}
    \subfigure[AudioSet-Drums]{\label{fig:drums_futcand_ssim}\includegraphics[width=0.49\linewidth]{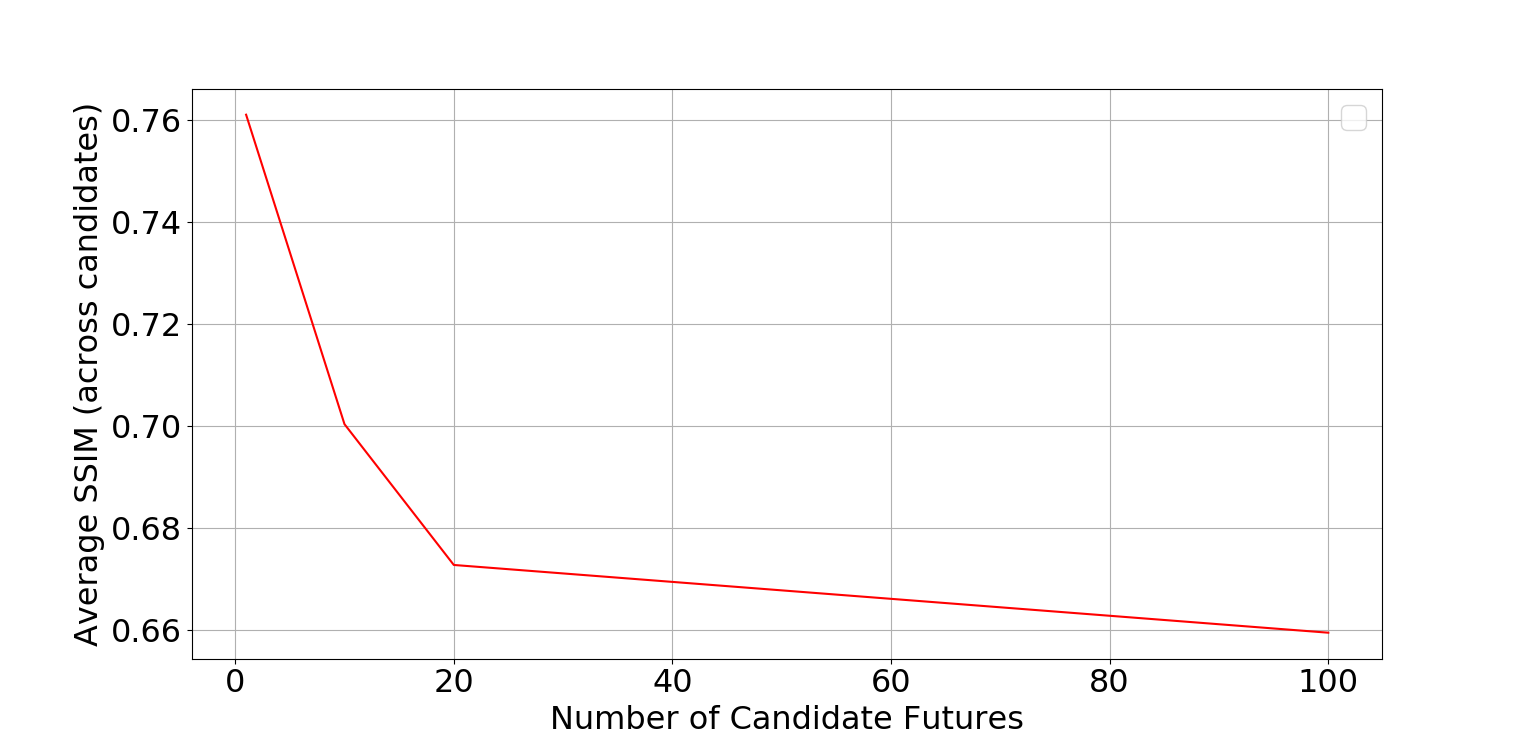}}
    \subfigure[YouTube-Painting]{\label{fig:painting_futcand_ssim}\includegraphics[width=.49\linewidth]{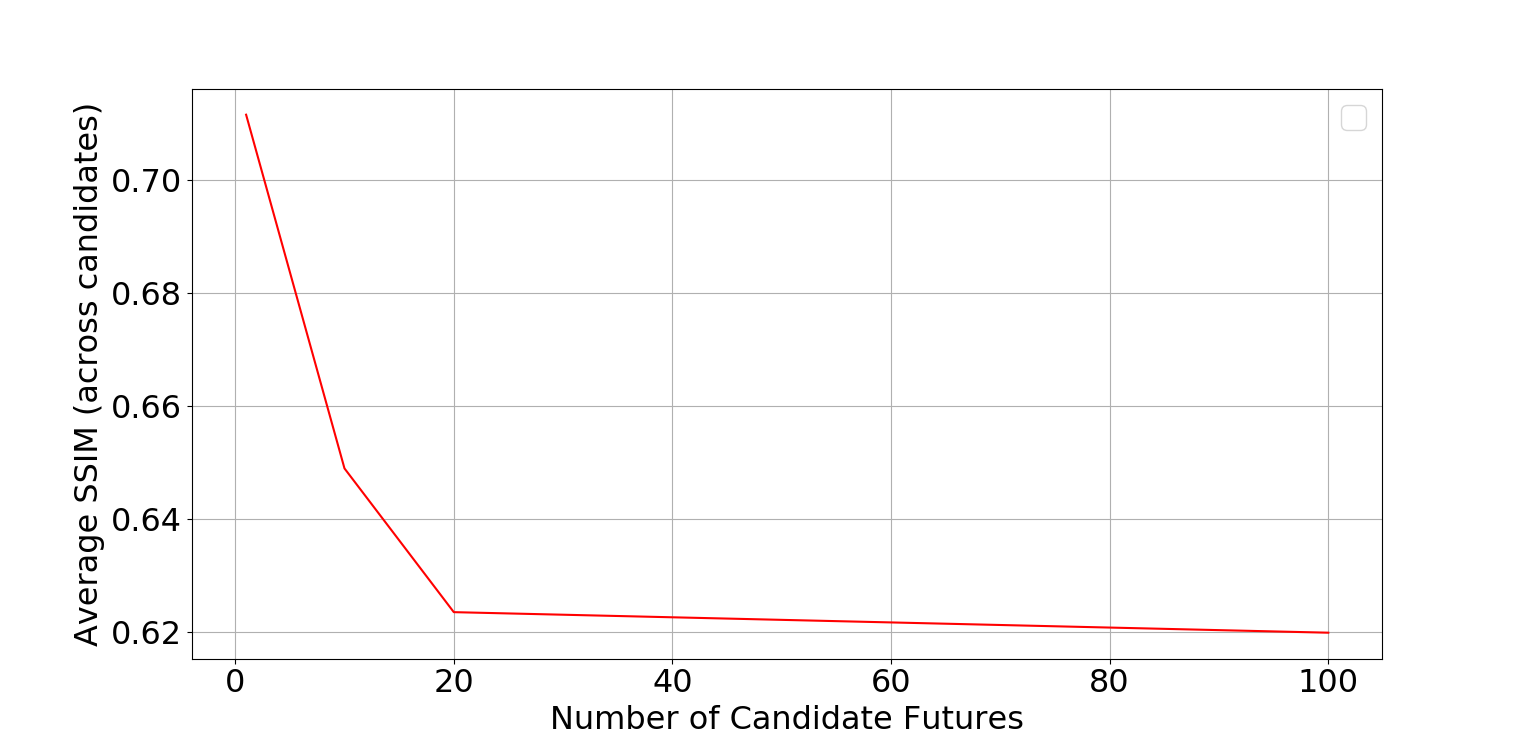}}
    \end{center}
   \caption{Diversity, measured by average SSIM across every pair of future candidate, in the generated frames on AudioSet-Drums (left) and YouTube Paintings (right) with increasing number of candidate futures.}
\end{figure}

Figures~\ref{fig:mnist_ablation_futcand_ssim} and ~\ref{fig:mnist_ablation_futcand_psnr} quantify the diversity in our model's generations. The plots reveal that the more the number of candidate frames (number of futures) that are sampled at every time step, during inference, the better is the approximation to the ground truth video. However, sampling more frames comes at the cost of computational complexity. Our experiments show that sampling 100 candidate frames at every time step was a good trade-off. We see the pattern of diverse sample generation, carry over to the real world datasets too, as observed in Figures~\ref{fig:drums_futcand_ssim} and~\ref{fig:painting_futcand_ssim}. Both figures suggest that as the number of candidate futures go up, the diversity of the generated samples increase which is why the average pairwise SSIM between the generated samples tends to go down. 

\subsection{Performance on a Single Generated Video Sample}
\begin{figure*}[t]
    \begin{center}
    \subfigure[SSIM]{\label{fig:mnist_ablation_1samp_ssim}\includegraphics[width=1.0\linewidth]{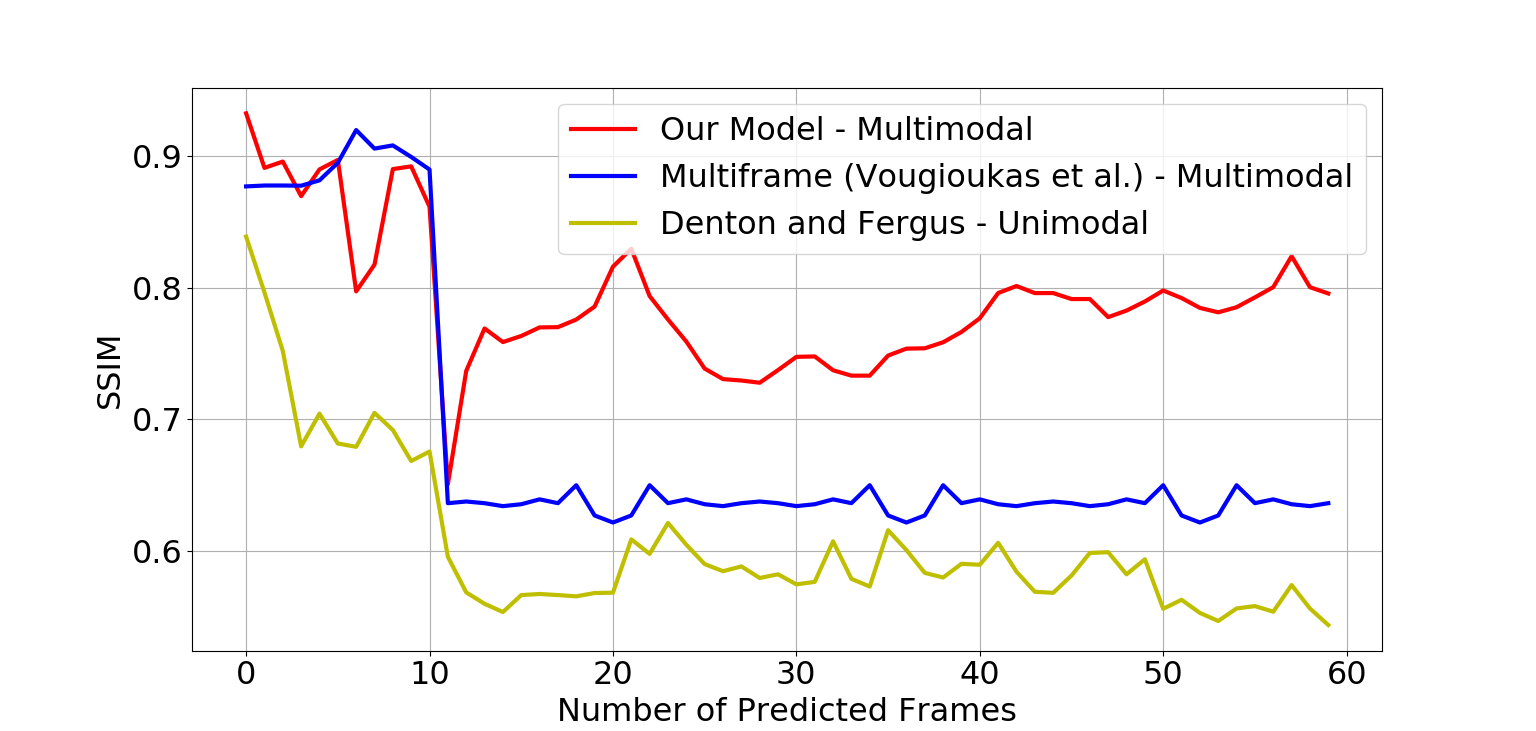}}
    \subfigure[PSNR]{\label{fig:mnist_ablation_1samp_psnr}\includegraphics[width=1.0\linewidth]{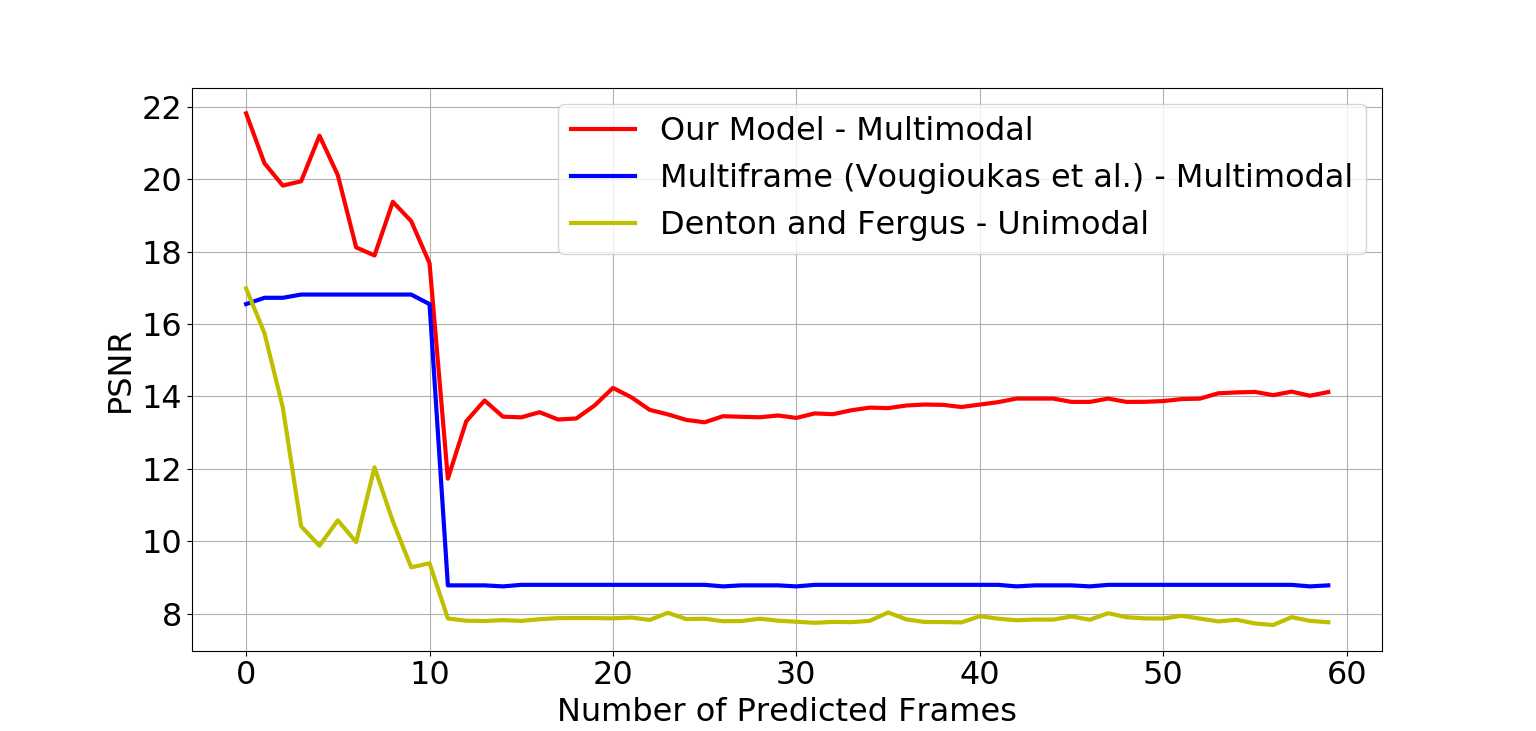}}
    \end{center}
   \caption{Performance on a random sample from the MovingMNIST with obstacle dataset vis-\'a-vis the most competitive baselines, measured by SSIM (top) and PSNR (bottom).}
\end{figure*}

Figures~\ref{fig:mnist_ablation_1samp_ssim} and ~\ref{fig:mnist_ablation_1samp_psnr} show the performance of a prototypical sample from the M3SO dataset, vis-\'a-vis competing state-of-the-art methods. Both these plots show that our approach, outperforms other methods by a significant margin. A closer look into these plots reveal more interesting details. For instance, we notice a sudden dip in performance of all methods at the $12$-th frame (frame index number 11). This is due to the sudden appearance of the block at this frame, whose location is not known in advance. However, our model's ability to perceive both audio and visual modalities, allows it to improve its performance as the digit interacts with the updated environment more and more. For instance, a collision with the block is indicated by a sound of a certain tone, which helps to localize the block with respect to the position of the digit. Such useful localization cues are absent in unimodal (vision only) approaches, resulting in poor performance. On the other hand, the approach of Multiframe - Vougioukas \etal~\cite{vougioukas2018end}, though multimodal, does not synchronize the audio and visual modalities, and thus fails.

\subsection{Training with Teacher Forcing}
We considered the impact of using \textit{Teacher-Forcing} for training our model, since such a strategy has shown promise for deterministic sequence-to-sequence models~\cite{yuan2017machine}. Here the model is trained with ground-truth frames for the first 100 epochs, but subsequently for every batch, a Bernoulli random variable is sampled to determine if the model is going to be trained with the ground-truth frames ($X_t$) or by feeding back the synthesized frames ($\hat{X}_t$) as input (as is the case during inference). We observe from the results in Table~\ref{tab:ablation_mnist_painiting} that for both M3SO-NB and for YouTube Painting, the \textit{Teacher-Forcing} variant performs similarly to our original training strategy. We surmise that this is due to the stochastic nature of our model, which permits it to adapt to variations in input data distribution.

\subsection{Audio-Video Synchronization}
\begin{figure*}[t]
    \begin{center}
    \includegraphics[width=1.0\linewidth]{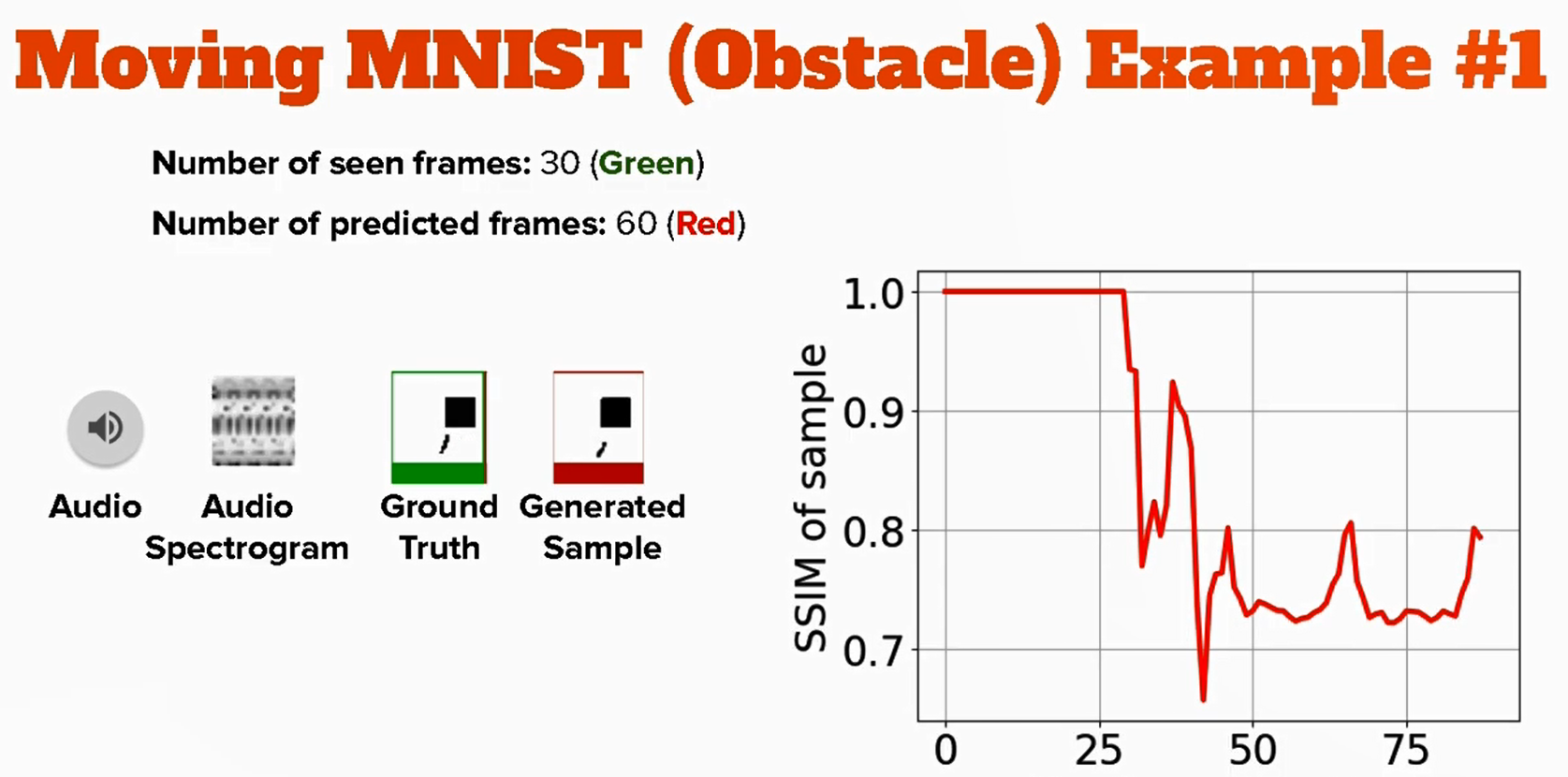}
    \end{center}
   \caption{A screenshot from the video illustrating how the frames predicted correlate with the ground-truth (as measured by SSIM) and the input audio signal. \textbf{Please see the attached video clip.}}
    \label{fig:sync_plot}
\end{figure*}

We further validate the ability of our technique to synchronize the audio and visual inputs. A time-evolving SSIM measure of a randomly chosen sample from the test set of M3SO, reveals how our method's predictions can adapt to the stochasticity of the input data  (see Figure~\ref{fig:sync_plot}). We refer the interested reader to the attached video to have a better understanding.

Additionally the importance of synchronized audio-visual context is evaluated. In order to do so, we train our model in the standard setup but at inference time we initialize the model with mismatched audio-visual inputs, i.e. the past visual context is not from the same sample from which the audio is drawn. The SSIM scores of the generated frames under this setting is shown in Table~\ref{tab:ablation_mnist_painiting} (`AV Mismatch'). As is evident from the sub-par performance of this setting, the alignment of the audio and visual inputs is critical for good generation. 

\section{Qualitative Results}
\begin{figure*}[]
    \begin{center}
    \includegraphics[width=1.0\linewidth]{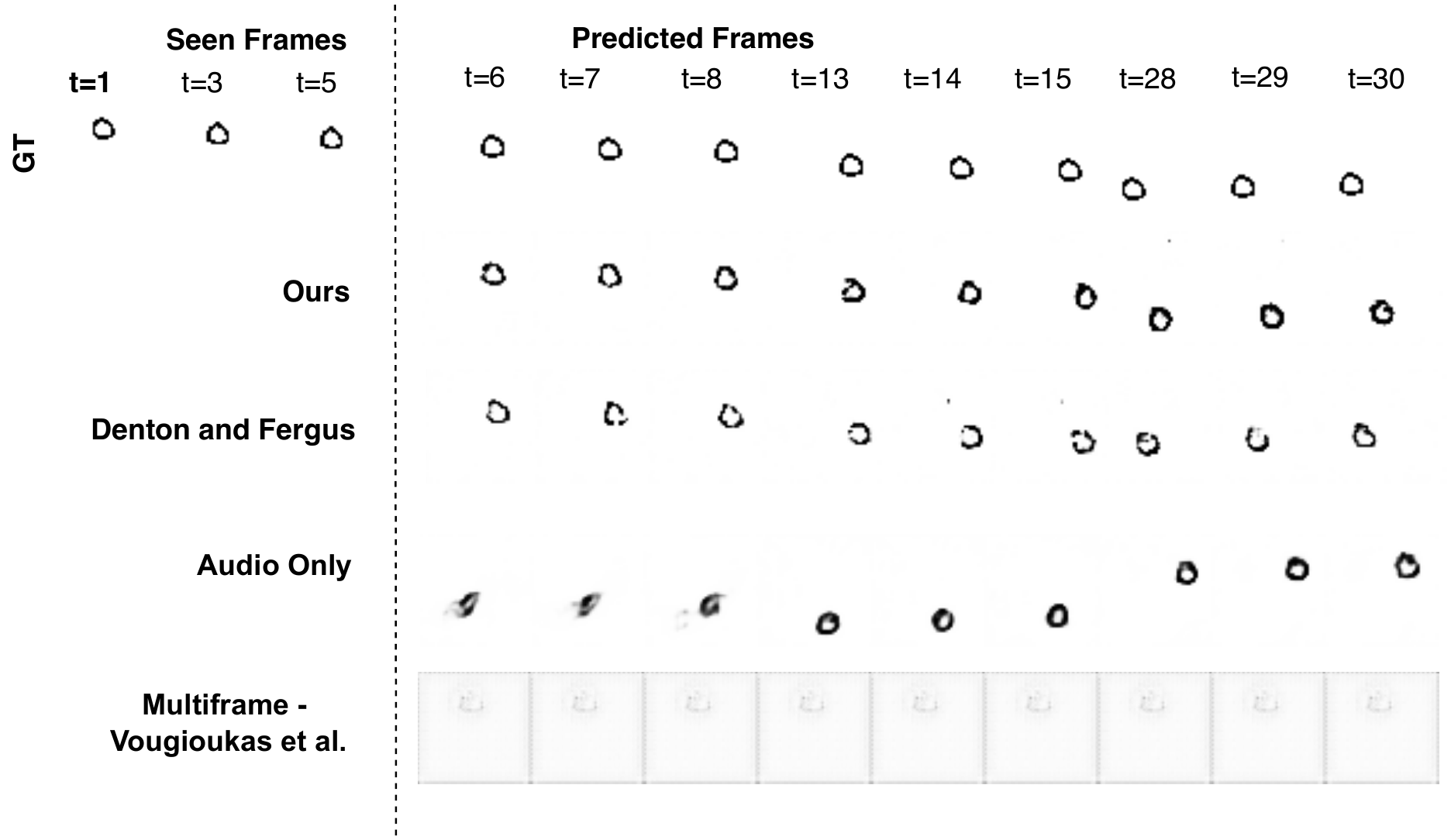}
    \end{center}
   \caption{Sample generations on the M3SO-NB dataset by our method vis-\'a-vis other baselines.}
    \label{fig:mnist_5_15_10}
\end{figure*}

\begin{figure*}[]
    \begin{center}
    \includegraphics[width=1.0\linewidth]{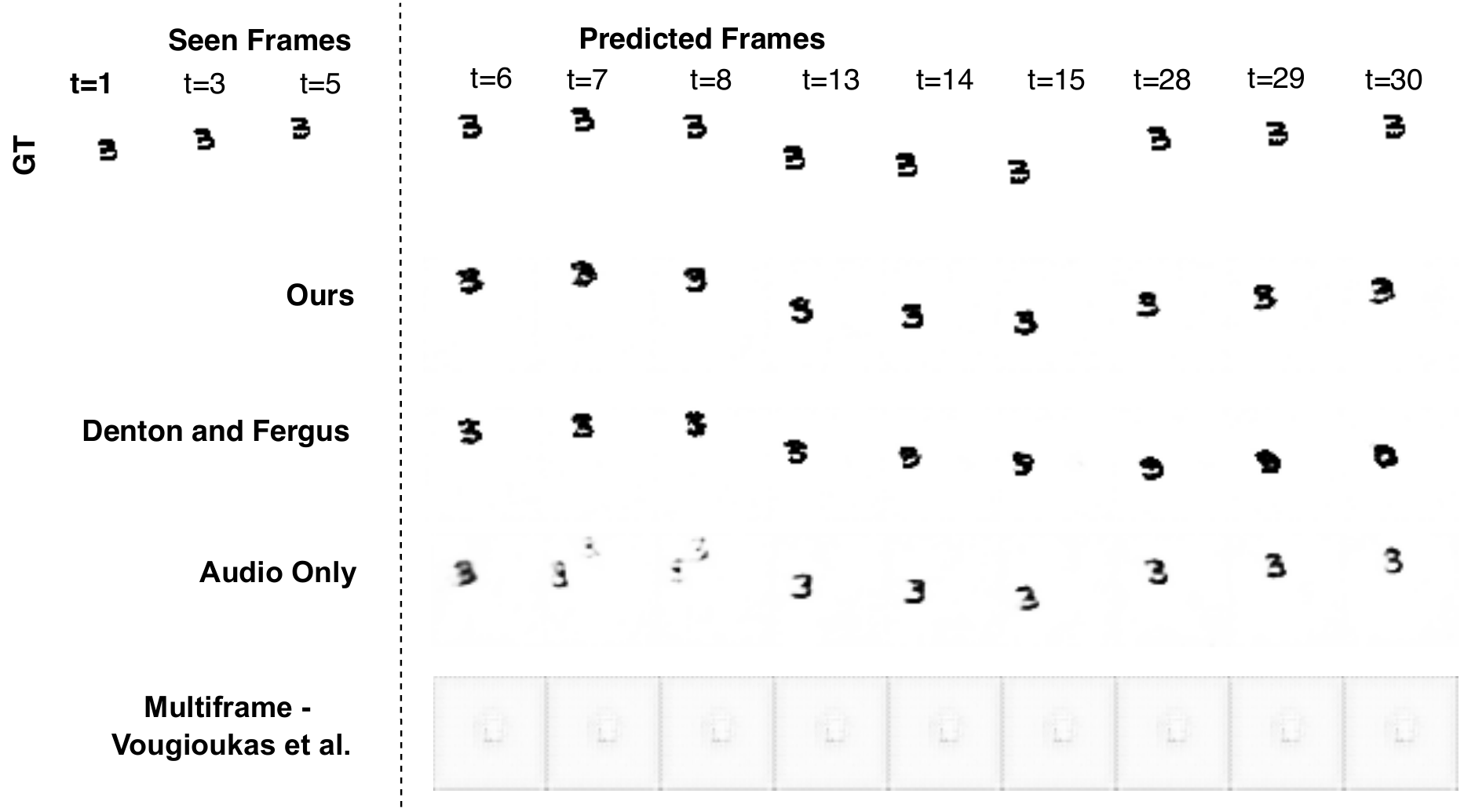}
    \end{center}
  \caption{Sample generations on the M3SO-NB dataset by our method vis-\'a-vis other baselines.}
    \label{fig:mnist_5_15_908}
\end{figure*}

In the following, we present qualitative results of video generation by our method vis-\'a-vis other state-of-the-art baselines on the Multimodal Moving MNIST dataset (both with and without obstacle) and the real-world YouTube Painting dataset and the AudioSet-Drums~\cite{gemmeke2017audio} datasets. Figures~\ref{fig:mnist_5_15_10}, ~\ref{fig:mnist_5_15_908}, show visualizations of the frames generated by our method and those by competing baselines on the M3SO-NB dataset. The figures make the case for the superiority of our method in capturing both a digit's appearance and location accurately. We did not observe much of a qualitative difference between the methods of Vougioukas \etal~\cite{vougioukas2018end} and its Multiframe version, across any of the datasets. Hence for brevity, we show only the latter. 

\begin{figure*}[]
    \begin{center}
    \includegraphics[width=1.0\linewidth]{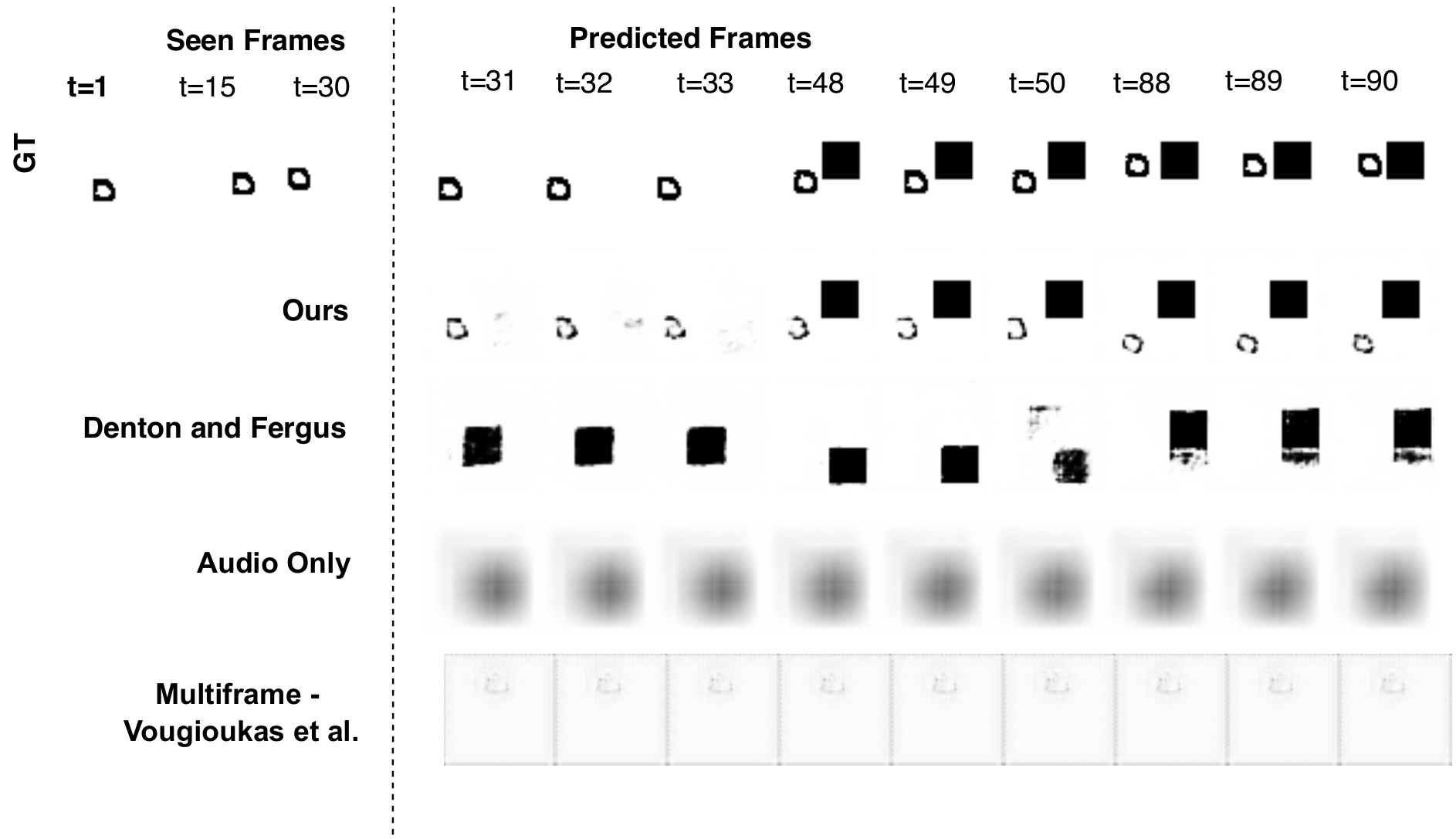}
    \end{center}
   \caption{Sample generations on the MovingMNIST with Surprise Obstacle dataset by our method vis-\'a-vis other baselines.}
    \label{fig:mnist_30_30_619}
\end{figure*}

\begin{figure*}[]
    \begin{center}
    \includegraphics[width=1.0\linewidth]{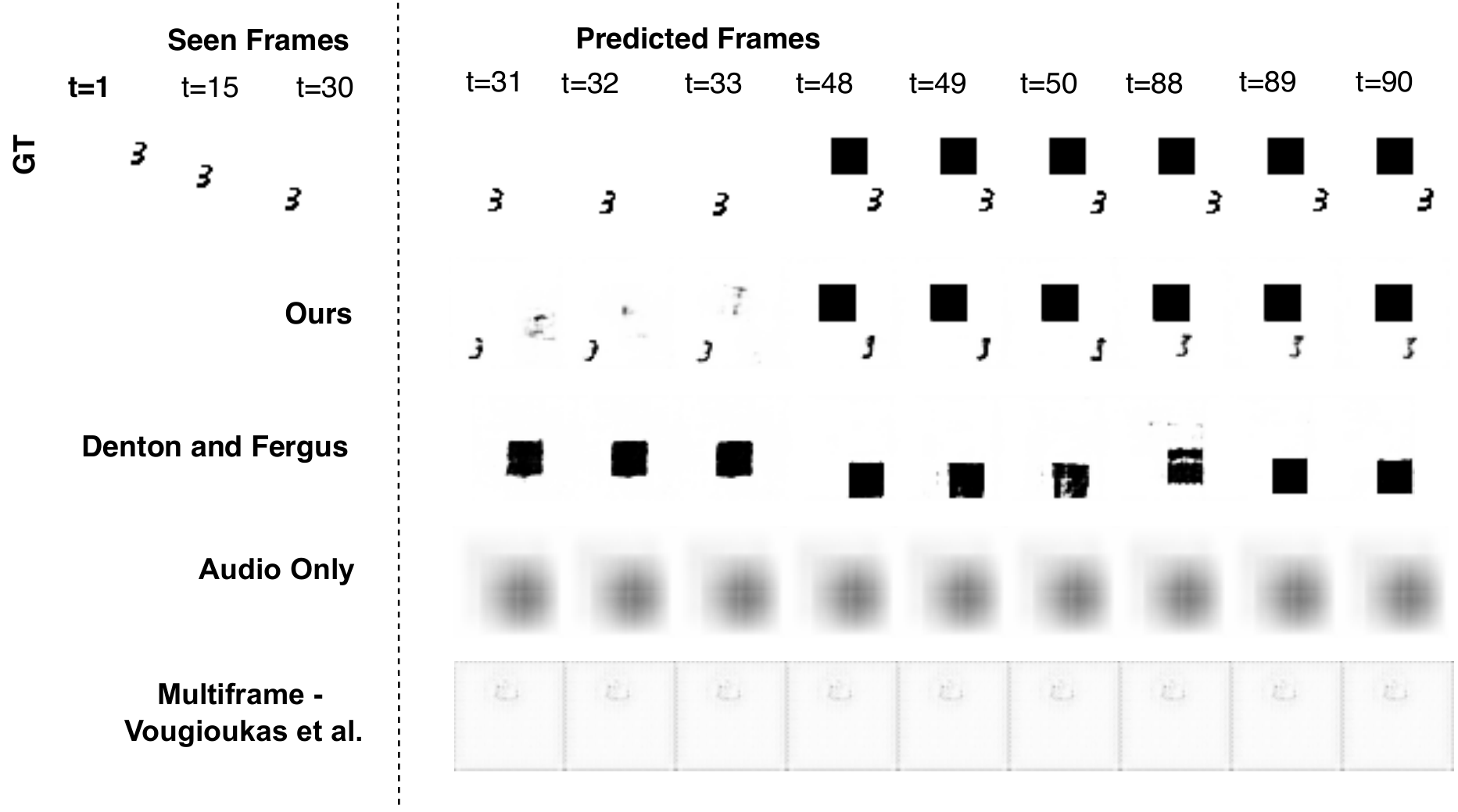}
    \end{center}
   \caption{Sample generations on the MovingMNIST with Surprise Obstacle dataset by our method vis-\'a-vis other baselines.}
    \label{fig:mnist_30_30_136}
\end{figure*}

Further in the case when the challenges are heightened by introducing an obstacle, we see that our approach demonstrates reasonable performance of localizing both the block and the digit, albeit its appearance is slightly morphed. This stretches beyond the performance of the baseline methods, significantly. In particular, we notice the total disappearance of the digits in each of the baselines. (see Figures~\ref{fig:mnist_30_30_619}, ~\ref{fig:mnist_30_30_136}. The distinction is more sharply observed, when we see the associated video (attachment in the supplementary materials). 

\begin{figure*}[]
    \begin{center}
    \includegraphics[width=1.0\linewidth]{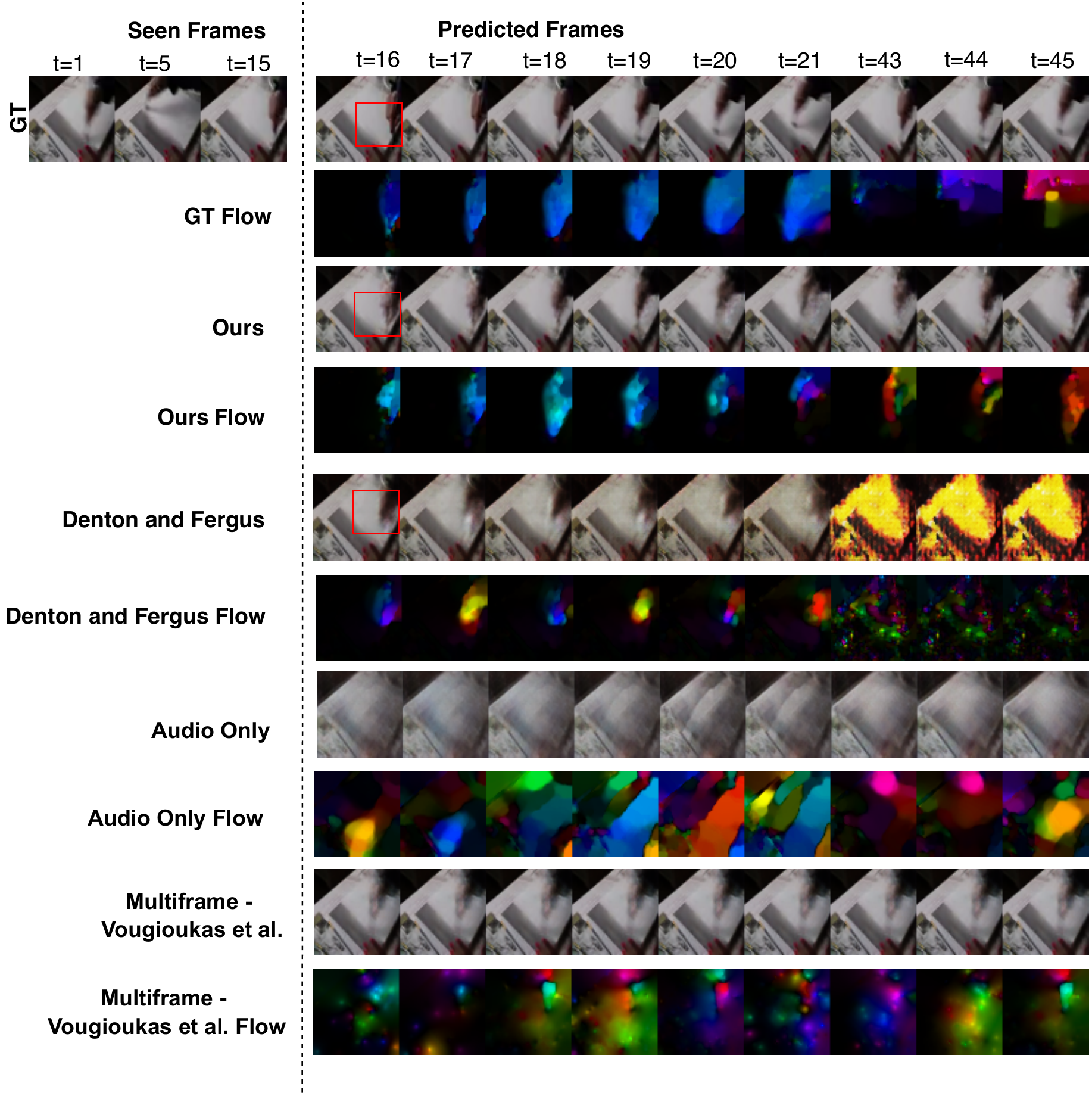}
    \end{center}
   \caption{Sample generations from the YouTube-Painting dataset by our method vis-\'a-vis other baselines and optical flows across frames. The red squares denotes regions of high motion.}
    \label{fig:painting_421}
\end{figure*}

\begin{figure*}[]
    \begin{center}
    \includegraphics[width=1.0\linewidth]{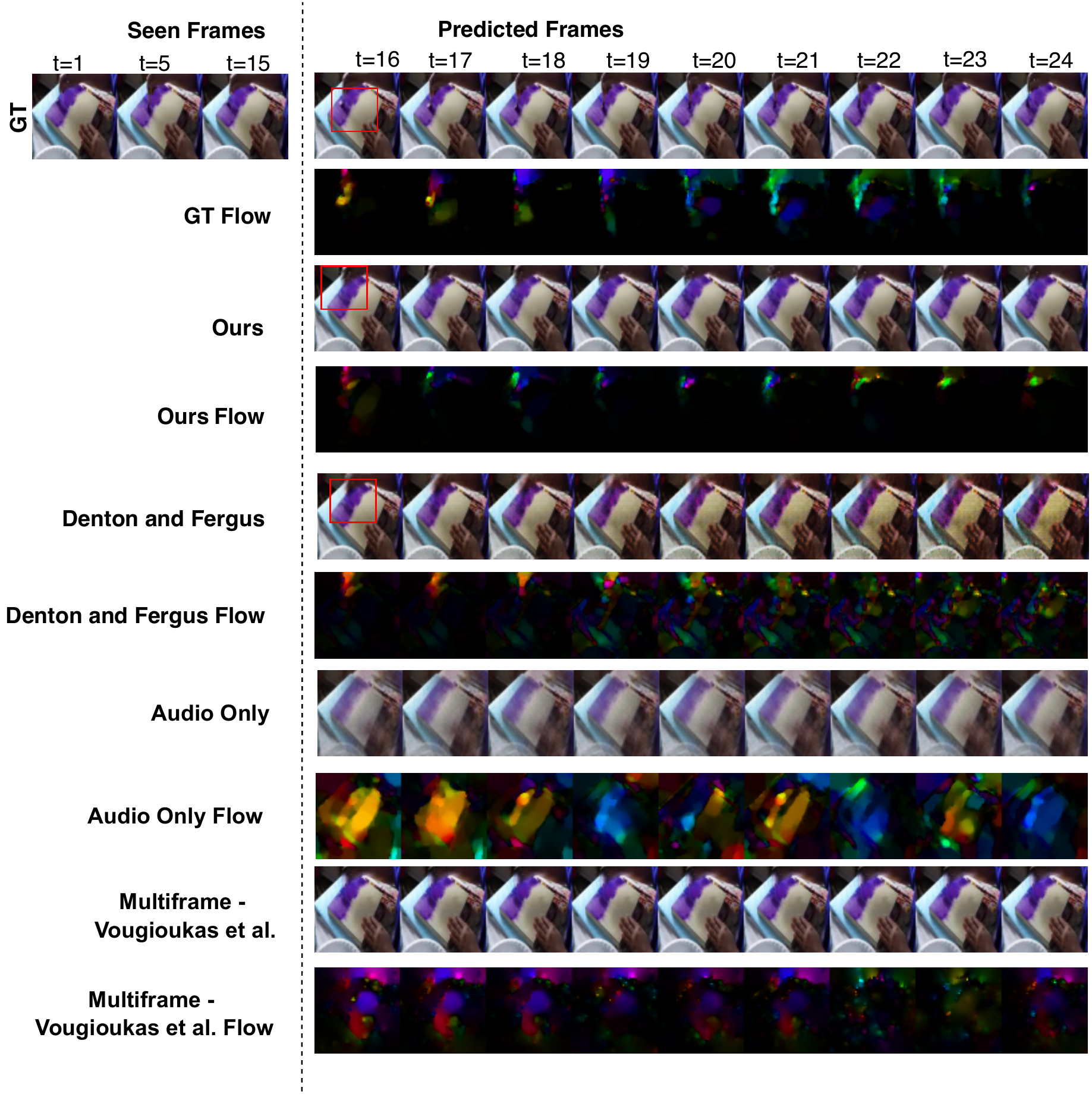}
    \end{center}
   \caption{Sample generations from the YouTube-Painting dataset by our method vis-\'a-vis other baselines and optical flows across frames. The red squares denotes regions of high motion.}
    \label{fig:painting_325}
\end{figure*}

\begin{figure*}[]
    \begin{center}
    \includegraphics[width=1.0\linewidth]{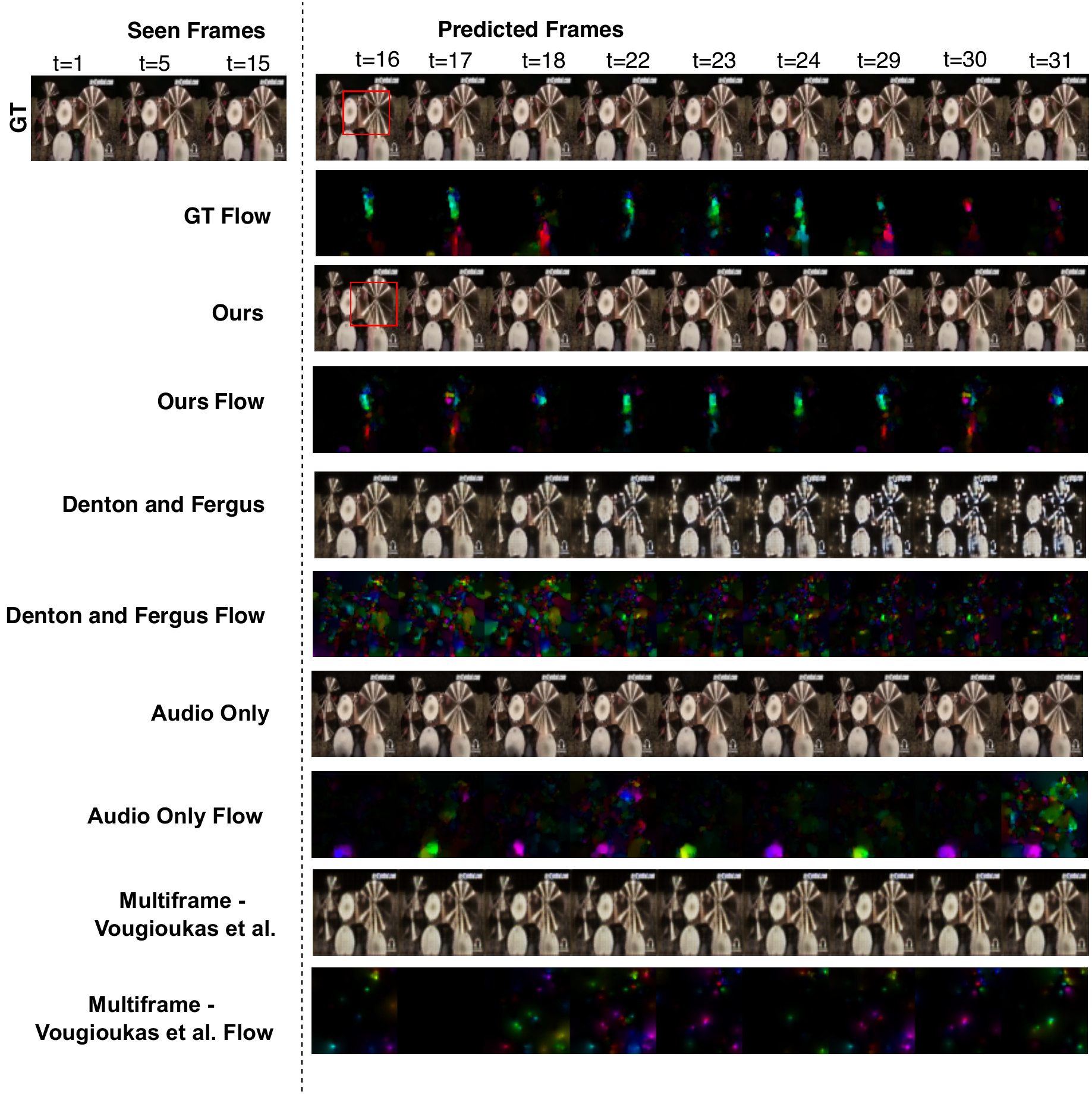}
    \end{center}
   \caption{Sample generations on the AudioSet-Drums dataset by our method vis-\'a-vis other baselines and optical flows across frames. The red square denotes regions of high motion.}
    \label{fig:drums_143}
\end{figure*}

In Figures~\ref{fig:painting_421}, ~\ref{fig:painting_325}, ~\ref{fig:drums_143}, we see how our method fares against the state-of-the-art in generating frames from the real world datasets of YouTube Painting  and AudioSet-Drums~\cite{gemmeke2017audio}, respectively. In all the cases, we see that our method is the closest to the ground-truth and further doesn't introduce artifacts, such as discoloration which some of the baseline methods suffer from. A comparison of the optical flow outputs corroborates this observation. Additionally, the relative crispness of the hand region is suggestive of the fact that our approach is also better at modeling the dynamics of the video. We recommend the interested reader to the accompanying videos to gain a better understanding of the generation performance.

\subsection{Diverse Sample Generations}
\begin{figure*}[]
    \begin{center}
    \includegraphics[width=1.0\linewidth]{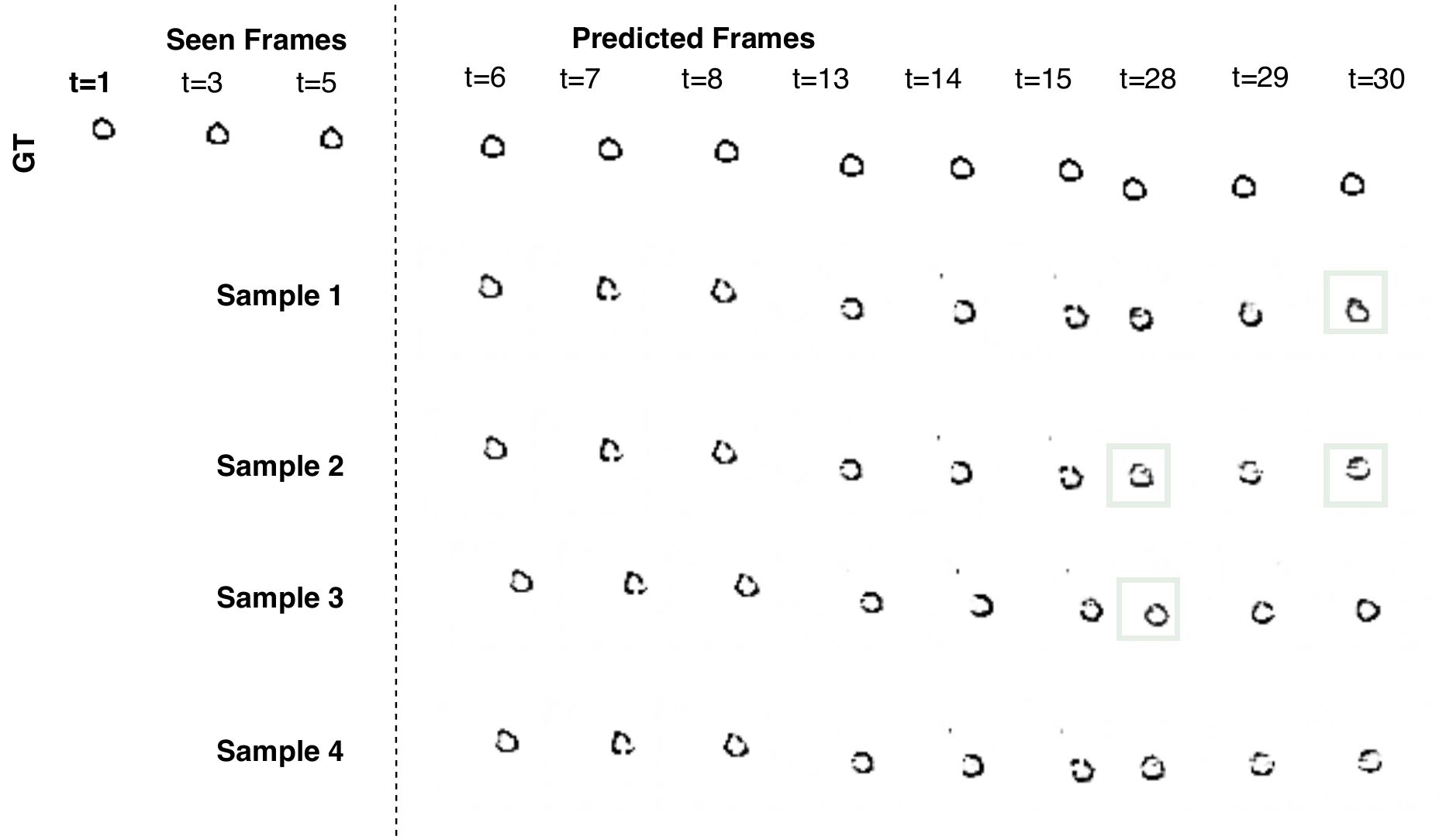}
    \end{center}
   \caption{Diverse sample generations on the M3SO-NB dataset by our method. The green square highlights frames where noticeable differences are observed across samples.}
    \label{fig:mnist_5_15_stochasticity_10}
\end{figure*}

\begin{figure*}[]
    \begin{center}
    \includegraphics[width=1.0\linewidth]{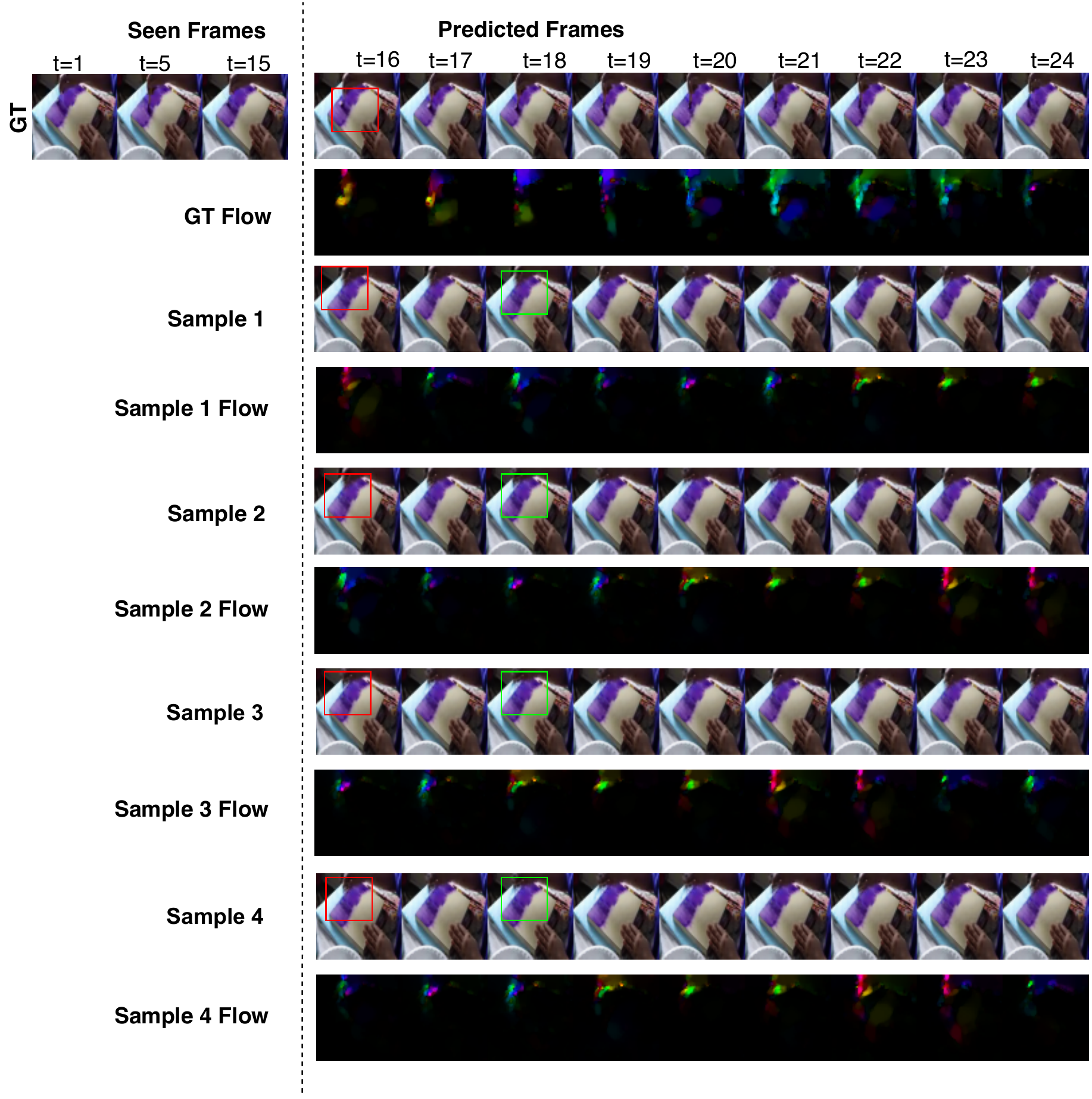}
    \end{center}
   \caption{Diverse sample generations on the YouTube Painting dataset by our method along with the optical flows between frames. The red square denotes regions of high motion, while the green square highlights frames where noticeable differences are observed across samples.}
    \label{fig:painting_stochasticity_325}
\end{figure*}

\begin{figure*}[]
    \begin{center}
    \includegraphics[width=1.0\linewidth]{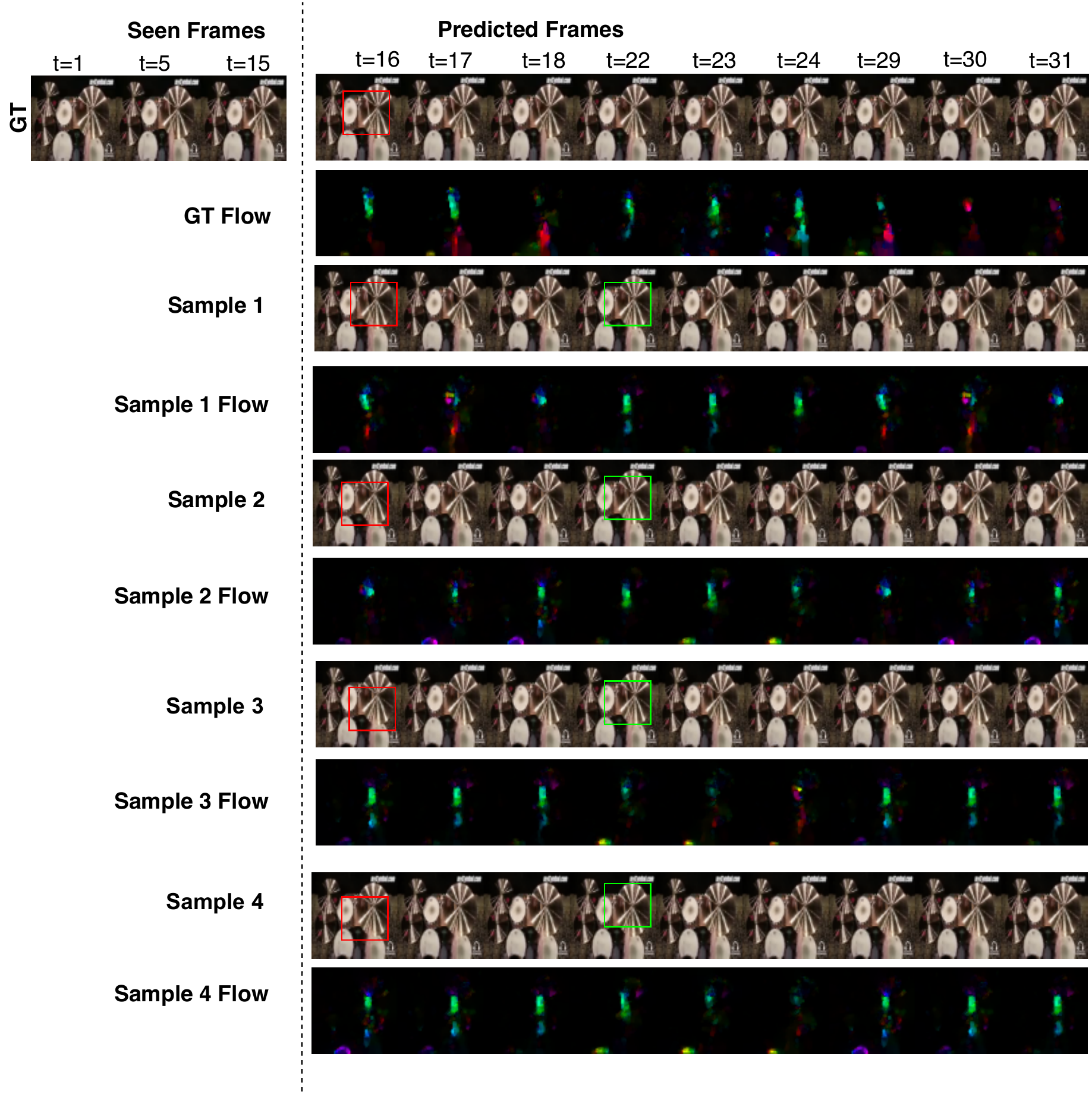}
    \end{center}
   \caption{Diverse sample generations on the AudioSet Drums dataset by our method along with the optical flows between frames. The red square denotes regions of high motion, while the green square highlights frames where noticeable differences are observed across samples.}
    \label{fig:drums_stochasticity_143}
\end{figure*}

In Figures ~\ref{fig:mnist_5_15_stochasticity_10},  ~\ref{fig:painting_stochasticity_325}, ~\ref{fig:drums_stochasticity_143}, we present qualitative visualizations of a set of diverse generations for every sample across all datasets. The green box highlights frames which are noticeably distinct across samples, underscoring the variety of the generated samples for both synthetic and real-world datasets. 

\begin{figure*}[]
    \begin{center}
    \includegraphics[width=1.0\linewidth]{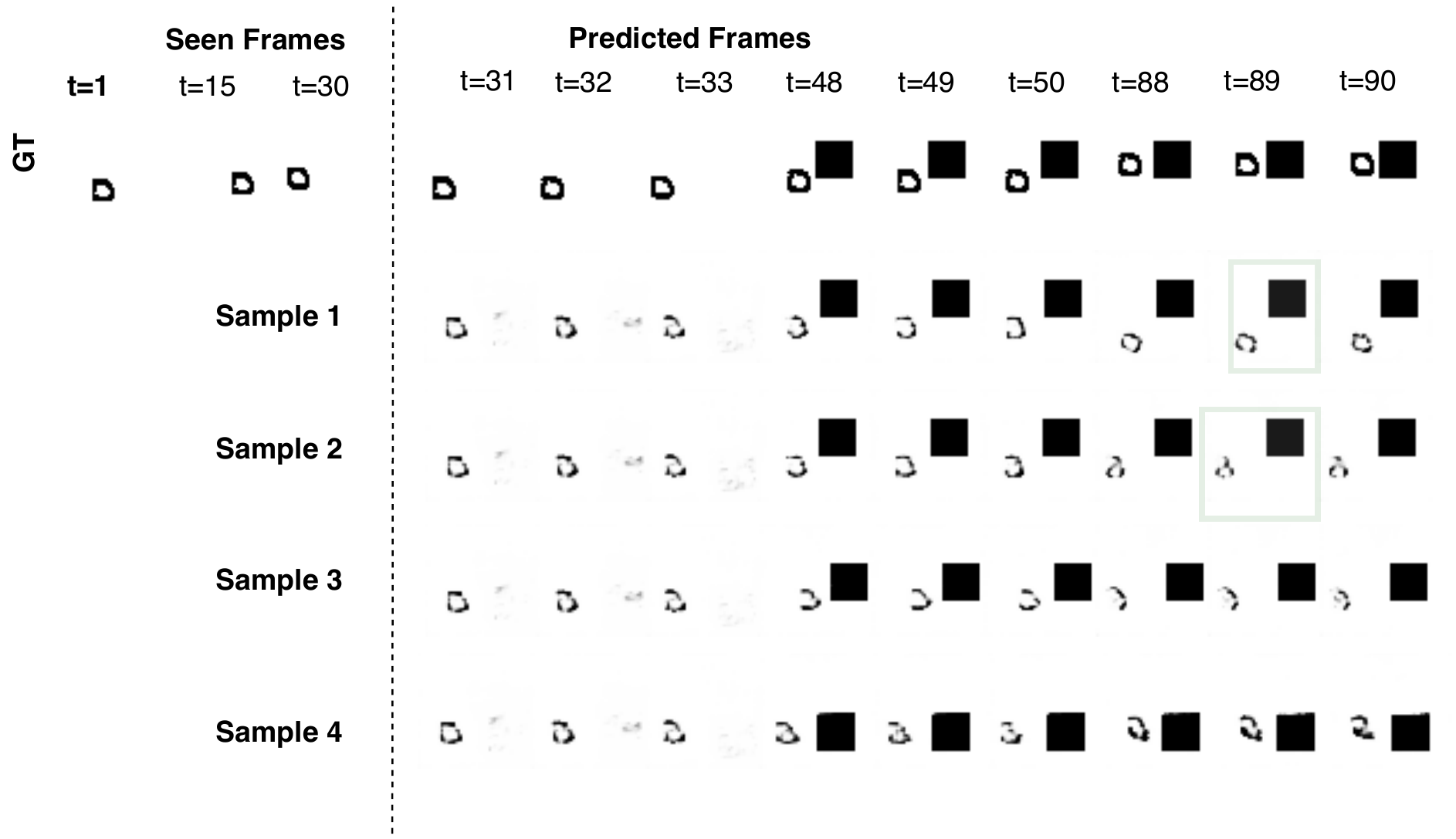}
    \end{center}
   \caption{Diverse sample generations on the Multimodal MovingMNIST with Surprise Obstacle dataset by our method. The green square highlights frames where noticeable differences are observed across samples.}
    \label{fig:mnist_30_30_stochasticity_619}
\end{figure*}

\begin{figure*}[]
    \begin{center}
    \includegraphics[width=1.0\linewidth]{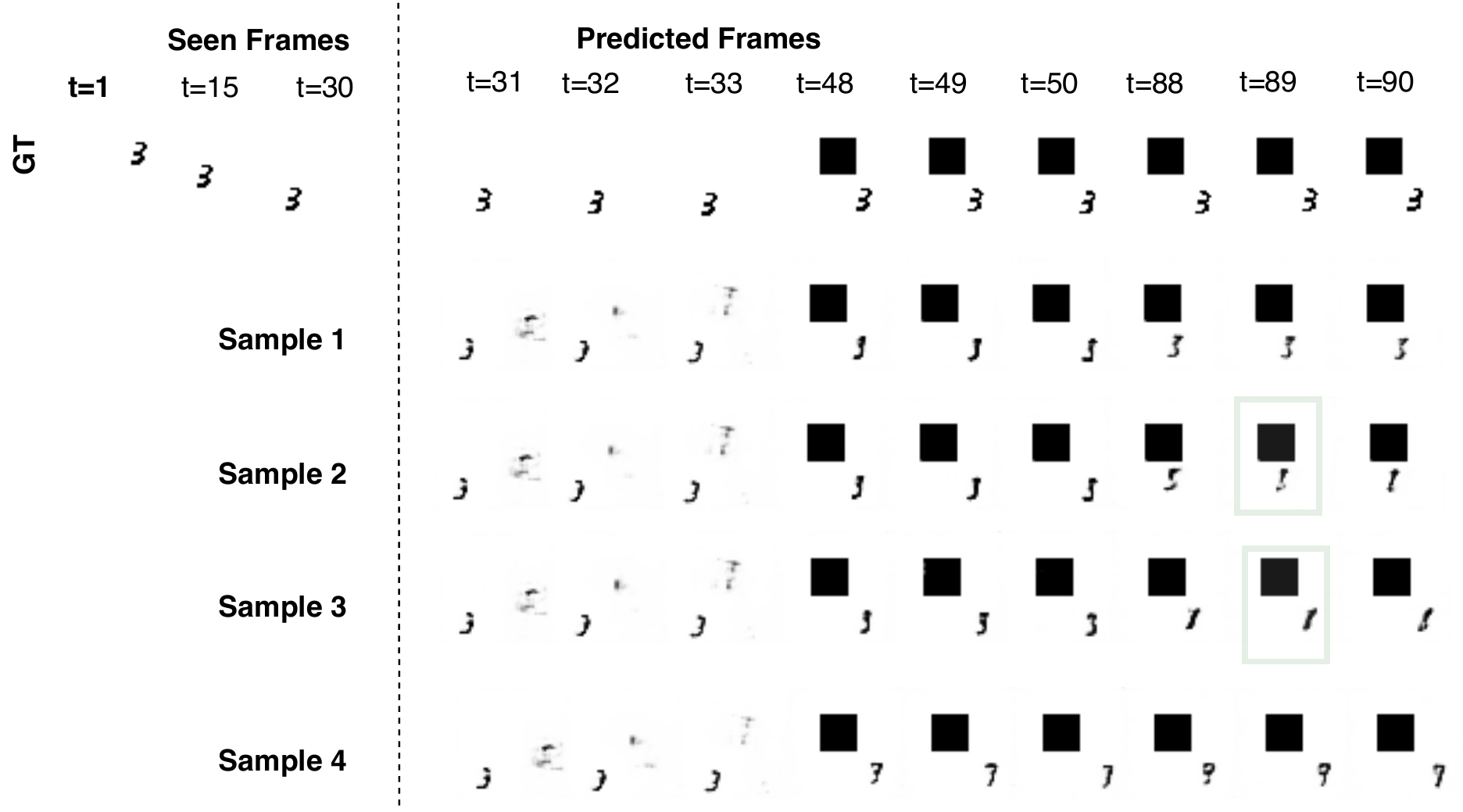}
    \end{center}
   \caption{Diverse sample generations on the Multimodal MovingMNIST with Surprise Obstacle dataset by our method. The green square highlights frames where noticeable differences are observed across samples.}
    \label{fig:mnist_30_30_stochasticity_136}
\end{figure*}

\subsection{Failure Cases}
\begin{figure*}[]
    \begin{center}
    \includegraphics[width=1.0\linewidth]{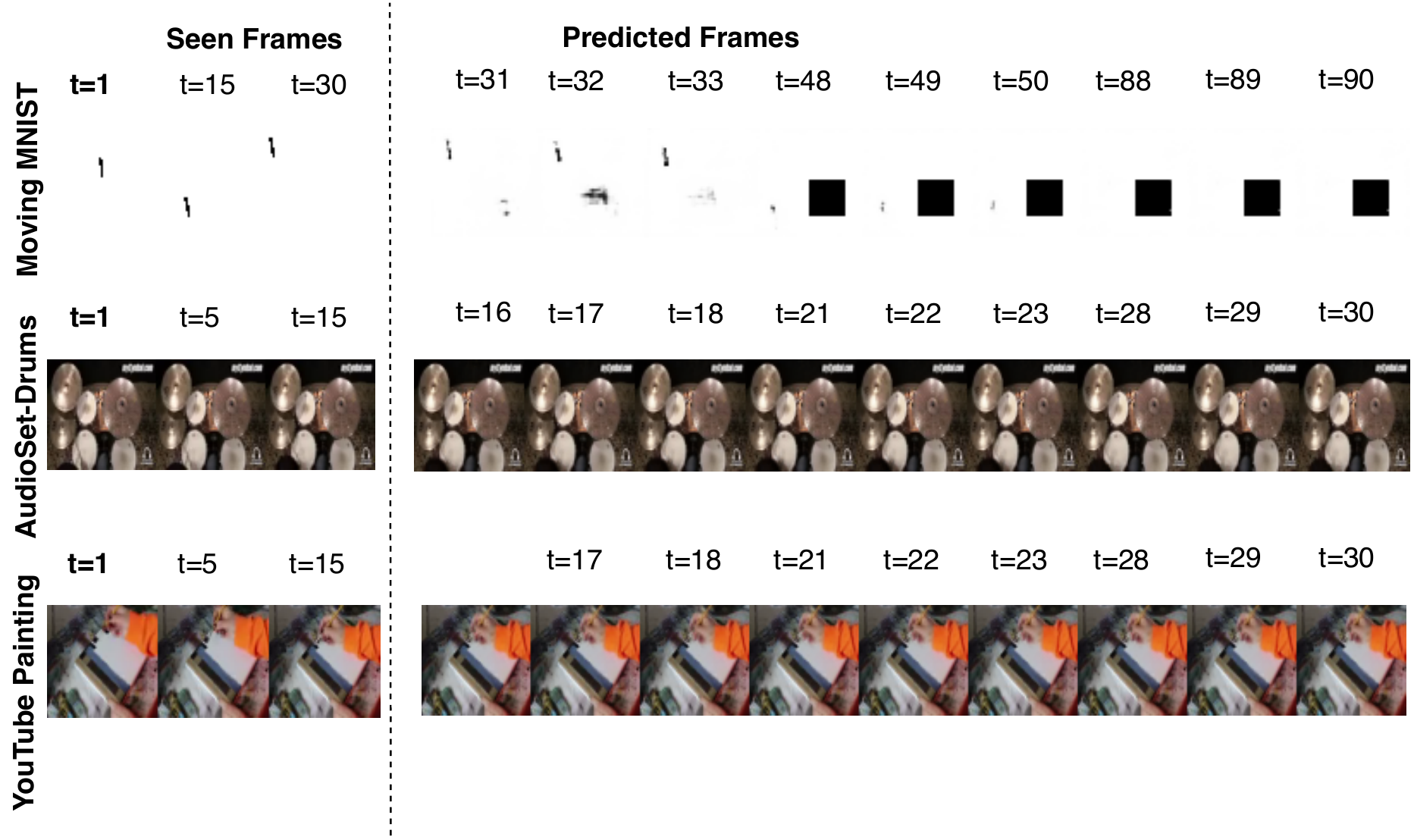}
    \end{center}
   \caption{An assortment of some of the failure cases of our method on the 3 datasets.}
    \label{fig:neg_ex}
\end{figure*}

Figure~\ref{fig:neg_ex} shows some scenarios where our model fails to generate visually compelling frames. This is mainly seen when the region of motion in the seen frames is localized to a small region. Our model, in such cases, essentially displays a static frame. This is typified by the slender `1' in Multimodal MovingMNIST or the limited hand motion (Figure~\ref{fig:neg_ex}) in the case of the YouTube Painting dataset. We intend to resolve this issue in our future work by replacing the Mean-Squared loss term in our objective, which uniformly penalizes all pixels, with a weighted version that would attend more to `regions of interest' - where more motion is observed.

\end{document}

%% file: definitions.tex
\newcommand{\vid}{V}
\newcommand{\dataset}{\mathcal{D}}
\newcommand{\set}[1]{\left\{#1\right\}}
\newcommand{\seq}[1]{\langle{#1}\rangle}

\DeclareMathOperator{\lstm}{LSTM}
\DeclareMathOperator{\kl}{KL}
\DeclareMathOperator{\softmax}{softmax}
\DeclareMathOperator{\concat}{concat}

\newcommand{\tuple}[1]{\left({#1}\right)}
\newcommand{\normal}{\mathcal{N}}
\newcommand{\dd}{\parallel}
\newcommand{\expect}{\mathop{\mathbb{E}}}
\newcommand{\etal}{\textit{et al.}}

%% file: S2S_single.bbl
\begin{thebibliography}{10}
\providecommand{\url}[1]{\texttt{#1}}
\providecommand{\urlprefix}{URL }
\providecommand{\doi}[1]{https://doi.org/#1}

\bibitem{wav2pix2019icassp}
Amanda~Duarte, Francisco~Roldan, e.a.: Wav2pix: Speech-conditioned face
  generation using generative adversarial networks. In: ICASSP (2019)

\bibitem{arandjelovic2017look}
Arandjelovic, R., Zisserman, A.: Look, listen and learn. In: Proceedings of the
  IEEE International Conference on Computer Vision. pp. 609--617 (2017)

\bibitem{Youtube:Taylor}
ASMR, T.: Painting ASMR (2019 (accessed November 5, 2019)),
  \url{https://www.youtube.com/playlist?list=PL5Y0dQ2DJHj47sK5jsbVkVpTQ9r7T090X}

\bibitem{aytar2016soundnet}
Aytar, Y., Vondrick, C., Torralba, A.: Soundnet: Learning sound representations
  from unlabeled video. In: Advances in neural information processing systems.
  pp. 892--900 (2016)

\bibitem{babaeizadeh2017stochastic}
Babaeizadeh, M., Finn, C., Erhan, D., Campbell, R.H., Levine, S.: Stochastic
  variational video prediction. arXiv preprint arXiv:1710.11252  (2017)

\bibitem{brock2018large}
Brock, A., Donahue, J., Simonyan, K.: Large scale gan training for high
  fidelity natural image synthesis. arXiv preprint arXiv:1809.11096  (2018)

\bibitem{chen2019hierarchical}
Chen, L., Maddox, R.K., Duan, Z., Xu, C.: Hierarchical cross-modal talking face
  generation with dynamic pixel-wise loss. In: CVPR (2019)

\bibitem{chen2017deep}
Chen, L., Srivastava, S., Duan, Z., Xu, C.: Deep cross-modal audio-visual
  generation. In: Proceedings of the on Thematic Workshops of ACM Multimedia
  2017. ACM (2017)

\bibitem{corlett2018conditioned}
Corlett, P.R., Powers, A.R.: Conditioned hallucinations: historic insights and
  future directions. World Psychiatry  \textbf{17}(3), ~361 (2018)

\bibitem{denton2018stochastic}
Denton, E., Fergus, R.: Stochastic video generation with a learned prior. In:
  International Conference on Machine Learning. pp. 1182--1191 (2018)

\bibitem{deshpande2018generative}
Deshpande, I., Zhang, Z., Schwing, A.G.: Generative modeling using the sliced
  wasserstein distance. In: Proceedings of the IEEE Conference on Computer
  Vision and Pattern Recognition. pp. 3483--3491 (2018)

\bibitem{finn2016unsupervised}
Finn, C., Goodfellow, I., Levine, S.: Unsupervised learning for physical
  interaction through video prediction. In: Advances in neural information
  processing systems. pp. 64--72 (2016)

\bibitem{fragkiadaki2015learning}
Fragkiadaki, K., Agrawal, P., Levine, S., Malik, J.: Learning visual predictive
  models of physics for playing billiards. arXiv preprint arXiv:1511.07404
  (2015)

\bibitem{gemmeke2017audio}
Gemmeke, J.F., Ellis, D.P., Freedman, D., Jansen, A., Lawrence, W., Moore,
  R.C., Plakal, M., Ritter, M.: Audio set: An ontology and human-labeled
  dataset for audio events. In: 2017 IEEE International Conference on
  Acoustics, Speech and Signal Processing (ICASSP). pp. 776--780. IEEE (2017)

\bibitem{goodfellow2014generative}
Goodfellow, I., Pouget-Abadie, J., Mirza, M., Xu, B., Warde-Farley, D., Ozair,
  S., Courville, A., Bengio, Y.: Generative adversarial nets. In: Advances in
  neural information processing systems. pp. 2672--2680 (2014)

\bibitem{gulrajani2017improved}
Gulrajani, I., Ahmed, F., Arjovsky, M., Dumoulin, V., Courville, A.C.: Improved
  training of wasserstein gans. In: Advances in neural information processing
  systems. pp. 5767--5777 (2017)

\bibitem{gupta2018imagine}
Gupta, T., Schwenk, D., Farhadi, A., Hoiem, D., Kembhavi, A.: Imagine this!
  scripts to compositions to videos. In: Proceedings of the European Conference
  on Computer Vision (ECCV). pp. 598--613 (2018)

\bibitem{hao2018cmcgan}
Hao, W., Zhang, Z., Guan, H.: Cmcgan: A uniform framework for cross-modal
  visual-audio mutual generation. In: AAAI (2018)

\bibitem{hao2018controllable}
Hao, Z., Huang, X., Belongie, S.: Controllable video generation with sparse
  trajectories. In: Proceedings of the IEEE Conference on Computer Vision and
  Pattern Recognition. pp. 7854--7863 (2018)

\bibitem{harwath2016unsupervised}
Harwath, D., Torralba, A., Glass, J.: Unsupervised learning of spoken language
  with visual context. In: Advances in Neural Information Processing Systems.
  pp. 1858--1866 (2016)

\bibitem{hochreiter1997long}
Hochreiter, S., Schmidhuber, J.: Long short-term memory. Neural computation
  \textbf{9}(8),  1735--1780 (1997)

\bibitem{hsieh2018learning}
Hsieh, J.T., Liu, B., Huang, D.A., Fei-Fei, L.F., Niebles, J.C.: Learning to
  decompose and disentangle representations for video prediction. In: Advances
  in Neural Information Processing Systems. pp. 517--526 (2018)

\bibitem{ioffe2015batch}
Ioffe, S., Szegedy, C.: Batch normalization: Accelerating deep network training
  by reducing internal covariate shift. In: International Conference on Machine
  Learning. pp. 448--456 (2015)

\bibitem{jamaludin2019you}
Jamaludin, A., Chung, J.S., Zisserman, A.: You said that?: Synthesising talking
  faces from audio. International Journal of Computer Vision pp. 1--13 (2019)

\bibitem{jia2016dynamic}
Jia, X., De~Brabandere, B., Tuytelaars, T., Gool, L.V.: Dynamic filter
  networks. In: Advances in Neural Information Processing Systems. pp. 667--675
  (2016)

\bibitem{karras2017audio}
Karras, T., Aila, T., Laine, S., Herva, A., Lehtinen, J.: Audio-driven facial
  animation by joint end-to-end learning of pose and emotion. ACM Transactions
  on Graphics (TOG)  \textbf{36}(4), ~94 (2017)

\bibitem{kidron2005pixels}
Kidron, E., Schechner, Y.Y., Elad, M.: Pixels that sound. In: CVPR (2005)

\bibitem{kim2019deep}
Kim, D., Woo, S., Lee, J.Y., Kweon, I.S.: Deep video inpainting. In: CVPR
  (2019)

\bibitem{kingma2014adam}
Kingma, D.P., Ba, J.: Adam: A method for stochastic optimization. arXiv
  preprint arXiv:1412.6980  (2014)

\bibitem{kingma2013auto}
Kingma, D.P., Welling, M.: Auto-encoding variational bayes. arXiv preprint
  arXiv:1312.6114  (2013)

\bibitem{kolouri2018sliced}
Kolouri, S., Pope, P.E., Martin, C.E., Rohde, G.K.: Sliced-wasserstein
  autoencoder: an embarrassingly simple generative model. arXiv preprint
  arXiv:1804.01947  (2018)

\bibitem{lamb2016discriminative}
Lamb, A., Dumoulin, V., Courville, A.: Discriminative regularization for
  generative models. arXiv preprint arXiv:1602.03220  (2016)

\bibitem{lecun1998gradient}
LeCun, Y., Bottou, L., Bengio, Y., Haffner, P., et~al.: Gradient-based learning
  applied to document recognition. Proceedings of the IEEE  \textbf{86}(11),
  2278--2324 (1998)

\bibitem{li2018video}
Li, Y., Min, M.R., Shen, D., Carlson, D., Carin, L.: Video generation from
  text. In: Thirty-Second AAAI Conference on Artificial Intelligence (2018)

\bibitem{lindell2019acoustic}
Lindell, D.B., Wetzstein, G., Koltun, V.: Acoustic non-line-of-sight imaging.
  In: CVPR (2019)

\bibitem{liu2017unsupervised}
Liu, M.Y., Breuel, T., Kautz, J.: Unsupervised image-to-image translation
  networks. In: Advances in neural information processing systems. pp. 700--708
  (2017)

\bibitem{luo2017unsupervised}
Luo, Z., Peng, B., Huang, D.A., Alahi, A., Fei-Fei, L.: Unsupervised learning
  of long-term motion dynamics for videos. In: Proceedings of the IEEE
  Conference on Computer Vision and Pattern Recognition. pp. 2203--2212 (2017)

\bibitem{oh2019speech2face}
Oh, T.H., Dekel, T., Kim, C., Mosseri, I., Freeman, W.T., Rubinstein, M.,
  Matusik, W.: Speech2face: Learning the face behind a voice. In: Proceedings
  of the IEEE Conference on Computer Vision and Pattern Recognition. pp.
  7539--7548 (2019)

\bibitem{owens2018audio}
Owens, A., Efros, A.A.: Audio-visual scene analysis with self-supervised
  multisensory features. In: Proceedings of the European Conference on Computer
  Vision (ECCV). pp. 631--648 (2018)

\bibitem{owens2016visually}
Owens, A., Isola, P., McDermott, J., Torralba, A., Adelson, E.H., Freeman,
  W.T.: Visually indicated sounds. In: Proceedings of the IEEE conference on
  computer vision and pattern recognition. pp. 2405--2413 (2016)

\bibitem{owens2016ambient}
Owens, A., Wu, J., McDermott, J.H., Freeman, W.T., Torralba, A.: Ambient sound
  provides supervision for visual learning. In: European conference on computer
  vision. pp. 801--816. Springer (2016)

\bibitem{pan2019video}
Pan, J., Wang, C., Jia, X., Shao, J., Sheng, L., Yan, J., Wang, X.: Video
  generation from single semantic label map. In: Proceedings of the IEEE
  Conference on Computer Vision and Pattern Recognition. pp. 3733--3742 (2019)

\bibitem{pavlov1910work}
Pavlov, I.P.: The work of the digestive glands. Charles Griffin, Limited;
  Exeter Street, Strand (1910)

\bibitem{ranzato2014video}
Ranzato, M., Szlam, A., Bruna, J., Mathieu, M., Collobert, R., Chopra, S.:
  Video (language) modeling: a baseline for generative models of natural
  videos. arXiv preprint arXiv:1412.6604  (2014)

\bibitem{ronneberger2015u}
Ronneberger, O., Fischer, P., Brox, T.: U-net: Convolutional networks for
  biomedical image segmentation. In: International Conference on Medical image
  computing and computer-assisted intervention. pp. 234--241. Springer (2015)

\bibitem{saito2017temporal}
Saito, M., Matsumoto, E., Saito, S.: Temporal generative adversarial nets with
  singular value clipping. In: Proceedings of the IEEE International Conference
  on Computer Vision. pp. 2830--2839 (2017)

\bibitem{shlizerman2018audio}
Shlizerman, E., Dery, L., Schoen, H., Kemelmacher-Shlizerman, I.: Audio to body
  dynamics. In: Proceedings of the IEEE Conference on Computer Vision and
  Pattern Recognition. pp. 7574--7583 (2018)

\bibitem{srivastava2015unsupervised}
Srivastava, N., Mansimov, E., Salakhudinov, R.: Unsupervised learning of video
  representations using lstms. In: ICML (2015)

\bibitem{sutskever2014sequence}
Sutskever, I., Vinyals, O., Le, Q.: Sequence to sequence learning with neural
  networks. Advances in NIPS  (2014)

\bibitem{suwajanakorn2017synthesizing}
Suwajanakorn, S., Seitz, S.M., Kemelmacher-Shlizerman, I.: Synthesizing obama:
  learning lip sync from audio. ACM Transactions on Graphics (TOG)
  \textbf{36}(4), ~95 (2017)

\bibitem{taylor2017deep}
Taylor, S., Kim, T., Yue, Y., Mahler, M., Krahe, J., Rodriguez, A.G., Hodgins,
  J., Matthews, I.: A deep learning approach for generalized speech animation.
  ACM Transactions on Graphics (TOG)  \textbf{36}(4), ~93 (2017)

\bibitem{tulyakov2018mocogan}
Tulyakov, S., Liu, M.Y., Yang, X., Kautz, J.: Mocogan: Decomposing motion and
  content for video generation. In: Proceedings of the IEEE conference on
  computer vision and pattern recognition. pp. 1526--1535 (2018)

\bibitem{vaswani2017attention}
Vaswani, A., Shazeer, N., Parmar, N., Uszkoreit, J., Jones, L., Gomez, A.N.,
  Kaiser, {\L}., Polosukhin, I.: Attention is all you need. In: Advances in
  neural information processing systems. pp. 5998--6008 (2017)

\bibitem{villegas2017decomposing}
Villegas, R., Yang, J., Hong, S., Lin, X., Lee, H.: Decomposing motion and
  content for natural video sequence prediction. arXiv preprint
  arXiv:1706.08033  (2017)

\bibitem{vondrick2016generating}
Vondrick, C., Pirsiavash, H., Torralba, A.: Generating videos with scene
  dynamics. In: NIPS (2016)

\bibitem{vougioukas2018end}
Vougioukas, K., Petridis, S., Pantic, M.: End-to-end speech-driven facial
  animation with temporal gans. arXiv preprint arXiv:1805.09313  (2018)

\bibitem{walker2017pose}
Walker, J., Marino, K., Gupta, A., Hebert, M.: The pose knows: Video
  forecasting by generating pose futures. In: Proceedings of the IEEE
  International Conference on Computer Vision. pp. 3332--3341 (2017)

\bibitem{wan2019towards}
Wan, C.H., Chuang, S.P., Lee, H.Y.: Towards audio to scene image synthesis
  using generative adversarial network. In: ICASSP (2019)

\bibitem{wang2018video}
Wang, T.C., Liu, M.Y., Zhu, J.Y., Liu, G., Tao, A., Kautz, J., Catanzaro, B.:
  Video-to-video synthesis. arXiv preprint arXiv:1808.06601  (2018)

\bibitem{wang2004image}
Wang, Z., Bovik, A.C., Sheikh, H.R., Simoncelli, E.P., et~al.: Image quality
  assessment: from error visibility to structural similarity. IEEE transactions
  on image processing  \textbf{13}(4),  600--612 (2004)

\bibitem{wu2019sliced}
Wu, J., Huang, Z., Acharya, D., Li, W., Thoma, J., Paudel, D.P., Gool, L.V.:
  Sliced wasserstein generative models. In: Proceedings of the IEEE Conference
  on Computer Vision and Pattern Recognition. pp. 3713--3722 (2019)

\bibitem{xue2016visual}
Xue, T., Wu, J., Bouman, K., Freeman, B.: Visual dynamics: Probabilistic future
  frame synthesis via cross convolutional networks. In: Advances in neural
  information processing systems. pp. 91--99 (2016)

\bibitem{yuan2017machine}
Yuan, X., Wang, T., Gulcehre, C., Sordoni, A., Bachman, P., Subramanian, S.,
  Zhang, S., Trischler, A.: Machine comprehension by text-to-text neural
  question generation. arXiv preprint arXiv:1705.02012  (2017)

\bibitem{snd_motions}
Zhao, H., Gan, C., Ma, W., Torralba, A.: The sound of motions. CoRR
  \textbf{abs/1904.05979} (2019)

\bibitem{zhao2018sound}
Zhao, H., Gan, C., Rouditchenko, A., Vondrick, C., McDermott, J., Torralba, A.:
  The sound of pixels. In: Proceedings of the European Conference on Computer
  Vision (ECCV). pp. 570--586 (2018)

\bibitem{zhao2018through}
Zhao, M., Li, T., Abu~Alsheikh, M., Tian, Y., Zhao, H., Torralba, A., Katabi,
  D.: Through-wall human pose estimation using radio signals. In: CVPR (2018)

\bibitem{zhou2019vision}
Zhou, H., Liu, Z., Xu, X., Luo, P., Wang, X.: Vision-infused deep audio
  inpainting. In: ICCV (2019)

\end{thebibliography}
